\documentclass[mal,biber]{nowfnt} 

\usepackage[utf8]{inputenc}
\makeatletter
\def\UTFviii@defined#1{%
  \ifx#1\relax
      ?%
  \else\expandafter
    #1%
  \fi
}

\makeatother
\usepackage{hyperref}

\usepackage{graphicx} 
\usepackage{subfig}

\usepackage{algorithm}
\usepackage{algorithmic}

\usepackage{yfonts}
\usepackage{amsmath}
\usepackage{amsfonts}
\usepackage{amssymb}
\DeclareMathOperator{\E}{\mathbb{E}}

\usepackage{tikz}
\usetikzlibrary{decorations.pathreplacing}
\usetikzlibrary{patterns}
\usetikzlibrary{calc}

\usetikzlibrary{quotes,arrows.meta}
\tikzset{
  annotated cuboid/.pic={
    \tikzset{%
      every edge quotes/.append style={midway, auto},
      /cuboid/.cd,
      #1
    }
    \draw [every edge/.append style={pic actions, densely dashed, opacity=.5}, pic actions]
    (0,0,0) coordinate (o) -- ++(-\cubescale*\cubex,0,0) coordinate (a) -- ++(0,-\cubescale*\cubey,0) coordinate (b) edge coordinate [pos=1] (g) ++(0,0,-\cubescale*\cubez)  -- ++(\cubescale*\cubex,0,0) coordinate (c) -- cycle
    (o) -- ++(0,0,-\cubescale*\cubez) coordinate (d) -- ++(0,-\cubescale*\cubey,0) coordinate (e) edge (g) -- (c) -- cycle
    (o) -- (a) -- ++(0,0,-\cubescale*\cubez) coordinate (f) edge (g) -- (d) -- cycle;
    \path [every edge/.append style={pic actions, |-|}]
    ;
  },
  /cuboid/.search also={/tikz},
  /cuboid/.cd,
  width/.store in=\cubex,
  height/.store in=\cubey,
  depth/.store in=\cubez,
  units/.store in=\cubeunits,
  scale/.store in=\cubescale,
  width=10,
  height=10,
  depth=10,
  units=cm,
  scale=.1,
}
\usepackage{verbatim}
\usetikzlibrary{fit}
\usetikzlibrary{backgrounds}
\usetikzlibrary{positioning}

\usepackage{textcomp}
\usepackage{mdframed}

\usetikzlibrary{shapes,arrows}
\usepackage{comment}
\usepackage{pgfplots}

\title{An Introduction to Deep Reinforcement Learning}

\booltrue{authortwocolumn} 
\maintitleauthorlist{
Vincent Fran\c{c}ois-Lavet \\
McGill University\\
vincent.francois-lavet@mcgill.ca
\and
Peter Henderson \\
McGill University\\
peter.henderson@mail.mcgill.ca
\and
Riashat Islam \\
McGill University\\
riashat.islam@mail.mcgill.ca
\and
Marc G. Bellemare \\
Google Brain\\
bellemare@google.com
\and
Joelle Pineau \\
Facebook, McGill University\\
jpineau@cs.mcgill.ca
}

\issuesetup
{%
copyrightowner={V.~Fran\c{c}ois-Lavet, P.~Henderson, R.~Islam, M.~G.~Bellemare and J.~Pineau},
volume        = 11,
issue         = 3-4,
pubyear       = 2018,
isbn          = 978-1-68083-538-0,
eisbn         = 978-1-68083-539-7,
doi           = 10.1561/2200000071,
firstpage     = 1,
 lastpage      = xxx 
}

\usepackage[style=authoryear,backend=biber]{biblatex}
\usepackage{mwe}

\author[1]{Fran\c{c}ois-Lavet,Vincent}
\author[2]{Henderson, Peter}
\author[3]{Islam, Riashat}
\author[4]{Bellemare, Marc G.}
\author[5]{Pineau, Joelle}

\affil[1]{McGill University; vincent.francois-lavet@mcgill.ca}
\affil[2]{McGill University; peter.henderson@mail.mcgill.ca}
\affil[3]{McGill University; riashat.islam@mail.mcgill.ca}
\affil[4]{Google Brain; bellemare@google.com}
\affil[5]{Facebook, McGill University; jpineau@cs.mcgill.ca}

\begin{document}

\makeabstracttitle

\begin{abstract}
Deep reinforcement learning is the combination of reinforcement learning (RL) and deep learning.
This field of research has been able to solve a wide range of complex decision-making tasks that were previously out of reach for a machine.
Thus, deep RL opens up many new applications in domains such as healthcare, robotics, smart grids, finance, and many more.
This manuscript provides an introduction to deep reinforcement learning models, algorithms and techniques.
Particular focus is on the aspects related to generalization and how deep RL can be used for practical applications.
We assume the reader is familiar with basic machine learning concepts.
\end{abstract}

\chapter{Introduction}
\section{Motivation}
A core topic in machine learning is that of sequential decision-making.
This is the task of deciding, from experience, the sequence of actions to perform in an uncertain environment in order to achieve some goals.
Sequential decision-making tasks cover a wide range of possible applications with the potential to impact many domains, such as robotics, healthcare, smart grids, finance, self-driving cars, and many more.

Inspired by behavioral psychology (see e.g., \cite{sutton1984temporal}), reinforcement learning (RL) proposes a formal framework to this problem.
The main idea is that an artificial agent may learn by interacting with its environment, similarly to a biological agent.
Using the experience gathered, the artificial agent should be able to optimize some objectives given in the form of cumulative rewards.
This approach applies in principle to any type of sequential decision-making problem relying on past experience.
The environment may be stochastic, the agent may only observe partial information about the current state, the observations may be high-dimensional (e.g., frames and time series), the agent may freely gather experience in the environment or, on the contrary, the data may be may be constrained (e.g., not access to an accurate simulator or limited data).

Over the past few years, RL has become increasingly popular due to its success in addressing challenging sequential decision-making problems.
Several of these achievements are due to the combination of RL with deep learning techniques \citep{lecun2015deep, schmidhuber2015deep, goodfellow2016deep}.
This combination, called deep RL, is most useful in problems with high dimensional state-space.
Previous RL approaches had a difficult design issue in the choice of features \citep{munos2002variable,bellemare2013arcade}.
However, deep RL has been successful in complicated tasks with lower prior knowledge thanks to its ability to learn different levels of abstractions from data. 
For instance, a deep RL agent can successfully learn from visual perceptual inputs made up of thousands of pixels \citep{mnih2015human}.
This opens up the possibility to mimic some human problem solving capabilities, even in high-dimensional space --- which, only a few years ago, was difficult to conceive.

Several notable works using deep RL in games have stood out for attaining super-human level in playing Atari games from the pixels \citep{mnih2015human}, mastering Go~\citep{silver2016mastering} or beating the world's top professionals at the game of Poker~\citep{brownlibratus,moravvcik2017deepstack}.
Deep RL also has potential for real-world applications such as
robotics \citep{levine2016end, gandhi2017learning, pinto2017asymmetric},
self-driving cars \citep{you2017virtual},
finance \citep{deng2017deep} and smart grids \citep{franccois2017contributions}, to name a few.
Nonetheless, several challenges arise in applying deep RL algorithms.
Among others, exploring the environment efficiently or being able to generalize a good behavior in a slightly different context are not straightforward.
Thus, a large array of algorithms have been proposed for the deep RL framework, depending on a variety of settings of the sequential decision-making tasks. 

\section{Outline}
The goal of this introduction to deep RL is to guide the reader towards effective use and understanding of core methods, as well as provide references for further reading.
After reading this introduction, the reader should be able to understand the key different deep RL approaches and algorithms and should be able to apply them.
The reader should also have enough background to investigate the scientific literature further and pursue research on deep RL.

In Chapter \ref{ch:intro_DL}, we introduce the field of machine learning and the deep learning approach.
The goal is to provide the general technical context and explain briefly where deep learning is situated in the broader field of machine learning.
We assume the reader is familiar with basic notions of supervised and unsupervised learning;
however, we briefly review the essentials.

In Chapter \ref{ch:intro_RL}, we provide the general RL framework along with the case of a Markov Decision Process (MDP).
In that context, we examine the different methodologies that can be used to train a deep RL agent.
On the one hand, learning a value function (Chapter \ref{ch:value-based_methods}) and/or a direct representation of the policy (Chapter \ref{ch:policy-based_methods}) belong to the so-called model-free approaches.
On the other hand, planning algorithms that can make use of a learned model of the environment belong to the so-called model-based approaches (Chapter \ref{ch:model-based}).


We dedicate Chapter \ref{ch:generalization} to the notion of generalization in RL.
Within either a model-based or a model-free approach, we discuss the importance of different elements:
(i)~feature selection,
(ii)~function approximator selection,
(iii)~modifying the objective function and
(iv)~hierarchical learning.
In Chapter \ref{ch:challenges_online}, we present the main challenges of using RL in the online setting.
In particular, we discuss the exploration-exploitation dilemma and the use of a replay memory.

In Chapter \ref{ch:benchmarks}, we provide an overview of different existing benchmarks for evaluation of RL algorithms. Furthermore, we present a set of best practices to ensure consistency and reproducibility of the results obtained on the different benchmarks.

In Chapter \ref{ch:different_settings}, we discuss more general settings than MDPs:
(i)~the Partially Observable Markov Decision Process (POMDP),
(ii)~the distribution of MDPs (instead of a given MDP) along with the notion of transfer learning,
(iii)~learning without explicit reward function and
(iv)~multi-agent systems.
We provide descriptions of how deep RL can be used in these settings.

In Chapter \ref{ch:real-world}, we present broader perspectives on deep RL.
This includes a discussion on applications of deep RL in various domains, along with the successes achieved and remaining challenges (e.g. robotics, self driving cars, smart grids, healthcare, etc.).
This also includes a brief discussion on the relationship between deep RL and neuroscience.

Finally, we provide a conclusion in Chapter \ref{ch:conclusion} with an outlook on the future development of deep RL techniques, their future applications, as well as the societal impact of deep RL and artificial intelligence.

\chapter{Machine learning and deep learning}
\label{ch:intro_DL}

Machine learning provides automated methods that can detect patterns in data and use them to achieve some tasks \citep{christopher2006pattern, murphy2012machine}.
Three types of machine learning tasks can be considered:
\begin{itemize}
\item \textit{Supervised learning} is the task of inferring a classification or regression from labeled training data.
\item \textit{Unsupervised learning} is the task of drawing inferences from datasets consisting of input data without labeled responses.
\item \textit{Reinforcement learning} (RL) is the task of learning how agents ought to take sequences of actions in an environment in order to maximize cumulative rewards.
\end{itemize}

To solve these machine learning tasks, the idea of function approximators is at the heart of machine learning.
There exist many different types of function approximators: linear models \citep{anderson1958introduction}, SVMs \citep{cortes1995support}, decisions tree \citep{liaw2002classification, geurts2006extremely}, Gaussian processes \citep{rasmussen2004gaussian}, deep learning \citep{lecun2015deep, schmidhuber2015deep, goodfellow2016deep}, etc.

In recent years, mainly due to recent developments in deep learning, machine learning has undergone dramatic improvements when learning from high-dimensional data such as time series, images and videos.
These improvements can be linked to the following aspects:
(i)~an exponential increase of computational power with the use of GPUs and distributed computing \citep{krizhevsky2012imagenet},
(ii)~methodological breakthroughs in deep learning \citep{srivastava2014dropout, ioffe2015batch, he2016deep, szegedy2016inception, klambauer2017self},
(iii)~a growing eco-system of softwares such as Tensorflow \citep{abadi2016tensorflow} 
and datasets such as ImageNet \citep{russakovsky2015imagenet}. 
All these aspects are complementary and, in the last few years, they have lead to a virtuous circle for the development of deep learning.

In this chapter, we discuss the supervised learning setting along with the key concepts of bias and overfitting.
We briefly discuss the unsupervised setting with tasks such as data compression and generative models.
We also introduce the deep learning approach that has become key to the whole field of machine learning.
Using the concepts introduced in this chapter, we cover the reinforcement learning setting in later chapters.



\section{Supervised learning and the concepts of bias and overfitting}
In its most abstract form, supervised learning consists in finding a function $\textstyle f:\mathcal X \rightarrow \mathcal Y$\label{ntn:f-XY} that takes as input $x \in \mathcal X$ and gives as output $y \in \mathcal Y$ ($\mathcal X$ and $\mathcal Y$ depend on the application):
\begin{equation}
y=f(x).
\end{equation}

\begin{figure}[ht!]
    \centering
    \includegraphics[width=1\textwidth]{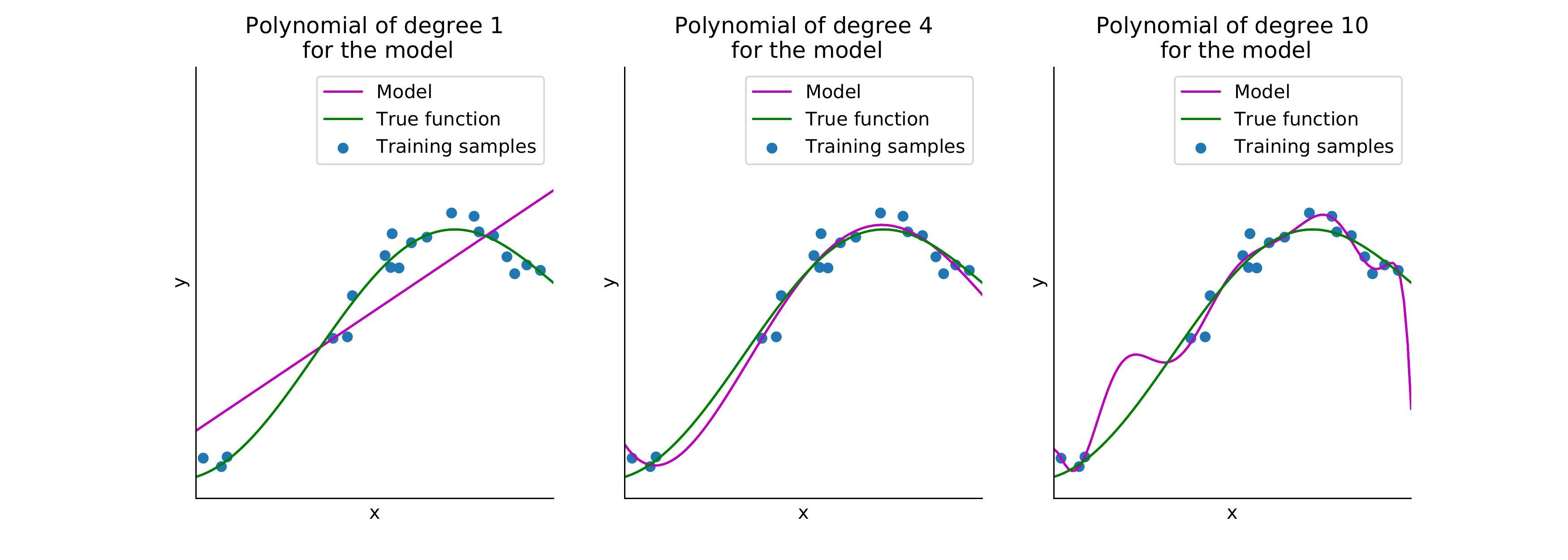}
    \caption
    [Illustration of overfitting and underfitting for a simple 1D regression task in supervised learning.]
    {Illustration of overfitting and underfitting for a simple 1D regression task in supervised learning (based on one example from the library scikit-learn \citep{pedregosa2011scikit}).
    In this illustration, the data points $(x,y)$ are noisy samples from a true function represented in green.
    In the left figure, the degree 1 approximation is underfitting, which means that it is not a good model, even for the training samples; on the right, the degree 10 approximation is a very good model for the training samples but is overly complex and fails to provide a good generalization.}
    \label{fig:under_over_f}
\end{figure}

A supervised learning algorithm can be viewed as a function that maps a dataset $D_{LS}$ of learning samples $(x,y) \overset{\text{i.i.d.}}{\sim} (X,Y)$ into a model. The prediction of such a model at a point $x \in \mathcal X$ of the input space is denoted by $f (x \mid D_{LS})$. Assuming a random sampling scheme, s.t. $D_{LS} \sim \mathcal D_{LS}$%
, $f (x \mid D_{LS})$ is a random variable, and so is its average error over the input space. The expected value of this quantity is given by:
\begin{equation}
\begin{split}
I[f] &
= \underset{X}{\mathbb E}~{\underset{D_{LS}} {\mathbb E}~{\underset{Y|X}{\mathbb E}{L\left(Y, f(X \mid D_{LS})\right)}}},\\
\end{split}
\label{eq:bias-overfit}
\end{equation}
where $L(\cdot , \cdot)$ is the loss function.
If $L(y, \hat y) = (y-\hat y)^2$, the error decomposes naturally into a sum of a bias term and a variance term\footnote{
The bias-variance decomposition \citep{geman1992neural} is given by:
\begin{equation}
{\underset{D_{LS}} {\mathbb E}~{\underset{Y|X}{\mathbb E}{(Y - f(X \mid D_{LS}))^2}}} = \sigma^2(x) + \text{bias}^2(x),
\end{equation}
where
\begin{equation}
\begin{split}
& \text{bias}^2(x) \triangleq \left(\mathbb E_{Y \mid x}(Y) - \mathbb E_{D_{LS}} f(x \mid D_{LS})\right)^2,\\
& \sigma^2(x) \triangleq \underbrace{\mathbb E_{Y|x}\left( Y - \mathbb E_{Y \mid x}(Y) \right)^2}_{\text{Internal variance}}
+ \underbrace{\mathbb E_{D_{LS}}{\Big(f(x \mid D_{LS}) - \mathbb E_{D_{LS}}{f(x \mid D_{LS})}\Big)^2}}_{\text{Parametric variance}},\\
\end{split}
\end{equation}
}. This bias-variance decomposition can be useful because it highlights a tradeoff between an error due to erroneous assumptions in the model selection/learning algorithm (the bias) and an error due to the fact that only a finite set of data is available to learn that model (the parametric variance). Note that the parametric variance is also called the overfitting error\footnote{For any given model, the parametric variance goes to zero with an arbitrary large dataset by considering the strong law of convergence.}.
Even though there is no such direct decomposition for other loss functions \citep{james2003variance}, there is always a tradeoff between a sufficiently rich model (to reduce the model bias, which is present even when the amount of data would be unlimited) and a model not too complex (so as to avoid overfitting to the limited amount of data). Figure \ref{fig:under_over_f} provides an illustration.

Without knowing the joint probability distribution, it is impossible to compute $I[f]$. Instead, we can compute the empirical error on a sample of data. Given $n$ data points $(x_i, y_i)$, the empirical error is
$$ I_S[f] = \frac{1}{n} \sum_{i=1}^n L(y_i,f(x_i)). $$

The \textit{generalization error} is the difference between the error on a sample set (used for training) and the error on the underlying joint probability distribution.
It is defined as

$$ G = I[f] - I_S[f]. $$

In machine learning, the complexity of the function approximator provides upper bounds on the generalization error. The generalization error can be bounded by making use of complexity measures, such as the Rademacher complexity \citep{bartlett2002rademacher}, or the VC-dimension \citep{vapnik1998statistical}. However, even though it lacks strong theoretical foundations, it has become clear in practice that the strength of deep neural networks is their generalization capability, even with a high number of parameters (hence a potentially high complexity) \citep{zhang2016understanding}.

\section{Unsupervised learning}
Unsupervised learning is a branch of machine learning that learns from data that do not have any label. It relates to using and identifying patterns in the data for tasks such as data compression or generative models.


Data compression or dimensionality reduction involve encoding information using a smaller representation (e.g., fewer bits) than the original representation.
For instance, an auto-encoder consists of an encoder and a decoder. The encoder maps the original image $x_i \in \mathbb R^M$ onto a low-dimensional representation $z_i = e(x_i; \theta_e)\in \mathbb R^m$, where $m << M$; the decoder maps these features back to a high-dimensional representation $d(z_i; \theta_d) \approx e^{-1}(z_i; \theta_e)$. Auto-encoders can be trained by optimizing for the reconstruction of the input through supervised learning objectives. 

Generative models aim at approximating the true data distribution of a training set so as to generate new data points from the distribution. Generative adversarial networks \citep{goodfellow2014generative} use an adversarial process, in which two models are trained simulatenously: a generative model G captures the data distribution, while a discriminative model D estimates whether a sample comes from the training data rather than G. The training procedure corresponds to a minimax two-player game.


\section{The deep learning approach}
Deep learning relies on a function $\textstyle f:\mathcal X \rightarrow \mathcal Y$ parameterized with $\theta \in \mathbb R^{n_\theta}$\label{ntn:n_theta} $ (n_\theta\in \mathbb N$):
\begin{equation}
y=f(x;\theta).
\end{equation}
A deep neural network is characterized by a succession of multiple processing layers. Each layer consists in a non-linear transformation and the sequence of these transformations leads to learning different levels of abstraction \citep{erhan2009visualizing, olah2017feature}.

First, let us describe a very simple neural network with one fully-connected hidden layer (see Fig \ref{fig:fc}). The first layer is given the input values (i.e., the input features) $x$ in the form of a column vector of size $n_x$ ($n_x \in \mathbb N$\label{ntn:natural_number}). The values of the next hidden layer are a transformation of these values by a non-linear parametric function, which is a matrix multiplication by $W_{1}$ of size $n_{h} \times n_x$ ($n_h \in \mathbb N$), plus a bias term $b_{1}$ of size $n_{h}$, followed by a non-linear transformation: 
\begin{equation}
h=A(W_{1} \cdot x + b_{1}),\label{ntn:NN_transfo}
\end{equation}
where A is the \textit{activation function}.
This non-linear activation function is what makes the transformation at each layer non-linear, which ultimately provides the expressivity of the neural network.
The hidden layer $h$ of size $n_{h}$ can in turn be transformed to other sets of values up to the last transformation that provides the output values $y$. In this case:
\begin{equation}
y=(W_{2} \cdot h + b_{2}),
\end{equation}
where $W_{2}$ is of size $n_y \times n_{h}$ and $b_{2}$ is of size $n_y$ ($n_y \in \mathbb N$).
\def\layersep{2.5cm}
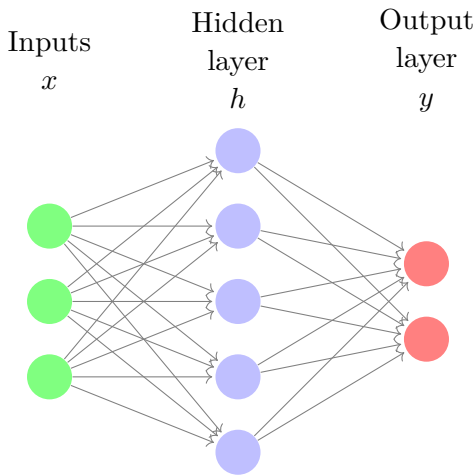
\begin{figure}[ht!]
 \centering
\begin{tikzpicture}[shorten >=1pt,->,draw=black!50, node distance=\layersep]
    \tikzstyle{every pin edge}=[<-,shorten <=1pt]
    \tikzstyle{neuron}=[circle,fill=black!25,minimum size=17pt,inner sep=0pt]
    \tikzstyle{input neuron}=[neuron, fill=green!50];
    \tikzstyle{output neuron}=[neuron, fill=red!50];
    \tikzstyle{hidden neuron}=[neuron, fill=blue!25];
    \tikzstyle{annot} = [text width=4em, text centered]

    \foreach \name / \y in {1,...,3}
        \node[input neuron] (I-\name) at (0,-\y-0.5) {};

    \foreach \name / \y in {1,...,5}
        \path[yshift=0.5cm]
            node[hidden neuron] (H-\name) at (\layersep,-\y cm) {};

    \foreach \name / \y in {1,...,2}
        \path[yshift=-1cm]
         node[output neuron] (O-\name) at (2*\layersep,-\y cm) {};

    \foreach \source in {1,...,3}
        \foreach \dest in {1,...,5}
            \path (I-\source) edge (H-\dest);

    \foreach \source in {1,...,5}
        \foreach \dest in {1,...,2}
        \path (H-\source) edge (O-\dest);

    \node[annot,above of=H-1, node distance=1.2cm] (hl) {Hidden layer\\$h$};
    \node[annot,left of=hl] {Inputs\\$x$};
    \node[annot,right of=hl] {Output layer\\$y$};
\end{tikzpicture}
\caption{Example of a neural network with one hidden layer.}
\label{fig:fc}
\end{figure}

All these layers are trained to minimize the empirical error $I_S[f]$.
The most common method for optimizing the parameters of a neural network is based on gradient descent via the backpropagation algorithm \citep{rumelhart1988learning}.
In the simplest case, at every iteration, the algorithm changes its internal parameters $\theta$ so as to fit the desired function:
\begin{equation}
\begin{split}
\theta \leftarrow \theta - \alpha \nabla_{\theta} I_S[f],
\end{split}
\end{equation}
where $\alpha$ is the learning rate.

In current applications, many different types of neural network layers have appeared beyond the simple feedforward networks just introduced. 
Each variation provides specific advantages, depending on the application (e.g., good tradeoff between bias and overfitting in a supervised learning setting).
In addition, within one given neural network, an arbitrarily large number of layers is possible, and the trend in the last few years is to have an ever-growing number of layers, with more than $100$ in some supervised learning tasks \citep{szegedy2017inception}. 
We merely describe here two types of layers that are of particular interest in deep RL (and in many other tasks).

\begin{figure}[ht!]
  \centering
  \begin{tikzpicture}[scale = 1.4]
    \node at (1.5,4){\begin{tabular}{c}input image\\or input feature map\end{tabular}};

    \draw (0,0) -- (3,0) -- (3,3) -- (0,3) -- (0,0);

    \draw[color=blue] (2,0.5) -- (2.5,0.5) -- (2.5,1) -- (2,1) -- (2,0.5);
    \draw[color=blue] (1,1) -- (1.5,1) -- (1.5,1.5) -- (1,1.5) -- (1,1);

    \draw [->, thick, color=blue] (3,3) to [out=-30,in=120] (4,2.2);
    \draw[color=blue] (3.5,3) -- (4,3) -- (4,3.5) -- (3.5,3.5) -- (3.5,3);
    \node[color=blue] at (3.5,3.7){filter};
    \node[color=blue] at (3.625,3.125){0};
    \node[color=blue] at (3.875,3.125){1};
    \node[color=blue] at (3.625,3.375){1};
    \node[color=blue] at (3.875,3.375){0};

    \draw[color=blue] (2.5,1) -- (5.6,0.5);
    \draw[color=blue] (2.5,0.5) -- (5.6,0.5);
    \node[color=blue] at (5.7,0.5){2};

    \draw[color=blue] (1.5,1.5) -- (4.75,0.9);
    \draw[color=blue] (1.5,1) -- (4.75,0.9);
    \node[color=blue] at (4.9,0.9){0};

    \node at (6.5,4.5){\begin{tabular}{c}output feature maps\end{tabular}};

    \draw[fill=black,opacity=0.2,draw=black] (6,2) -- (8,2) -- (8,4) -- (6,4) -- (6,2);
    \draw[fill=black,opacity=0.2,draw=black] (5.5,1.5) -- (7.5,1.5) -- (7.5,3.5) -- (5.5,3.5) -- (5.5,1.5);
    \draw[fill=black,opacity=0.2,draw=black] (4,0) -- (6.2,0) -- (6.2,2.2) -- (4,2.2) -- (4,0);

    \foreach \i in {0,...,11}{
      \foreach \j in {0,...,11}{
        \pgfmathparse{11-\j}
      \ifthenelse{ \i=\j \OR \i=\pgfmathresult }{
      \node at (0.125+\i*0.25,0.125+\j*0.25){1};
      }{
      \node at (0.125+\i*0.25,0.125+\j*0.25){0};
      }
      }
    }

    \foreach \i in {1,...,11}
    {
      \draw[dashed,opacity=0.3] (\i*0.25,0) -- (\i*0.25,3);
      \draw[dashed,opacity=0.3] (0,\i*0.25) -- (3,\i*0.25);
    }

    \foreach \i in {1,...,10}
    {
      \draw[dashed,opacity=0.3] (4+\i*0.2,0) -- (4+\i*0.2,2.2);
      \draw[dashed,opacity=0.3] (4,\i*0.2) -- (4+2.2,\i*0.2);
    }
  \end{tikzpicture}
  \caption
  [Illustration of a convolutional layer with one input feature map that is convolved by different filters to yield the output feature maps.]
  {Illustration of a convolutional layer with one input feature map that is convolved by different filters to yield the output feature maps. The parameters that are learned for this type of layer are those of the filters. For illustration purposes, some results are displayed for one of the output feature maps with a given filter (in practice, that operation is  followed by a non-linear activation function).}
  \label{fig:convolutional-layer}
\end{figure}
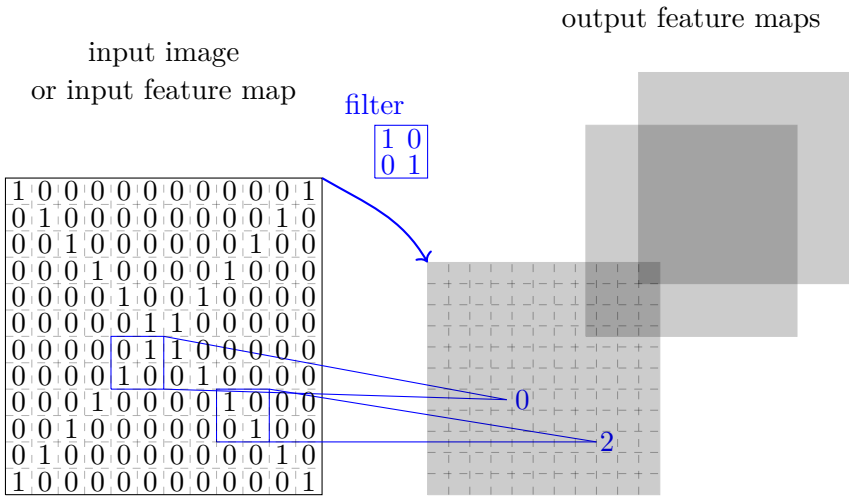

Convolutional layers \citep{lecun1995convolutional} are particularly well suited for images and sequential data (see Fig \ref{fig:convolutional-layer}), mainly due to their translation invariance property.
The layer's parameters consist of a set of learnable filters (or kernels), which have a small receptive field and which apply a convolution operation to the input, passing the result to the next layer. As a result, the network learns filters that activate when it detects some specific features. In image classification, the first layers learn how to detect edges, textures and patterns; then the following layers are able to detect parts of objects and whole objects \citep{erhan2009visualizing, olah2017feature}.
In fact, a convolutional layer is a particular kind of feedforward layer, with the specificity that many weights are set to $0$ (not learnable) and that other weights are shared.

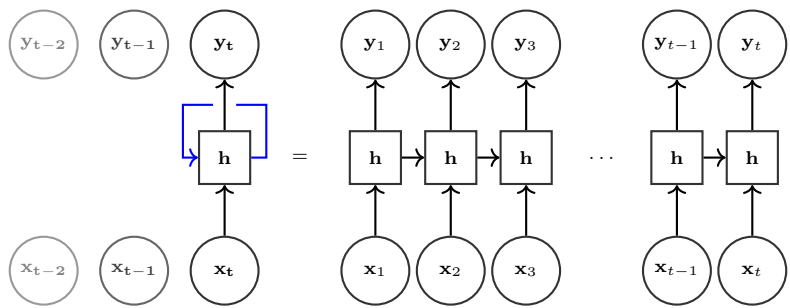
\begin{figure}[t]
\centering
\begin{tikzpicture}[->,thick]
\scriptsize
\tikzstyle{main}=[circle, minimum size = 9mm, thick, draw =black!80, node distance = 12mm]
\tikzstyle{rr}=[rounded rectangle, rounded rectangle west arc=5pt, rounded rectangle east arc=50pt, minimum size = 7mm, thick, draw =black!80, node distance = 12mm]

    \node[main, fill = white!100] (y) at (-1,1.5) {$\mathbf{y_t}$};
    \node[rr, fill = white!100] (h) at (-1,0) {$\mathbf{h}$};
    \node[main, fill = white!100] (x) at (-1,-1.5) {$\mathbf{x_t}$};
        \path (x) edge (h);
        \path (h) edge (y);
    \node[opacity=0.75,main, fill = white!100, left of=x] (x-1) {$\mathbf{x_{t-1}}$};
    \node[opacity=0.5,main, fill = white!100, left of=x-1] (x-2) {$\mathbf{x_{t-2}}$};
    \node[opacity=0.75,main, fill = white!100, left of=y] (y-1) {$\mathbf{y_{t-1}}$};
    \node[opacity=0.5,main, fill = white!100, left of=y-1] (y-2) {$\mathbf{y_{t-2}}$};

    \draw [-, color=blue] (h.east) -- ($(h.east)+(0.2,0)$) -- ($(h.east)+(0.2,0.7)$) -- ($(h.east)+(-0.2,0.7)$);
    \draw [->, color=blue] ($(h.west)+(0.2,0.7)$) -- ($(h.west)+(-0.2,0.7)$) -- ($(h.west)+(-0.2,0)$) -- (h.west);

    \node[] at (0,0) {$=$};

\foreach \name in {1,...,3}
    \node[main, fill = white!100] (y\name) at (\name,1.5) {$\mathbf{y}_\name$};
\foreach \name in {1,...,3}
    \node[rr, fill = white!100] (h\name) at (\name,0) {$\mathbf{h}$};
\foreach \name in {1,...,3}
    \node[main, fill = white!100] (x\name) at (\name,-1.5) {$\mathbf{x}_\name$};

\node[] (h4) at (4,0) {$\ldots$};
\node[main, fill = white!100] (y5) at (5,1.5) {$\mathbf{y}_{t-1}$};
\node[rr, fill = white!100] (h5) at (5,0) {$\mathbf{h}$};
\node[main, fill = white!100] (x5) at (5,-1.5) {$\mathbf{x}_{t-1}$};
\node[main, fill = white!100] (y6) at (6,1.5) {$\mathbf{y}_t$};
\node[rr, fill = white!100] (h6) at (6,0) {$\mathbf{h}$};
\node[main, fill = white!100] (x6) at (6,-1.5) {$\mathbf{x}_t$};

\foreach \h in {1,...,6}
       {\ifthenelse{\h = 4}{}{
        \path (x\h) edge (h\h);
        \path (h\h) edge (y\h);
       }}
\foreach \current/\next in {1/2,2/3,3/4,4/5,5/6}
       {\ifthenelse{\current = 4 \OR \next = 4}{}{
        \path (h\current) edge (h\next);
       }}
\end{tikzpicture}
\caption
[Illustration of a simple recurrent neural network.]
{Illustration of a simple recurrent neural network. The layer denoted by "h" may represent any non linear function that takes two inputs and provides two outputs. On the left is the simplified view of a recurrent neural network that is applied recursively to $(x_{t},y_{t})$ for increasing values of $t$ and where the blue line presents a delay of one time step. On the right, the neural network is unfolded with the implicit requirement of presenting all inputs and outputs simultaneously.}
\label{fig:recurrent-layer}
\end{figure}

Recurrent layers are particularly well suited for sequential data (see Fig \ref{fig:recurrent-layer}). Several different variants provide particular benefits in different settings. One such example is the long short-term memory network (LSTM) \citep{hochreiter1997long}, which is able to encode information from long sequences, unlike a basic recurrent neural network. Neural Turing Machines (NTMs) \citep{graves2014neural} are another such example. In such systems, a differentiable "external memory" is used for inferring even longer-term dependencies than LSTMs with low degradation.

Several other specific neural network architectures have also been studied to improve generalization in deep learning. For instance, it is possible to design an architecture in such a way that it automatically focuses on only some parts of the inputs with a mechanism called attention \citep{xu2015show, vaswani2017attention}. Other approaches aim to work with symbolic rules by learning to create programs \citep{reed2015neural, neelakantan2015neural, johnson2017inferring, chen2017learning}. 

To be able to actually apply the deep RL methods described in the later chapters, the reader should have practical knowledge of applying deep learning methods in simple supervised learning settings (e.g., MNIST classification). For information on topics such as the importance of input normalizations, weight initialization techniques, regularization techniques and the different variants of gradient descent techniques, the reader can refer to several reviews on the subject \citep{lecun2015deep, schmidhuber2015deep, goodfellow2016deep} as well as references therein.

In the following chapters, the focus is on reinforcement learning, in particular on methods that scale to deep neural network function approximators.
These methods allows for learning a wide variety of challenging sequential decision-making tasks directly from rich high-dimensional inputs.

\chapter{Introduction to reinforcement learning}
\label{ch:intro_RL}
Reinforcement learning (RL) is the area of machine learning that deals with sequential decision-making.
In this chapter, we describe how the RL problem can be formalized as an agent that has to make decisions in an environment to optimize a given notion of cumulative rewards.
It will become clear that this formalization applies to a wide variety of tasks and captures many essential features of artificial intelligence such as a sense of cause and effect as well as a sense of uncertainty and nondeterminism.
This chapter also introduces the different approaches to learning sequential decision-making tasks and how deep RL can be useful.

A key aspect of RL is that an agent \textit{learns} a good behavior.
This means that it modifies or acquires new behaviors and skills incrementally.
Another important aspect of RL is that it uses trial-and-error \textit{experience} (as opposed to e.g., dynamic programming that assumes full knowledge of the environment a priori).
Thus, the RL agent does not require complete knowledge or control of the environment; it only needs to be able to interact with the environment and collect information.
In an \textit{offline} setting, the experience is acquired a priori, then it is used as a batch for learning (hence the offline setting is also called batch RL).
This is in contrast to the \textit{online} setting where data becomes available in a sequential order and is used to progressively update the behavior of the agent.
In both cases, the core learning algorithms are essentially the same but the main difference is that in an online setting, the agent can influence how it gathers experience so that it is the most useful for learning.
This is an additional challenge mainly because the agent has to deal with the \textit{exploration/exploitation} dilemma while learning (see \S\ref{sec:explo-exploit} for a detailed discussion).
But learning in the online setting can also be an advantage since the agent is able to gather information specifically on the most interesting part of the environment.
For that reason, even when the environment is fully known, RL approaches may provide the most computationally efficient approach in practice as compared to some dynamic programming methods that would be inefficient due to this lack of specificity.

\section{Formal framework}

\subsection*{The reinforcement learning setting}
\begin{figure}[!ht]
\centering
\tikzstyle{block} = [rectangle, draw, text width=7em, text centered, rounded corners, minimum height=3em]
\tikzstyle{line} = [draw, -latex]

\begin{tikzpicture}[node distance = 6em, auto, thick]
    \node [block] (Agent) {Agent};
    \node [block, below of=Agent] (Environment) {Environment\\$s_t \rightarrow s_{t+1}$};

     \path [line] (Agent.180) --++ (-4em,-0em) |- node [near start]{$a_t$} (Environment.180);
     \path [line] (Environment.10) --++ (4.25em,0em) |- node [near start] {$\omega_{t+1}$} (Agent.-10);
     \path [line] (Environment.-10) --++ (6em,0em) |- node [near start, right] {$r_{t}$} (Agent.10);
\end{tikzpicture}

\caption{Agent-environment interaction in RL.}
\label{fig:RL_problem}
\end{figure}
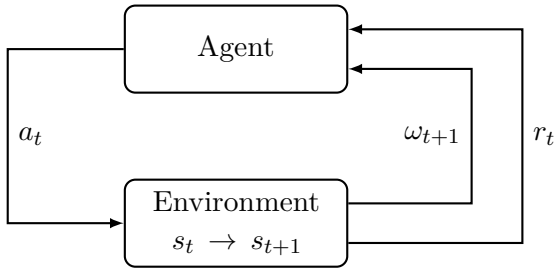

The general RL problem is formalized as a discrete time stochastic control process where an agent interacts with its environment in the following way: the agent starts, in a given state within its environment $s_{0} \in \mathcal S$, by gathering an initial observation $\omega_0 \in \Omega$\label{ntn:obs_space}.
At each time step~$t$, the agent has to take an action $a_t \in \mathcal A$\label{ntn:act_space}. As illustrated in Figure \ref{fig:RL_problem}, it follows three consequences: (i)~the agent obtains a reward $r_{t} \in \mathcal R$\label{ntn:rwd_space}, (ii)~the state transitions to $s_{t+1} \in \mathcal S$\label{ntn:stt_space}, and (iii)~the agent obtains an observation $\omega_{t+1} \in  \Omega$.
This control setting was first proposed by \cite{bellman1957dynamic} and later extended to learning by \cite{barto1983neuronlike}. Comprehensive treatment of RL fundamentals are provided by \cite{sutton1998introduction}. Here, we review the main elements of RL before delving into deep RL in the following chapters.

\subsection*{The Markov property}

For the sake of simplicity, let us consider first the case of Markovian stochastic control processes \citep{norris1998markov}.
\begin{definition}
A discrete time stochastic control process is Markovian (i.e., it has the Markov property) if
\begin{itemize}
\item $\mathbb P(\omega_{t+1} \mid \omega_t, a_t) = \mathbb P(\omega_{t+1} \mid \omega_t, a_t, \ldots, ,\omega_0, a_0)$, and
\item $\mathbb P(r_{t} \mid \omega_t, a_t) = \mathbb P(r_{t} \mid \omega_t, a_t, \ldots, ,\omega_0, a_0)$.
\end{itemize}
\end{definition}
The Markov property means that the future of the process only depends on the current observation, and the agent has no interest in looking at the full history.

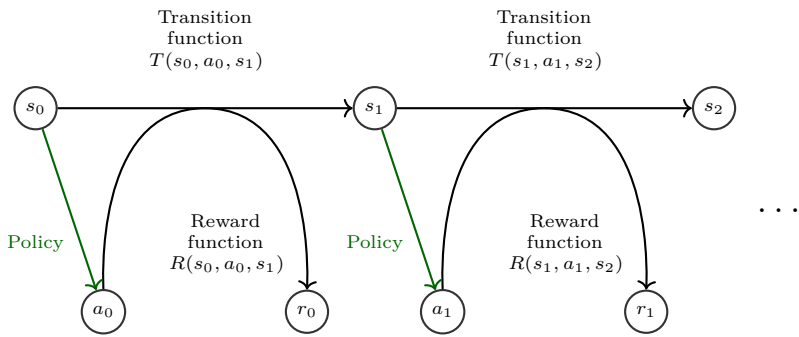
\begin{figure}[ht!]
\centering
\def\dist{5}
\def\ra{0.9}
\begin{tikzpicture}[->,thick, scale=0.9]
\scriptsize
\tikzstyle{main}=[circle, minimum size = \ra, thick, draw =black!80, node distance = 12mm]
\tikzstyle{rr}=[rounded rectangle, rounded rectangle west arc=5pt, rounded rectangle east arc=50pt, minimum size = 7mm, thick, draw =black!80, node distance = 12mm]

\foreach \name in {0,...,2}
    \node[main, fill = white!100] (s\name) at (\dist*\name,0) {$s_\name$};
\foreach \name in {0,...,1}
    \node[main, fill = white!100] (a\name) at (\dist*\name+1,-3) {$a_\name$};
\foreach \name in {0,...,1}
    \node[main, fill = white!100] (r\name) at (\dist*\name+4,-3) {$r_\name$};
\foreach \name in {0,...,1}{
    \node[] (tb\name) at (\dist*\name+2,0) {};
    \node[] (tf\name) at (\dist*\name+3,0) {};
}

\node[font=\Large] (dots) at (\dist*2+1,-1.5) {\ldots};

\foreach \name in {0,...,1}
    \draw [] plot [smooth, tension=2] coordinates { (a\name.north) (\dist*\name+\dist/2,0) (r\name.north) };

\foreach \cur/\nex in {0/1,1/2}
       {
        \path (s\cur) edge (s\nex);
        \path [black!60!green] (s\cur) edge (a\cur);
        \node [black!60!green] (pol\cur) at (\dist*\cur,-2) {Policy};
        \node [text width=1.5cm, align=center] (pol\cur) at (\dist*\cur+2.8,-2) {Reward function $R(s_\cur, a_\cur, s_\nex)$};
        \node [text width=1.8cm, align=center] (pol\cur) at (\dist*\cur+\dist/2,1) {Transition function $T(s_\cur, a_\cur, s_\nex)$};
       }

\end{tikzpicture}
\caption{Illustration of a MDP. At each step, the agent takes an action that changes its state in the environment and provides a reward.}
\label{fig:MDP}
\end{figure}

A Markov Decision Process (MDP) \citep{bellman1957markovian} is a discrete time stochastic control process defined as follows:
\begin{definition}
An MDP is a 5-tuple $(\mathcal S,\mathcal A,T,R,\gamma)$ where:
\begin{itemize}
\item $\mathcal S$ is the state space,
\item $\mathcal A$ is the action space,
\item $T: \mathcal S \times \mathcal A \times \mathcal S \to [0,1]$\label{ntn:trans_fct} is the transition function (set of conditional transition probabilities between states),
\item $R: \mathcal S \times \mathcal A \times \mathcal S \to \mathcal R$\label{ntn:rwd_fct} is the reward function, where $\mathcal R$ is a continuous set of possible rewards in a range $R_{\text{max}} \in \mathbb{R}^+$\label{ntn:R_max} (e.g., $[0,R_{\text{max}}]$),
\item $\gamma \in [0, 1)$ is the discount factor.
\end{itemize}
\end{definition}
The system is fully observable in an MDP, which means that the observation is the same as the state of the environment: $\omega_t = s_t$.
At each time step $t$, the probability of moving to $s_{t+1}$ is given by the state transition function $T(s_t,a_t,s_{t+1})$
and the reward is given by a bounded reward function $R(s_t,a_t,s_{t+1}) \in \mathcal R$. This is illustrated in Figure \ref{fig:MDP}.
Note that more general cases than MDPs are introduced in Chapter~\ref{ch:different_settings}.

\subsection*{Different categories of policies}
A policy defines how an agent selects actions.
Policies can be categorized under the criterion of being either stationary or non-stationary.
A non-stationary policy depends on the time-step and is useful for the finite-horizon context where the cumulative rewards that the agent seeks to optimize are limited to a finite number of future time steps \citep{bertsekas1995dynamic}. In this introduction to deep RL, infinite horizons are considered and the policies are stationary\footnote{The formalism can be directly extended to the finite horizon context. In that case, the policy and the cumulative expected returns should be time-dependent.}.

Policies can also be categorized under a second criterion of being either deterministic or stochastic:
\begin{itemize}
\item In the deterministic case, the policy is described by $\pi(s):\mathcal S \rightarrow \mathcal A$.
\item In the stochastic case, the policy is described by $\pi(s,a):\mathcal S \times \mathcal A \rightarrow [0,1]$ where $\pi(s,a)$ denotes the probability that action $a$ may be chosen in state $s$.
\end{itemize}

\subsection*{The expected return}
Throughout this survey, we consider the case of an RL agent whose goal is to find a policy $\pi(s,a) \in \Pi$\label{ntn:policy}
, so as to optimize an expected return $V^\pi(s):\mathcal S \rightarrow \mathbb R$ (also called V-value function) such that
\begin{equation}
V^\pi(s)=\mathbb E \left[ \sum \nolimits_{k=0}^{\infty} \gamma^k r_{t+k} \mid s_t=s, \pi\right],
\label{def_V}
\end{equation}
where:
\begin{itemize}
\item $r_{t} = \underset{a \sim \pi(s_t,\cdot)}{\mathbb E} R \big(s_{t},a, s_{t+1} \big)$,
\item $\mathbb P \big( s_{t+1} | s_{t}, a_t \big) = T(s_{t},a_t,s_{t+1})$ with $a_t \sim \pi(s_t,\cdot)$,
\end{itemize}
From the definition of the expected return, the optimal expected return can be defined as:
\begin{equation}
V^*(s)=\operatorname*{max}_{\pi\in \Pi} V^\pi(s).
\label{def_opti_V}
\end{equation}

In addition to the V-value function, a few other functions of interest can be introduced. The Q-value function $Q^\pi(s,a):\mathcal S \times A \rightarrow \mathbb R$ is defined as follows: 
\begin{equation}
Q^\pi(s,a)=\operatorname*{\mathbb{E}} \left[ \sum \nolimits_{k=0}^{\infty} \gamma^{k} r_{t+k} \mid s_t=s, a_t=a, \pi \right].
\label{def_Q}
\end{equation}
This equation can be rewritten recursively in the case of an MDP using Bellman's equation:
\begin{equation}
Q^{\pi}(s, a) =  \sum_{s' \in S} T(s,a,s') \left( R(s,a,s')+\gamma Q^{\pi}(s', a=\pi(s')) \right).
\label{def_Q_recursive}
\end{equation}
Similarly to the V-value function, the optimal Q-value function $Q^*(s,a)$ can also be defined as
\begin{equation}
Q^*(s,a)=\operatorname*{max}_{\pi\in \Pi} Q^\pi(s,a).
\label{def_opti_Q}
\end{equation}
The particularity of the Q-value function as compared to the V-value function is that the optimal policy can be obtained directly from $Q^*(s,a)$:
\begin{equation}
\pi^*(s)=\operatorname*{argmax}_{a \in \mathcal A} Q^*(s,a).
\label{def_opti_pi}
\end{equation}
The optimal V-value function $V^*(s)$ is the expected discounted reward when in a given state $s$ while following the policy $\pi^*$ thereafter.
The optimal Q-value $Q^*(s, a)$ is the expected discounted return when in a given state $s$ and for a given action $a$ while following the policy $\pi^*$ thereafter.

It is also possible to define the advantage function 
\begin{equation}
A^{\pi}(s,a)=Q^{\pi}(s,a)-V^{\pi}(s).
\label{def_A}
\end{equation}
This quantity describes how good the action $a$ is, as compared to the expected return when following directly policy $\pi$.

Note that one straightforward way to obtain estimates of either $V^\pi(s)$, $Q^\pi(s,a)$ or $A^\pi(s,a)$ is to use Monte Carlo methods, i.e. defining an estimate by performing several simulations from $s$ while following policy $\pi$. In practice, we will see that this may not be possible in the case of limited data. In addition, even when it is possible, we will see that other methods should usually be preferred for computational efficiency.

\section{Different components to learn a policy}
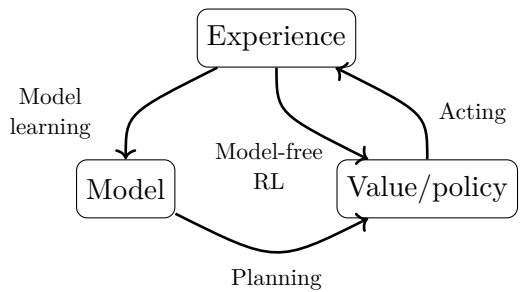
\begin{figure}[ht!]
 \centering
\tikzstyle{block} = [rectangle, draw, text centered, rounded corners, minimum height=2em]
\tikzstyle{line} = [draw, -latex]
\begin{tikzpicture}
  \tikzstyle{transition}=[rectangle,thick,draw=black!75,
          fill=black!20,minimum size=4mm]

  \node[block](experience) at (0,0) {Experience};
  \node[block](valpol) at (2,-2) {Value/policy};
  \node[block](model) at (-2,-2) {Model};

  \node(act) at (2,-1) {};
  \node(plan) at (0,-3) {};
  \node(mod) at (-2,-1) {};
  \node(dir) at (0,-1) {};

  \node[right of=act, xshift=-4mm, scale=0.8] {Acting};
  \node[left of=mod, align=center, scale=0.8] {Model\\learning};
  \node[below of=plan, yshift=8mm, scale=0.8] {Planning};
  \node[left of=dir, xshift=9mm, yshift=-7mm, align=center, scale=0.8] {Model-free\\RL};

\draw[line width=1pt, ->] (experience) .. controls(mod) .. (model);
\draw[line width=1pt, ->] (model) .. controls(plan) .. (valpol);
\draw[line width=1pt, ->] (valpol) .. controls(act) .. (experience);
\draw[line width=1pt, ->] (experience) .. controls(dir) .. (valpol);

\end{tikzpicture}
 \caption{General schema of the different methods for RL. The direct approach uses a representation of either a value function or a policy to act in the environment. The indirect approach makes use of a model of the environment.}
 \label{fig:different_approach_RL}
\end{figure}

An RL agent includes one or more of the following components:
\begin{itemize}
\item a representation of a \textit{value function} that provides a prediction of how good each state or each state/action pair is,
\item a direct representation of the \textit{policy} $\pi(s)$ or $\pi(s,a)$, or
\item a \textit{model} of the environment (the estimated transition function and the estimated reward function) in conjunction with a planning algorithm.
\end{itemize}
The first two components are related to what is called \textit{model-free} RL and are discussed in Chapters \ref{ch:value-based_methods}, \ref{ch:policy-based_methods}.
When the latter component is used, the algorithm is referred to as \textit{model-based} RL, which is discussed in Chapter \ref{ch:model-based}. A combination of both and why using the combination can be useful is discussed in \S\ref{sec:integr_learn_plan}.
A schema with all possible approaches is provided in Figure \ref{fig:different_approach_RL}. 

For most problems approaching real-world complexity, the state space is high-dimensional (and possibly continuous).
In order to learn an estimate of the model, the value function or the policy, there are two main advantages for RL algorithms to rely on deep learning:
\begin{itemize}
\item Neural networks are well suited for dealing with high-dimensional sensory inputs (such as times series, frames, etc.) and, in practice, they do not require an exponential increase of data when adding extra dimensions to the state or action space (see Chapter \ref{ch:intro_DL}).
\item In addition, they can be trained incrementally and make use of additional samples obtained as learning happens.
\end{itemize}

\section{Different settings to learn a policy from data}
\label{sec:approach_from_data}
We now describe key settings that can be tackled with RL.

\subsection{Offline and online learning}
Learning a sequential decision-making task appears in two cases: (i)~in the offline learning case where only limited data on a given environment is available and (ii)~in an online learning case where, in parallel to learning, the agent gradually gathers experience in the environment.
In both cases, the core learning algorithms introduced in Chapters~\ref{ch:value-based_methods} to \ref{ch:model-based} are essentially the same.
The specificity of the batch setting is that the agent has to learn from limited data without the possibility of interacting further with the environment.
In that case, the idea of generalization introduced in Chapter~\ref{ch:generalization} is the main focus.
In the online setting, the learning problem is more intricate and learning without requiring a large amount of data (sample efficiency) is not only influenced by the capability of the learning algorithm to generalize well from the limited experience.
Indeed, the agent has the possibility to gather experience via an \textit{exploration/exploitation strategy}.
In addition, it can use a \textit{replay memory} to store its experience so that it can be reprocessed at a later time.
Both the exploration and the replay memory will be discussed in Chapter~\ref{ch:challenges_online}).
In both the batch and the online settings, a supplementary consideration is also the computational efficiency, which, among other things, depends on the efficiency of a given gradient descent step.
All these elements will be introduced with more details in the following chapters.
A general schema of the different elements that can be found in most deep RL algorithms is provided in Figure~\ref{general_schema}.


\pgfdeclarelayer{background}
\pgfdeclarelayer{foreground}
\pgfsetlayers{background,main,foreground}

\tikzstyle{block}=[draw, fill=blue!20, text width=8.5em,
    text centered, minimum height=2.5em, rounded corners=2pt]
\def\blockdist{4}
\def\edgedist{4}

\begin{figure}[!ht]
\centering
\begin{tikzpicture}[thick,scale=0.8, every node/.style={scale=0.8}]
    \node (pol) [block, text width=9.2em]  {
          {Policies\\
                            \scriptsize
                        \begin{tabular}{l}
                    \hline
                    \text{Exploration/Exploitation}\\
                    \text{dilemma} \\
                        \end{tabular}}
    };
    \path (pol.north)+(0,0.5*\blockdist) node (contr) [block] {
      {Controllers\\
                            \scriptsize
                        \begin{tabular}{l}
                    \hline
                    \textbullet\ train/validation\\
                                and test phases\\
                    \textbullet\ hyperparameters\\
                        management\\
                        \end{tabular}}
    };
    \path (pol.west)+(-1*\blockdist,0) node (rm) [block] {
      {Replay memory\\
                            \scriptsize
                        \begin{tabular}{l}
                        \end{tabular}}
    };
    \path (contr.west)+(-\blockdist,0) node (learning) [block] {
      {Learning \\algorithms\\
                            \scriptsize
                    \begin{tabular}{l}
                    \hline
Value-based RL\\
Policy-based RL\\
Model-based RL\\
                    \end{tabular}}
    };

    \path (learning.west)+(-\blockdist*0.6,0) node (fa) [block] {
      {Function \\ Approximators \\
                            \scriptsize
                    \begin{tabular}{l}
                    \hline
                      \textbullet\ Convolutions \\
                      \textbullet\ Recurrent cells \\
      \textbullet\ ...
                    \end{tabular}}
    };

    \path (pol.south)+(-1*\blockdist,-1*\blockdist) node (env) [text width=8.5em,
    text centered, fill=yellow!20, rounded corners, draw=black!50, dashed, minimum height=10em] {
          {ENVIRONMENT \\
  }
    };

    \path (pol.south west)+(-2*\blockdist,-0.4) node (AGENT) {AGENT};
    \draw [->] (pol) -- (env.north east);
    \draw [-] (contr) -- (pol);
    \draw [-] (learning) -- (fa);
    \draw [-] (learning) -- node [above] {} (contr);
    \draw [->] (learning) -- (pol.north west);
    \draw [dashed,<-] (rm.105) -- (learning.-97);
    \draw [->] (rm.75) -- (learning.-83);
    \draw [->] (env.100) -- (rm);

    \begin{pgfonlayer}{background}
        \path (fa.west |- learning.north)+(-0.5,0.3) node (a) {};
        \path (AGENT.south -| pol.east)+(+0.3,-0.2) node (b) {};
        \path[fill=yellow!20,rounded corners, draw=black!50, dashed]
            (a) rectangle (b);
        \path (rm.north west)+(-0.2,0.2) node (a) {};
    \end{pgfonlayer}
\end{tikzpicture}
\caption{General schema of deep RL methods.}
\label{general_schema}
\end{figure}
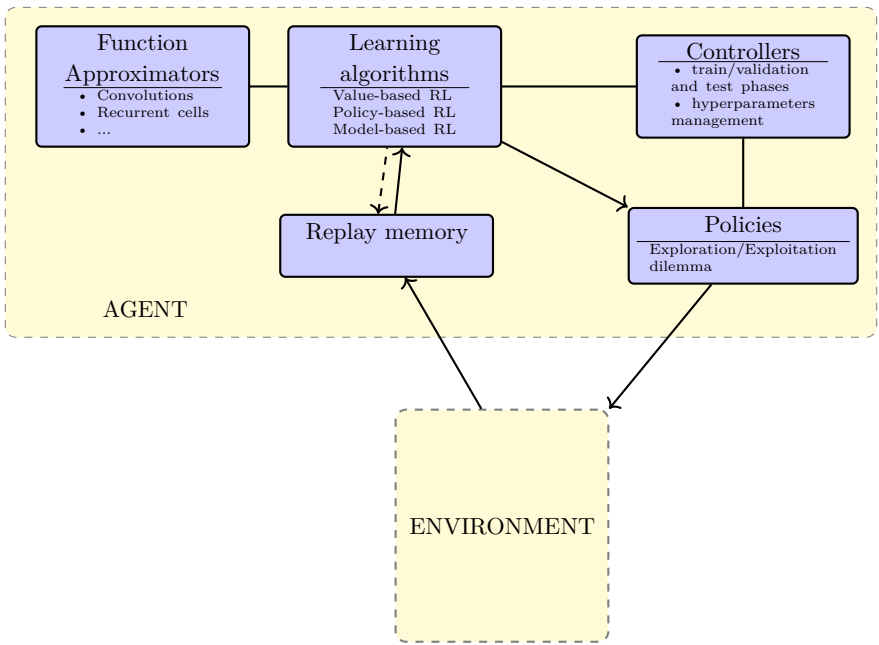

\subsection{Off-policy and on-policy learning}
According to \cite{sutton1998introduction}, \guillemotleft~on-policy methods attempt to evaluate or improve the policy that is used to make decisions, whereas off-policy methods evaluate or improve a policy different from that used to generate the data~\guillemotright.
In off-policy based methods, learning is straightforward when using trajectories that are not necessarily obtained under the current policy, but from a different behavior policy $\beta(s,a)$.
In those cases, experience replay allows reusing samples from a different behavior policy.
On the contrary, on-policy based methods usually introduce a bias when used with a replay buffer as the trajectories are usually not obtained solely under the current policy $\pi$.
As will be discussed in the following chapters, this makes off-policy methods sample efficient as they are able to make use of any experience; in contrast, on-policy methods would, if specific care is not taken, introduce a bias when using off-policy trajectories. 

\chapter{Value-based methods for deep RL}
\label{ch:value-based_methods}
The value-based class of algorithms aims to build a value function, which subsequently lets us define a policy.
We discuss hereafter one of the simplest and most popular value-based algorithms, the Q-learning algorithm \citep{watkins1989learning} and its variant, the fitted Q-learning, that uses parameterized function approximators \citep{gordon1996stable}.
We also specifically discuss the main elements of the deep Q-network (DQN) algorithm \citep{mnih2015human} which has achieved superhuman-level control when playing ATARI games from the pixels by using neural networks as function approximators.
We then review various improvements of the DQN algorithm and provide resources for further details.
At the end of this chapter and in the next chapter, we discuss the intimate link between value-based methods and policy-based methods.

\section{Q-learning}
\label{sec:q-learning}

The basic version of Q-learning keeps a lookup table of values $Q(s, a)$ (Equation \ref{def_Q}) with one entry for every state-action pair.
In order to learn the optimal Q-value function, the Q-learning algorithm makes use of the Bellman equation for the Q-value function \citep{bellman1962applied} whose unique solution is $Q^*(s, a)$:
\begin{equation}
Q^*(s, a) = (\mathcal B Q^*)(s, a),
\label{Bellman_Q}
\end{equation}
where $\mathcal B$ is the Bellman operator mapping any function $K:\mathcal S \times \mathcal A \rightarrow \mathbb{R}$ into another function $\mathcal S \times \mathcal A \rightarrow \mathbb{R}$ and is defined as follows:
\begin{equation}
(\mathcal BK)(s,a) = \sum_{s' \in S} T(s,a,s') \left( R(s,a,s')+\gamma \operatorname*{max}_{a' \in \mathcal A} K(s',a') \right).
\label{op_mapping_Bellman_Q}
\end{equation}
By Banach's theorem, the fixed point of the Bellman operator $\mathcal B$ exists since it is a contraction mapping\footnote{The Bellman operator is a contraction mapping because it can be shown that for any pair of bounded functions $K, K': S \times A \rightarrow \mathbb{R}$, the following condition is respected:
$$\lVert TK-TK'\lVert_\infty \le \gamma \lVert K-K'\lVert_\infty.$$}.
In practice, one general proof of convergence to the optimal value function is available \citep{watkins1992q} under the conditions that:
\begin{itemize}
\item the state-action pairs are represented discretely, and
\item all actions are repeatedly sampled in all states (which ensures sufficient exploration, hence not requiring access to the transition model).
\end{itemize}

This simple setting is often inapplicable due to the high-dimensional (possibly continuous) state-action space.
In that context, a parameterized value function $Q(s, a; \theta)$\label{ntn:Q_net} is needed, where $\theta$ refers to some parameters that define the Q-values.

\section{Fitted Q-learning}
Experiences are gathered in a given dataset $D$ in the form of tuples $<s,a,r,s'>$ where the state at the next time-step $s'$\label{ntn:next_state} is drawn from $T(s,a,\cdot)$ and the reward $r$ is given by $R(s,a,s')$.
In fitted Q-learning \citep{gordon1996stable}, the algorithm starts with some random initialization of the Q-values $Q(s, a; \theta_0 )$ where $\theta_0$ refers to the initial parameters (usually such that the initial Q-values should be relatively close to 0 so as to avoid slow learning). Then, an approximation of the Q-values at the $k^{th}$ iteration $Q(s, a; \theta_k )$ is updated towards the target value
\begin{equation}
Y_k^Q = r +\gamma \operatorname*{max}_{a' \in \mathcal A} Q(s',a';\theta_{k}),
\label{eq:target_Y}
\end{equation}
where $\theta_k$ refers to some parameters that define the Q-values at the $k^{th}$ iteration. 

In neural fitted Q-learning (NFQ) \citep{riedmiller2005neural}, the state can be provided as an input to the Q-network and a different output is given for each of the possible actions.
This provides an efficient structure that has the advantage of obtaining the computation of $\operatorname*{max}_{a' \in A} Q(s',a';\theta_k)$ in a single forward pass in the neural network for a given $s'$.
The Q-values are parameterized with a neural network $Q(s,a;\theta_k)$ where the parameters $\theta_k$ are updated by stochastic gradient descent (or a variant) by minimizing the square loss:
\begin{equation}
\text{L}_{DQN} = \left( Q(s, a; \theta_k )-Y_k^Q \right)^2.
\label{eq:loss_DQN}
\end{equation}
Thus, the Q-learning update amounts in updating the parameters:
\begin{equation}
\theta_{k+1}=\theta_{k}+\alpha \left(Y_k^Q - Q (s,a; \theta_k)\right) \nabla_{\theta_k} Q(s ,a ; \theta_k),
\label{QlearningitNN}
\end{equation}
where $\alpha$\label{ntn:learning_rate} is a scalar step size called the learning rate.
Note that using the square loss is not arbitrary. Indeed, it ensures that $Q(s, a; \theta_k )$ should tend without bias to the expected value of the random variable $Y_k^Q$ \footnote{The minimum of $\mathbb E[(Z-c)^2]$ occurs when the constant $c$ equals the expected value of the random variable $Z$.}. 
Hence, it ensures that $Q(s, a; \theta_k )$ should tend to $Q^*(s,a)$ after many iterations in the hypothesis that the neural network is well-suited for the task and that the experience gathered in the dataset $D$ is sufficient (more details will be given in Chapter \ref{ch:generalization}).

When updating the weights, one also changes the target. Due to the generalization and extrapolation abilities of neural networks, this approach can build large errors at different places in the state-action space\footnote{Note that even fitted value iteration with linear regression can diverge \citep{boyan1995generalization}. However, this drawback does not happen when using linear function approximators that only have interpolation abilities such as kernel-based regressors (k-nearest neighbors, linear and multilinear interpolation, etc.) \citep{gordon1999approximate} or tree-based ensemble methods \citep{ernst2005tree}. However, these methods have not proved able to handle successfully high-dimensional inputs.}.
Therefore, the contraction mapping property of the Bellman operator in Equation \ref{op_mapping_Bellman_Q} is not enough to guarantee convergence.
It is verified experimentally that these errors may propagate with this update rule and, as a consequence, convergence may be slow or even unstable \citep{baird1995residual,tsitsiklis1997analysis,gordon1999approximate,riedmiller2005neural}.
Another related damaging side-effect of using function approximators is the fact that Q-values tend to be overestimated due to the \textrm{max} operator \citep{van2016deep}. Because of the instabilities and the risk of overestimation, specific care has be taken to ensure proper learning.

\section{Deep Q-networks}

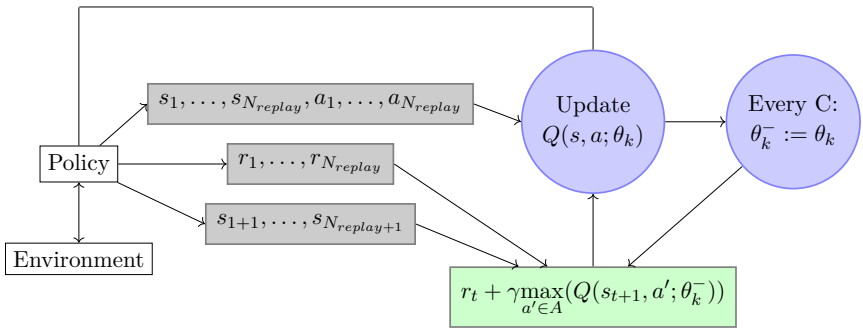
\begin{figure}[ht!]
 \centering
\scalebox{0.8}{
\begin{tikzpicture}
 [	NN/.style={circle,draw=blue!50,fill=blue!20,thick, inner sep=5pt,minimum size=6mm},
     inputs/.style={rectangle,draw=black!50,fill=black!20,thick, inner sep=5pt,minimum size=4mm},
     target/.style={rectangle,draw=black!50,fill=green!20,thick, inner sep=5pt,minimum size=4mm},
   Vhat/.style={circle,draw=black!50,fill=black!20,thick, inner sep=5pt,minimum size=4mm},
   ddt/.style={rectangle,draw=black!50,thick, inner sep=5pt,minimum size=4mm}];

  \node at (0,0) 	[NN, align=center] (NN) {Update \\$Q(s,a;\theta_k)$};
  \node [NN, right=of NN, align=center] (NN_t) {Every C:\\$\theta_k^{-}:=\theta_k$};
  \node [left=of NN,yshift=-7mm, xshift=-11mm,] 	[inputs] (rt) {$r_{1}, \ldots, r_{N_{replay}}$};
  \node [above=of rt, yshift=-7mm] 	[inputs] (st) {$s_{1}, \ldots, s_{N_{replay}}, a_{1}, \ldots, a_{N_{replay}}$};
  \node [below=of rt,yshift=7mm] 	[inputs] (st1) {$s_{1+1}, \ldots, s_{N_{replay+1}}$};
  \node [below=of NN,yshift=-2mm] 	[target] (target) {$r_t + \gamma \underset{a' \in A}{\operatorname{max}} ( Q(s_{t+1},a'; \theta_k^{-}) )$};
  \node[inner sep=0,minimum size=0,above=of NN, yshift=-3mm] (k) {}; 
  \node[draw,left=of rt, xshift=-8mm, align=center] (k3) {Policy}; 
  \node[draw,below=of k3, align=center] (env) {Environment}; 
  \node[inner sep=0,minimum size=0] (k1) at (k -| k3) {}; 
  \node[inner sep=0,minimum size=0] (k2) at (target -| k3) {}; 

 \draw [->] (st.east) -- (NN.west) node[midway, above] {};
 \draw [->] (st1.east) -- (target);
 \draw [->] (rt.east) -- (target);
 \draw [->] (target) -- (NN) node[midway, right, yshift=1mm, xshift=-1mm] {};
 \draw [-] (NN) -- (k);
 \draw [-] (k) -- (k1);
  \draw [-] (k1) -- (k3);
  \draw [->] (k3) -- (st.west);  \draw [->] (k3) -- (rt.west);  \draw [->] (k3) -- (st1.west);
  \draw [->] (NN) -- (NN_t);
  \draw [->] (NN_t) -- (target);
  \draw [<->] (k3) -- (env);

\end{tikzpicture}
}
 \caption[Sketch of the DQN algorithm.]{Sketch of the DQN algorithm. $Q(s,a;\theta_k)$ is initialized to random values (close to 0) everywhere in its domain and the replay memory is initially empty; the target Q-network parameters $\theta_k^-$ are only updated every C iterations with the Q-network parameters $\theta_k$ and are held fixed between updates; the update uses a mini-batch (e.g., 32 elements) of tuples $<s, a>$ taken randomly in the replay memory along with the corresponding mini-batch of target values for the tuples.}
 \label{ONFQ_schema}
\end{figure}

Leveraging ideas from NFQ, the deep Q-network (DQN) algorithm introduced by \citet{mnih2015human} is able to obtain strong performance in an online setting for a variety of ATARI games, directly by learning from the pixels. It uses two heuristics to limit the instabilities:
\begin{itemize}
\item The target Q-network in Equation \ref{eq:target_Y} is replaced by $Q(s',a';\theta_k^{-})$ where its parameters $\theta_k^{-}$ are updated only every $C \in \mathbb N$ iterations with the following assignment: $\theta_k^{-}=\theta_k$. This prevents the instabilities to propagate quickly and it reduces the risk of divergence as the target values $Y_k^Q$ are kept fixed for $C$ iterations. The idea of target networks can be seen as an instantiation of fitted Q-learning, where each period between target network updates corresponds to a single fitted Q-iteration.
\item In an online setting, the replay memory \citep{lin1992self} keeps all information for the last $N_{\text{replay}} \in \mathbb N$ time steps, where the experience is collected by following an $\epsilon$-greedy policy\footnote{It takes a random action with probability $\epsilon$ and follows the policy given by $\operatorname*{argmax}_{a \in \mathcal A} Q(s, a; \theta_k )$ with probability $1-\epsilon$.}.
The updates are then made on a set of tuples $<s,a,r,s'>$ (called mini-batch) selected randomly within the replay memory. This technique allows for updates that cover a wide range of the state-action space. In addition, one mini-batch update has less variance compared to a single tuple update. Consequently, it provides the possibility to make a larger update of the parameters, while having an efficient parallelization of the algorithm.
\end{itemize}
A sketch of the algorithm is given in Figure \ref{ONFQ_schema}. 

In addition to the target Q-network and the replay memory, DQN uses other important heuristics.
To keep the target values in a reasonable scale and to ensure proper learning in practice, rewards are clipped between -1 and +1.
Clipping the rewards limits the scale of the error derivatives and makes it easier to use the same learning rate across multiple games (however, it introduces a bias).
In games where the player has multiple lives, one trick is also to associate a terminal state to the loss of a life such that the agent avoids these terminal states (in a terminal state the discount factor is set to 0).

In DQN, many deep learning specific techniques are also used.
In particular, a preprocessing step of the inputs is used to reduce the input dimensionality, to normalize inputs (it scales pixels value into [-1,1]) and to deal with some specificities of the task.
In addition, convolutional layers are used for the first layers of the neural network function approximator and the optimization is performed using a variant of stochastic gradient descent called RMSprop \citep{rmsprop}.

\section{Double DQN}
\label{sec:double}
The max operation in Q-learning (Equations \ref{op_mapping_Bellman_Q}, \ref{eq:target_Y}) uses the same values both to select and to evaluate an action.
This makes it more likely to select overestimated values in case of inaccuracies or noise, resulting in overoptimistic value estimates.
Therefore, the DQN algorithm induces an upward bias.
The double estimator method uses two estimates for each variable, which allows for the selection of an estimator and its value to be uncoupled \citep{hasselt2010double}.
Thus, regardless of whether errors in the estimated Q-values are due to stochasticity in the environment, function approximation, non-stationarity, or any other source, this allows for the removal of the positive bias in estimating the action values.
In Double DQN, or DDQN \citep{van2016deep}, the target value $Y_k^Q$ is replaced by
\begin{equation}
Y_k^{DDQN} = r +\gamma Q(s',\operatorname*{argmax}_{a \in \mathcal A} Q(s',a;\theta_k);\theta_k^{-}),
\label{eq:target_YDDQN}
\end{equation}
which leads to less overestimation of the Q-learning values, as well as improved stability, hence improved performance.
As compared to DQN, the target network with weights $\theta_t^{-}$ are used for the evaluation of the current greedy action. Note that the policy is still chosen according to the values obtained by the current weights $\theta$.

\section{Dueling network architecture}
\label{sec:dueling}
In \citep{wang2015dueling}, the neural network architecture decouples the value and advantage function $A^{\pi}(s,a)$ (Equation \ref{def_A}), which leads to improved performance.
The Q-value function is given by
\begin{equation}
\begin{split}
Q(s,a;\theta^{(1)} & , \theta^{(2)},\theta^{(3)}) = V\left(s;\theta^{(1)},\theta^{(3)}\right) \\
& + \left( A\left(s,a;\theta^{(1)},\theta^{(2)}\right) - \operatorname*{max}_{a' \in \mathcal A} A\left(s,a';\theta^{(1)},\theta^{(2)}\right) \right).
\end{split}
\label{eq:QDueling}
\end{equation}

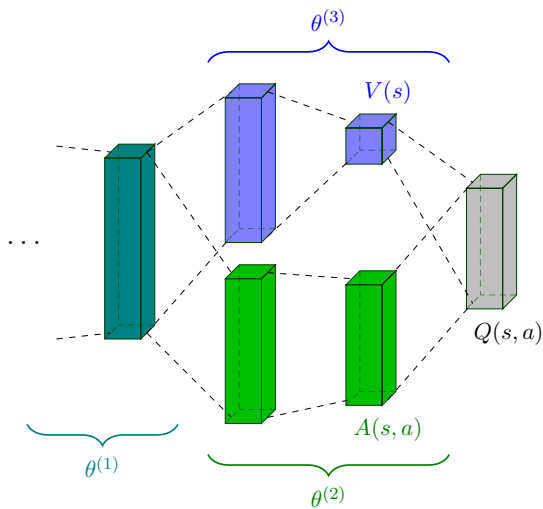
\begin{figure}[ht!]
 \centering
\scalebox{0.8}{
\begin{tikzpicture}

  \pic [fill=blue!50!green, draw=green!25!black] at (-4-0.1,0.5-0.1) {annotated cuboid={width=6, height=30, depth=6}};

  \pic [fill=blue!50, draw=green!25!black] at (-2-0.1,1.5-0.1) {annotated cuboid={width=6, height=24, depth=6}};
  \pic [fill=green!75!black, draw=green!25!black] at (-2-0.1,-1.5-0.1) {annotated cuboid={width=6, height=24, depth=6}};

  \pic [fill=blue!50, draw=green!25!black] at (0-0.1,1-0.1) {annotated cuboid={width=6, height=6, depth=6}};
  \pic [fill=green!75!black, draw=green!25!black] at (0-0.1,-1.6-0.1) {annotated cuboid={width=6, height=20, depth=6}};

  \pic [fill=black!25, draw=green!25!black] at (2-0.1,0-0.1) {annotated cuboid={width=6, height=20, depth=6}};

\node[font=\Large] (dots) at (-6,-1) {\ldots};

 \draw [dashed] node[] {} (-4-1.5,0.6) -- node[] {} (-4-0.6,0.5);
 \draw [dashed] node[] {} (-4-1.5,-2.6) -- node[] {} (-4-0.6,-2.5);

  \draw [dashed] node[] {} (-4,0.5) -- node[] {} (-2-0.6,1.5);
 \draw [dashed] node[] {} (-4,-2.5) -- node[] {} (-2-0.6,-0.9);
 \draw [dashed] node[] {} (-4,0.5) -- node[] {} (-2-0.6,-1.5);
 \draw [dashed] node[] {} (-4,-2.5) -- node[] {} (-2-0.6,-3.9);

\draw [dashed] node[] {} (-2,1.5) -- node[] {} (0-0.6,1);
 \draw [dashed] node[] {} (-2,-0.9) -- node[] {} (0-0.6,1-0.6);
 \draw [dashed] node[] {} (-2,-1.5) -- node[] {} (0-0.6,-1.6);
 \draw [dashed] node[] {} (-2,-3.9) -- node[] {} (0-0.6,-1.6-2);

 \draw [dashed] node[] {} (0,1) -- node[] {} (2-0.6,0);
 \draw [dashed] node[] {} (0,1-0.6) -- node[] {} (2-0.6,0-2);
 \draw [dashed] node[] {} (0,-1.6) -- node[] {} (2-0.6,0);
 \draw [dashed] node[] {} (0,-1.6-2) -- node[] {} (2-0.6,0-2);

\node[] (Q) at (2,-2.5) {$Q(s,a)$};
\node[color=green!50!black] (A) at (0,-4.1) {$A(s,a)$};
\node[color=blue] (V) at (0,1.5) {$V(s)$};

\draw [thick, blue,decorate,decoration={brace,amplitude=10pt},xshift=0.4pt,yshift=-0.4pt](-3,2.) -- (1,2.) node[blue,midway,yshift=0.7cm] {$\theta^{(3)}$};
\draw [thick, green!50!black,decorate,decoration={brace,amplitude=10pt,mirror},xshift=0.4pt,yshift=-0.4pt](-3,-4.5) -- (1,-4.5) node[green!50!black,midway,yshift=-0.7cm] {$\theta^{(2)}$};
\draw [thick, blue!50!green,decorate,decoration={brace,amplitude=10pt,mirror},xshift=0.4pt,yshift=-0.4pt](-6,-4) -- (-3.5,-4) node[blue!50!green,midway,yshift=-0.7cm] {$\theta^{(1)}$};

\end{tikzpicture}
}
 \caption{Illustration of the dueling network architecture with the two streams that separately estimate the value $V(s)$ and the advantages $A(s,a)$. The boxes represent layers of a neural network and the grey output implements equation \ref{eq:QDueling} to combine $V(s)$ and $A(s,a)$.}
 \label{fig:dueling}
\end{figure}

Now, for $a^* = \operatorname*{argmax}_{a' \in \mathcal A} Q(s, a'; \theta^{(1)},\theta^{(2)},\theta^{(3)})$, we obtain $Q(s, a^*;\theta^{(1)},\theta^{(2)},\theta^{(3)}) = V (s; \theta^{(1)},\theta^{(3)})$. As illustrated in Figure \ref{fig:dueling}, the stream $V (s; \theta^{(1)},\theta^{(3)})$ provides an estimate of the value function, while the other stream produces an estimate of the advantage function. The learning update is done as in DQN and it is only the structure of the neural network that is modified.

In fact, even though it loses the original semantics of $V$ and $A$, a slightly different approach is preferred in practice because it increases the stability of the optimization:
\begin{equation}
\begin{split}
Q(s,a &; \theta^{(1)} ,\theta^{(2)} ,\theta^{(3)}) = V\left(s;\theta^{(1)},\theta^{(3)}\right)\\
& + \left( A\left(s,a;\theta^{(1)},\theta^{(2)}\right) - \frac{1}{\lvert \mathcal A \rvert} \sum_{a' \in \mathcal A} A \left(s,a';\theta^{(1)},\theta^{(2)}\right) \right).
\end{split}
\label{eq:QDueling2}
\end{equation}
In that case, the advantages only need to change as fast as the mean, which appears to work better in practice \citep{wang2015dueling}.

\section{Distributional DQN}
\label{sec:distrib}
The approaches described so far in this chapter all directly approximate the expected return in a value function.
Another approach is to aim for a richer representation through a value distribution, i.e. the distribution of possible cumulative returns \citep{jaquette1973markov, morimura2010nonparametric}.
This value distribution provides more complete information of the intrinsic randomness of the rewards and transitions of the agent within its environment (note that it is not a measure of the agent's uncertainty about the environment).

The value distribution $Z^\pi$ is a mapping from state-action pairs to distributions of returns when following policy $\pi$. It has an expectation equal to $Q^\pi$:
$$Q^\pi(s,a) = \mathbb E Z^\pi(s,a).$$
This random return is also described by a recursive equation, but one of a distributional nature:
\begin{equation}
Z^\pi(s,a)=R(s,a,S')+\gamma Z^\pi(S',A'),
\label{eq:distribQit}
\end{equation}
where we use capital letters to emphasize the random nature of the next state-action pair $(S',A')$ and $A' \sim \pi(\cdot|S')$.
The distributional Bellman equation states that the distribution of $Z$ is characterized by the interaction of three random variables: the reward $R(s,a,S')$, the next state-action $(S', A')$, and its random return $Z^\pi(S',A')$.

\begin{figure}[ht!]
\centering
\subfloat[Example MDP.]{ 
\begin{tikzpicture}[->, >=stealth', auto, semithick, node distance=3cm, baseline]
\tikzstyle{state}=[fill=white, draw=black, thick, text=black, scale=1]
\node[state]    (A)      at (0,1.5)               {$(s,a)$};
\node    (A1)      [color=green!75!black]  at (0,2)            {$\pi_1$};
\node    (A2)      [color=cyan]   at (0,1)            {$\pi_2$};
\node[state]    (B)[above right of=A, yshift=-1cm]   {$s^{(1)}$};
\node[state]    (C)[below right of=A, yshift=1cm]   {$s^{(2)}$};
\node[state]    (D) at (2.5,1.5)   {$s^{(3)}$};
\node[]    (dummy)      at (4.5,-0.7)               {};
\path
(A) edge[bend left, color=green!75!black]  node[above, yshift=0.2cm] {$\mathbb P=1$}      (B)
    edge[color=cyan]  node[above, yshift=0.2cm] {$\mathbb P=0.2$}     (D)
    edge[bend right, color=cyan] node[below, yshift=-0.2cm] {$\mathbb P=0.8$}      (C)
(B) edge[loop right]  node {$\frac{R_{\text{max}}}{5}$}   (B)
(C) edge[loop right]  node {$0$}   (C)
(D) edge[loop right]  node {$R_{\text{max}}$}   (D)
;
\end{tikzpicture}
}
\subfloat[Sketch (in an idealized version) of the estimate of resulting value distribution $\hat Z^{\pi_1}$ and $\hat Z^{\pi_2}$ as well as the estimate of the Q-values $\hat Q^{\pi_1}, \hat Q^{\pi_2}$.]{
\begin{tikzpicture}
[	distrib/.style={rectangle,draw=black!50,fill=black!10,thick, inner sep=5pt,minimum size=4mm, align=center},
  task/.style={circle,draw=black!50,fill=black!5,thick, inner sep=5pt,minimum size=25mm, align=center},
  RLalgo/.style={rectangle,draw=black!50,fill=black!10,thick, inner sep=10pt,minimum width=30mm, align=center},
    declare function={fct1(\x,\k,\theta) = 0.8*1/((2*3.14)^0.5*0.1)*exp(-(1/2)*((\x-0)/(0.1))^2);},
    declare function={fct2(\x,\k,\theta) = 0.2*1/((2*3.14)^0.5*0.1)*exp(-(1/2)*((\x-5)/(0.1))^2);},
    declare function={fct3(\x,\k,\theta) = 1*1/((2*3.14)^0.5*0.1)*exp(-(1/2)*((\x-1)/(0.1))^2);},
    baseline
];
\node[color=cyan]  (C) at (-0.3,3.5)   {$\hat Z^{\pi_1}$};
\node[color=green!75!black]  (C) at (-0.3,3.)   {$\hat Z^{\pi_2}$};

\begin{axis}[
  no markers, domain=-0.3:6, samples=100,
  axis lines=left, xlabel=, ylabel=,
  every axis y label/.style={at=(current axis.above origin),anchor=east},
  every axis x label/.style={at=(current axis.right of origin),anchor=north},
  height=5cm, width=6cm,
  xtick={0,5}, ytick=\empty,
  xticklabels={0,$\frac{R_{\text{max}}}{1-\gamma}$},
  xmax=6.5,
  ymax=5.
  ]
\addplot [very thick,cyan!20!black] {fct1(x,2,2)};
\addplot [fill=cyan!20, draw=none] {fct1(x,2,2)} \closedcycle;
\addplot [very thick,cyan!20!black] {fct2(x,2,2)};
\addplot [fill=cyan!20, draw=none] {fct2(x,2,2)} \closedcycle;
\addplot [very thick,green!20!black] {fct3(x,2,2)};
\addplot [fill=green!20, draw=none] {fct3(x,2,2)} \closedcycle;
\end{axis}
\draw[color=green!50!black, thick] (0.83,0) -- (0.83,3.5);
\draw[color=cyan, thick] (0.86,0) -- (0.86,3.5);

\node[color=green!50!black]  (C) at (0.55,3.8)   {$\hat Q^{\pi_1}$};
\node[] at (0.95,3.7)   {$\approx$};
\node[color=cyan]  (C) at (1.45,3.8)   {$\hat Q^{\pi_2}$};

\end{tikzpicture}
}

 \caption{For two policies illustrated on Fig (a), the illustration on Fig (b) gives the value distribution $Z^{(\pi)}(s,a)$ as compared to the expected value $Q^\pi(s,a)$. On the left figure, one can see that $\pi_1$ moves with certainty to an absorbing state with reward at every step $\frac{R_{\text{max}}}{5}$, while $\pi_2$ moves with probability $0.2$ and $0.8$ to absorbing states with respectively rewards at every step $R_{\text{max}}$ and $0$. From the pair $(s,a)$, the policies $\pi_1$ and $\pi_2$ have the same expected return but different value distributions.}
 \label{fig:distrib}
\end{figure}
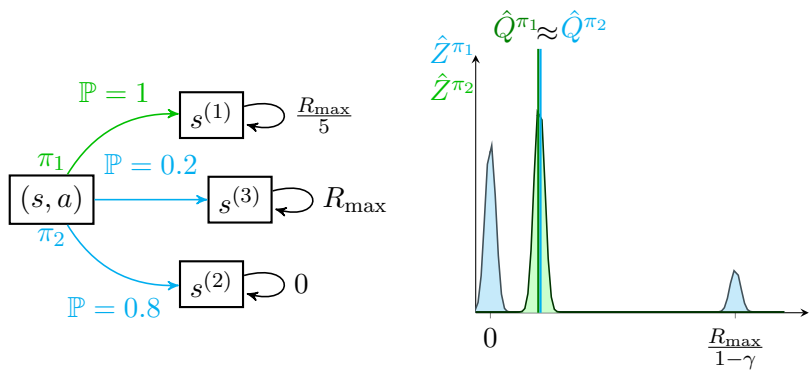

It has been shown that such a distributional Bellman equation can be used in practice, with deep learning as the function approximator \citep{bellemare2017distributional, dabney2017distributional,rowland2018analysis}.
This approach has the following advantages:
\begin{itemize}
\item It is possible to implement risk-aware behavior (see e.g., \cite{morimura2010nonparametric}).
\item It leads to more performant learning in practice. This may appear surprising since both DQN and the distributional DQN aim to maximize the expected return (as illustrated in Figure \ref{fig:distrib}). One of the main elements is that the distributional perspective naturally provides a richer set of training signals than a scalar value function $Q(s,a)$. These training signals that are not a priori necessary for optimizing the expected return are known as \textit{auxiliary tasks} \citep{jaderberg2016reinforcement} and lead to an improved learning (this is discussed in \S\ref{sec:aux_tasks}).
\end{itemize}

\section{Multi-step learning}
\label{sec:multi_step}
In DQN, the target value used to update the Q-network parameters (given in Equation \ref{eq:target_Y}) is estimated as the sum of the immediate reward and a contribution of the following steps in the return. That contribution is estimated based on its own value estimate at the next time-step. For that reason, the learning algorithm is said to \textit{bootstrap} as it recursively uses its own value estimates \citep{sutton1988learning}.

This method of estimating a target value is not the only possibility.
Non-bootstrapping methods learn directly from returns (Monte Carlo) and an intermediate solution is to use a multi-step target \citep{sutton1988learning,watkins1989learning,peng1994incremental,singh1996reinforcement}.
Such a variant in the case of DQN can be obtained by using the n-step target value given by:
\begin{equation}
Y_k^{Q,n} = \sum_{t=0}^{n-1} \gamma^t r_{t} +\gamma^{n} \operatorname*{max}_{a' \in A} Q(s_{n},a';\theta_k)
\label{eq:target_Y_multi}
\end{equation}
where $(s_0, a_0, r_0,\cdots, s_{n-1}, a_{n-1}, r_{n-1},s_{n})$ is any trajectory of $n+1$ time steps with $s=s_0$ and $a=a_0$.
A combination of different multi-steps targets can also be used:
\begin{equation}
Y_k^{Q,n} = \sum_{i=0}^{n-1} \lambda_i \left( \sum_{t=0}^{i} \gamma^t r_{t} +\gamma^{i+1} \operatorname*{max}_{a' \in A} Q\left(s_{i+1},a';\theta_k\right) \right)
\label{eq:target_Y_combi_multi}
\end{equation}
with $\sum_{i=0}^{n-1} \lambda_i=1$. In the method called $TD(\lambda)$ \citep{sutton1988learning}, $n \rightarrow \infty$ and $\lambda_i$ follow a geometric law: $\lambda_i \propto \lambda^i$ where $0 \le \lambda \le 1$.

\paragraph{To bootstrap or not to bootstrap?}

Bootstrapping has both advantages and disadvantages. On the negative side, using pure bootstrapping methods (such as in DQN) are prone to instabilities when combined with function approximation because they make recursive use of their own value estimate at the next time-step. On the contrary, methods such as n-step Q-learning rely less on their own value estimate because the estimate used is decayed by $\gamma^n$ for the n$^{th}$ step backup. In addition, methods that rely less on bootstrapping can propagate information more quickly from delayed rewards as they learn directly from returns \citep{sutton1996generalization}. Hence they might be more computationally efficient.

Bootstrapping also has advantages. The main advantage is that using value bootstrap allows learning from off-policy samples.
Indeed, methods that do not use pure bootstrapping, such as n-step Q-learning with $n>1$ or $TD(\lambda)$, are in principle on-policy based methods that would introduce a bias when used with trajectories that are not obtained solely under the behavior policy $\mu$ (e.g., stored in a replay buffer).

The conditions required to learn efficiently and safely with eligibility traces from off-policy experience are provided by \cite{munos2016safe,harutyunyan2016}.
In the control setting, the retrace operator \citep{munos2016safe} considers a sequence of target policies $\pi$ that depend on the sequence of Q-functions (such as $\epsilon$-greedy policies), and seek to approximate $Q^*$ (if $\pi$ is greedy or becomes increasingly greedy w.r.t. the Q estimates).
It leads to the following target:
\begin{small}
\begin{equation}
Y = Q(s,a) + \left[ \sum_{t \ge 0} \gamma^t \left( \prod_{c_s=1}^{t} c_s \right) \left(r_t + \gamma \mathbb E_\pi Q(s_{t+1},a') - Q(s_t,a_t)\right) \right]
\end{equation}
\end{small}%
where $c_s=\lambda \min \left(1,\frac{\pi(s,a)}{\mu(s,a)}\right)$ with $0 \le \lambda \le 1$ and $\mu$ is the behavior policy (estimated from observed samples). This way of updating the Q-network has guaranteed convergence, does not suffer from a high variance and it does not cut the traces unnecessarily when $\pi$ and $\mu$ are close. Nonetheless, one can note that estimating the target is more expansive to compute as compared to the one-step target (such as in DQN) because the Q-value function has to be estimated on more states.

\section{Combination of all DQN improvements and variants of DQN}

The original DQN algorithm can combine the different variants discussed in \S \ref{sec:double} to \S \ref{sec:multi_step} (as well as some discussed in Chapter \ref{sec:explo-exploit}) and that has been studied by \cite{hessel2017rainbow}. Their experiments show that the combination of all the previously mentioned extensions to DQN provides state-of-the-art performance on the Atari 2600 benchmarks, both in terms of sample efficiency and final performance. Overall, a large majority of Atari games can be solved such that the deep RL agents surpass the human level performance. 

Some limitations remain with DQN-based approaches. Among others, these types of algorithms are not well-suited to deal with large and/or continuous action spaces. In addition, they cannot explicitly learn stochastic policies. 
Modifications that address these limitations will be discussed in the following Chapter \ref{ch:policy-based_methods}, where we discuss policy-based approaches.
Actually, the next section will also show that value-based and policy-based approaches can be seen as two facets of the same model-free approach. Therefore, the limitations of discrete action spaces and deterministic policies are only related to DQN.

One can also note that value-based or policy-based approaches do not make use of any model of the environment, which limits their sample efficiency. 
Ways to combine model-free and model-based approaches will be discussed in Chapter \ref{ch:model-based}.

\chapter{Policy gradient methods for deep RL}
\label{ch:policy-based_methods}

This section focuses on a particular family of reinforcement learning algorithms that use policy gradient methods. These methods optimize a performance objective (typically the expected cumulative reward) by finding a good policy (e.g a neural network parameterized policy) thanks to variants of stochastic gradient ascent with respect to the policy parameters.
Note that policy gradient methods belong to a broader class of policy-based methods that includes, among others, evolution strategies.
These methods use a learning signal derived from sampling instantiations of policy parameters and the set of policies is developed towards policies that achieve better returns \citep[e.g.,][]{salimans2017evolution}.

In this chapter, we introduce the stochastic and deterministic gradient theorems that provide gradients on the policy parameters in order to optimize the performance objective. Then, we present different RL algorithms that make use of these theorems.

\section{Stochastic Policy Gradient}
\label{sec:SPG}
The expected return of a stochastic policy $\pi$ starting from a given state $s_0$ from Equation \ref{def_V} can be written as \citep{sutton2000policy}: 
\begin{equation}
\begin{split}
V^{\pi}(s_0) & = \int_{\mathcal{S}} \rho^\pi(s) \int_{\mathcal{A}} \pi(s,a) R'(s,a) da ds,
 \end{split}
\end{equation}
where $R'(s,a)=\int_{s' \in \mathcal S} T(s,a,s') R(s,a,s')$\label{ntn:rwd_fct_prime} and $\rho^\pi(s)$ is the discounted state distribution defined as
$$\rho^\pi(s)=\sum_{t=0}^{\infty} \gamma^t Pr\{s_t=s | s_0, \pi \}.$$\label{ntn:disc_st_dis}

For a differentiable policy $\pi_w$\label{ntn:diff_pol}, the fundamental result underlying these algorithms is the policy gradient theorem \citep{sutton2000policy}:

\begin{equation}
\begin{split}
\nabla_w V^{\pi_w}(s_0) & = \int_{\mathcal{S}} \rho^{\pi_w}(s) \int_{\mathcal{A}} \nabla_w  \pi_w(s,a) Q^{\pi_w}(s,a) da ds. \\
 \end{split}
 \label{eq:po_grad}
\end{equation}
This result allows us to adapt the policy parameters $w$: $\Delta w \propto \nabla_w V^{\pi_w}(s_0)$ from experience.
This result is particularly interesting since the policy gradient does not depend on the gradient of the state distribution (even though one might have expected it to).
The simplest way to derive the policy gradient estimator (i.e., estimating $\nabla_w V^{\pi_w}(s_0)$ from experience) is to use a \textit{score function gradient estimator}, commonly known as the REINFORCE algorithm \citep{williams1992simple}. The likelihood ratio trick can be exploited as follows to derive a general method of estimating gradients from expectations:

\begin{equation}
\label{likelihood-ratio-trick}
\begin{split}
\nabla_w \pi_w(s,a) & = \pi_w(s,a)\frac{\nabla_w \pi_w(s,a)}{\pi_w(s,a)} \\
 & = \pi_w(s,a) \nabla_w \log(\pi_w(s,a)).
\end{split}
\end{equation}
Considering Equation \ref{likelihood-ratio-trick}, it follows that

\begin{equation}
\begin{split}
\nabla_w V^{\pi_w}(s_0) & = \mathbb{E}_{s \sim \rho^{\pi_w}, a \sim \pi_w} \left[ \nabla_w \left(\log~\pi_w (s,a) \right)Q^{\pi_w}\left(s,a\right) \right].
 \end{split}
 \label{eq:score function gradient estimator}
\end{equation}
Note that, in practice, most policy gradient methods effectively use undiscounted state distributions, without hurting their performance \citep{thomas2014bias}.

So far, we have shown that policy gradient methods should include a policy evaluation followed by a policy improvement.
On the one hand, the policy evaluation estimates $Q^{\pi_{w}}$.
On the other hand, the policy improvement takes a gradient step to optimize the policy $\pi_{w}(s,a)$ with respect to the value function estimation.
Intuitively, the policy improvement step increases the probability of the actions proportionally to their expected return.

The question that remains is how the agent can perform the policy evaluation step, i.e., how to obtain an estimate of $Q^{\pi_w}(s,a)$.
The simplest approach to estimating gradients is to replace the Q function estimator with a cumulative return from entire trajectories.
In the Monte-Carlo policy gradient, we estimate the $Q^{\pi_w}(s,a)$ from rollouts on the environment while following policy $\pi_{w}$.
The Monte-Carlo estimator is an unbiased well-behaved estimate when used in conjunction with the back-propagation of a neural network policy, as it estimates returns until the end of the trajectories (without instabilities induced by bootstrapping).
However, the main drawback is that the estimate requires on-policy rollouts and can exhibit high variance. Several rollouts are typically needed to obtain a good estimate of the return. 
A more efficient approach is to instead use an estimate of the return given by a value-based approach, as in actor-critic methods discussed in \S \ref{sec:actor-critic}.

We make two additional remarks.
First, to prevent the policy from becoming deterministic, it is common to add an entropy regularizer to the gradient.
With this regularizer, the learnt policy can remain stochastic.
This ensures that the policy keeps exploring.

Second, instead of using the value function $Q^{\pi_w}$ in Eq. \ref{eq:score function gradient estimator}, an advantage value function $A^{\pi_{w}}$ can also be used. While $Q^{\pi_{w}}(s,a)$ summarizes the performance of each action for a given state under policy $\pi_{w}$, the advantage function $A^{\pi_{w}}(s,a)$ provides a measure of comparison for each action to the expected return at the state $s$, given by $V^{\pi_{w}}(s)$.
Using $A^{\pi_{w}}(s,a)=Q^{\pi_{w}}(s,a)-V^{\pi_{w}}(s)$ has usually lower magnitudes than $Q^{\pi_{w}}(s,a)$.
This helps reduce the variance of the gradient estimator $\nabla_w V^{\pi_w}(s_0)$ in the policy improvement step, while not modifying the expectation\footnote{Indeed, subtracting a baseline that only depends on $s$ to $Q^{\pi_{w}}(s,a)$ in Eq. \ref{eq:po_grad} does not change the gradient estimator because $\forall s$, $\int_{\mathcal{A}} \nabla_w  \pi_w(s,a) da=0$.}.
In other words, the value function $V^{\pi_{w}}(s)$ can be seen as a \textit{baseline} or \textit{control variate} for the gradient estimator. 
When updating the neural network that fits the policy, using such a baseline allows for improved numerical efficiency -- i.e. reaching a given performance with fewer updates -- because the learning rate can be bigger.

\section{Deterministic Policy Gradient}
\label{sec:DPG}
The policy gradient methods may be extended to deterministic policies.
The Neural Fitted Q Iteration with Continuous Actions (NFQCA) \citep{hafner2011reinforcement} and the Deep Deterministic Policy Gradient (DDPG) \citep{silver2014deterministic, lillicrap2015continuous} algorithms introduce the direct representation of a policy in such a way that it can extend the NFQ and DQN algorithms to overcome the restriction of discrete actions.

Let us denote by $\pi(s)$ the deterministic policy: $\pi(s): \mathcal S~\rightarrow~\mathcal A$.
In discrete action spaces, a direct approach is to build the policy iteratively with:
\begin{equation}
\pi_{k+1}(s) = \underset{a \in \mathcal A}{\operatorname{argmax}}~Q^{\pi_k} (s, a),
\end{equation}
where $\pi_{k}$\label{ntn:pol_kth} is the policy at the $k^{th}$ iteration.
In continuous action spaces, a greedy policy improvement becomes problematic, requiring a global maximisation at every step.
Instead, let us denote by $\pi_w(s)$ a differentiable deterministic policy. In that case, a simple and computationally attractive alternative is to move the policy in the direction of the gradient of Q, which leads to the Deep Deterministic Policy Gradient (DDPG) algorithm \citep{lillicrap2015continuous}:
\begin{equation}
\nabla_w  V^{\pi_w}(s_0) = \mathbb{E}_{s \sim \rho^{\pi_w}} \left[ \nabla_w \left(\pi_w \right) \nabla_a \left(Q^{\pi_w}(s,a)\right)|_{a=\pi_w(s)} \right].
\end{equation}
This equation implies relying on $\nabla_a \left(Q^{\pi_w}(s,a)\right)$ (in addition to $\nabla_w \pi_w$), which usually requires using actor-critic methods (see \S \ref{sec:actor-critic}).

\section{Actor-Critic Methods}
\label{sec:actor-critic}
As we have seen in \S \ref{sec:SPG} and \S \ref{sec:DPG}, a policy represented by a neural network can be updated by gradient ascent for both the deterministic and the stochastic case.
In both cases, the policy gradient typically requires an estimate of a value function for the current policy.
One common approach is to use an actor-critic architecture that consists of two parts: an actor and a critic \citep{konda2000actor}.
The actor refers to the policy and the critic to the estimate of a value function (e.g., the Q-value function).
In deep RL, both the actor and the critic can be represented by non-linear neural network function approximators \citep{mnih2016asynchronous}.
The actor uses gradients derived from the policy gradient theorem and adjusts the policy parameters $w$.
The critic, parameterized by $\theta$, estimates the approximate value function for the current policy~$\pi$: $Q(s,a;\theta) \approx Q^{\pi}(s,a)$.


\subsubsection{The critic}
From a (set of) tuples $<s,a,r,s'>$, possibly taken from a replay memory, the simplest off-policy approach to estimating the critic is to use a pure bootstrapping algorithm $TD(0)$ where, at every iteration, the current value $Q(s,a;\theta)$ is updated towards a target value:
\begin{equation}
Y_k^Q = r + \gamma Q(s', a=\pi(s'); \theta)
\label{eq:TD_sarsa}
\end{equation}
This approach has the advantage of being simple, yet it is not computationally efficient as it uses a pure bootstrapping technique that is prone to instabilities and has a slow reward propagation backwards in time \citep{sutton1996generalization}.
This is similar to the elements discussed in the value-based methods in \S \ref{sec:multi_step}.

The ideal is to have an architecture that is
\begin{itemize}
\item sample-efficient such that it should be able to make use of both off-policy and and on-policy trajectories (i.e., it should be able to use a replay memory), and
\item computationally efficient: it should be able to profit from the stability and the fast reward propagation of on-policy methods for samples collected from near on-policy behavior policies.
\end{itemize}
There are many methods that combine on- and off-policy data for policy evaluation \citep{precup2000eligibility}. The algorithm $Retrace(\lambda)$ \citep{munos2016safe} has the advantages that (i)~it can make use of samples collected from any behavior policy without introducing a bias and (ii)~it is efficient as it makes the best use of samples collected from near on-policy behavior policies.
That approach was used in actor-critic architectures described by \citet{DBLP:journals/corr/WangBHMMKF16, gruslys2017reactor}.
These architectures are sample-efficient thanks to the use of a replay memory, and computationally efficient since they use multi-step returns which improves the stability of learning and increases the speed of reward propagation backwards in time.

\subsubsection{The actor}
From Equation \ref{eq:score function gradient estimator}, the off-policy gradient in the policy improvement phase for the stochastic case is given as:
\begin{equation}
\nabla_w V^{\pi_w}(s_0) = \mathbb{E}_{s \sim \rho^{\pi_\beta}, a \sim \pi_\beta} \left[ \nabla_\theta \left(\log~\pi_w (s,a) \right)Q^{\pi_w}\left(s,a\right) \right].
\end{equation}
where $\beta$ is a behavior policy generally different than $\pi$, which makes the gradient generally biased.
This approach usually behaves properly in practice but the use of a biased policy gradient estimator makes difficult the analysis of its convergence without the GLIE assumption \citep{munos2016safe,gruslys2017reactor}\footnote{Greedy in the Limit with Infinite Exploration (GLIE) means that the behavior policies are required to become greedy (no exploration) in the limit of an online learning setting where the agent has gathered an infinite amount of experience. It is required that
\guillemotleft~(i) each action is executed infinitely often in every state that is visited infinitely often, and (ii) in the limit, the learning policy is greedy with respect to the Q-value function with probability 1~\guillemotright \citep{singh2000convergence}.}.

In the case of actor-critic methods, an approach to perform the policy gradient on-policy without experience replay has been investigated with the use of asynchronous methods, where multiple agents are executed in parallel and the actor-learners are trained asynchronously \citep{mnih2016asynchronous}.
The parallelization of agents also ensures that each agent experiences different parts of the environment at a given time step. 
In that case, n-step returns can be used without introducing a bias.
This simple idea can be applied to any learning algorithm that requires on-policy data and it removes the need to maintain a replay buffer.
However, this asynchronous trick is not sample efficient.

An alternative is to combine off-policy and on-policy samples to trade-off both the sample efficiency of off-policy methods and the stability of on-policy gradient estimates.
For instance, Q-Prop \citep{DBLP:journals/corr/GuLGTL16} uses a Monte Carlo on-policy gradient estimator, while reducing the variance of the gradient estimator by using an off-policy critic as a control variate.
One limitation of Q-Prop is that it requires using on-policy samples for estimating the policy gradient.

\section{Natural Policy Gradients}
Natural policy gradients are inspired by the idea of natural gradients for the updates of the policy.
Natural gradients can be traced back to the work of \cite{DBLP:journals/neco/Amari98} and has been later adapted to reinforcement learning \citep{DBLP:conf/nips/Kakade01}.

Natural policy gradient methods use the steepest direction given by the Fisher information metric, which uses the manifold of the objective function.
In the simplest form of steepest ascent for an objective function $J(w)$, the update is of the form $\bigtriangleup w \propto \nabla_{w} J(w)$. In other words, the update follows the direction that maximizes $\left(J(w) - J(w+\bigtriangleup w)\right)$ under a constraint on $|| \bigtriangleup w ||_2$.
In the hypothesis that the constraint on $\bigtriangleup w$ is defined with another metric than $L_2$, the first-order solution to the constrained optimization problem typically has the form $\bigtriangleup w \propto B^{-1} \nabla_{w} J(w)$ where $B$ is an $n_w \times n_w$ matrix.
In natural gradients, the norm uses the Fisher information metric, given by a local quadratic approximation to the KL divergence $D_{KL}(\pi^{w}||\pi^{w+\Delta w})$.
The natural gradient ascent for improving the policy $\pi_{w}$ is given by
\begin{equation}
    \bigtriangleup w \propto F_{w}^{-1} \nabla_w V^{\pi_w}(\cdot),
\end{equation}
where $F_{w}$ is the Fisher information matrix given by
\begin{equation}
    F_{w} = \E_{\pi_{w}} [\nabla_{w} \log \pi_{w} (s,\cdot) (\nabla_{w} \log \pi_{w}(s,\cdot))^T ].
\end{equation}
Policy gradients following $\nabla_{w} V^{\pi_w}(\cdot)$ are often slow because they are prone to getting stuck in local plateaus. 
Natural gradients, however, do not follow the usual steepest direction in the parameter space, but the steepest direction with respect to the Fisher metric.
Note that, as the angle between natural and ordinary gradient is never larger than ninety degrees, convergence is also guaranteed when using natural gradients.

The caveat with natural gradients is that, in the case of neural networks and their large number of parameters, it is usually impractical to compute, invert, and store the Fisher information matrix \citep{schulman2015trust}.
This is the reason why natural policy gradients are usually not used in practice for deep RL;
however alternatives inspired by this idea have been found and they are discussed in the following section.

\section{Trust Region Optimization}
As a modification to the natural gradient method, policy optimization methods based on a \textit{trust region} aim at improving the policy while changing it in a controlled way.
These constraint-based policy optimization methods focus on restricting the changes in a policy using the KL divergence between the action distributions.
By bounding the size of the policy update, trust region methods also bound the changes in state distributions guaranteeing improvements in policy.

TRPO \citep{schulman2015trust} uses constrained updates and advantage function estimation to perform the update, resulting in the reformulated optimization given by 
\begin{equation}
\max_{\bigtriangleup w} \E_{s \sim \rho^{\pi_w},a \sim \pi_w} \left[  \frac{\pi_{w+\bigtriangleup w} (s,a)}{ \pi_{w} (s,a)} A^{\pi_w}(s,a) \right]
\end{equation}
subject to $\E D_\text{KL} \left( \pi_{w} (s, \cdot) || \pi_{w+\bigtriangleup w}(s,\cdot) \right) \leq \delta$, where $\delta \in \mathbb R$ is a hyperparameter.
From empirical data, TRPO uses a conjugate gradient with KL constraint to optimize the objective function.

Proximal Policy Optimization (PPO) \citep{schulman2017proximal} is a variant of the TRPO algorithm, which formulates the constraint as a penalty or a clipping objective, instead of using the KL constraint.
Unlike TRPO, PPO considers modifying the objective function to penalize changes to the policy that move $r_t(w)=\frac{\pi_{w+\bigtriangleup w} (s,a)}{ \pi_{w} (s,a)}$ away from 1.
The clipping objective that PPO maximizes is given by
\begin{equation}
\underset{s \sim \rho^{\pi_w}, a \sim \pi_w}{\E} \bigg[ \min \Big( r_t(w) A^{\pi_w}(s,a), \text{clip} \big(  r_t(w), 1-\epsilon, 1 + \epsilon \big) A^{\pi_w}(s,a)  \Big) \bigg]
\end{equation}
where $\epsilon \in \mathbb R$ is a hyperparameter. This objective function clips the probability ratio to constrain the changes of $r_t$ in the interval $ \left[ 1-\epsilon, 1+\epsilon  \right]  $.



\section{Combining policy gradient and Q-learning}
\label{sec:comb_pol_Q}
Policy gradient is an efficient technique for improving a policy in a reinforcement learning setting.
As we have seen, this typically requires an estimate of a value function for the current policy and a sample efficient approach is to use an actor-critic architecture that can work with off-policy data.

These algorithms have the following properties unlike the methods based on DQN discussed in Chapter~\ref{ch:value-based_methods}:
\begin{itemize}
\item They are able to work with continuous action spaces. This is particularly interesting in applications such as robotics, where forces and torques can take a continuum of values.
\item They can represent stochastic policies, which is useful for building policies that can explicitly explore. This is also useful in settings where the optimal policy is a stochastic policy (e.g., in a multi-agent setting where the Nash equilibrium is a stochastic policy).
\end{itemize}

However, another approach is to combine policy gradient methods directly with off-policy Q-learning \citep{o2016pgq}.
In some specific settings, depending on the loss function and the entropy regularization used, value-based methods and policy-based methods are equivalent \citep{fox2015taming,o2016pgq,haarnoja2017reinforcement,schulman2017equivalence}.
For instance, when adding an entropy regularization, Eq. \ref{eq:score function gradient estimator} can be written as
\begin{equation}
\begin{split}
\nabla_w V^{\pi_w}(s_0) & = \mathbb{E}_{s, a} \left[ \nabla_w \left(\log~\pi_w (s,a) \right)Q^{\pi_w}\left(s,a\right) \right] + \alpha \mathbb E_s \nabla_w H^{\pi_w}(s).
 \end{split}
 \label{eq:score function gradient estimator with entrop}
\end{equation}
where $H^\pi(s)=-\sum_a \pi(s, a) \log \pi(s, a)$. 
From this, one can note that an optimum is satisfied by the following policy:
$\pi_w(s, a) = exp(A^{\pi_w}(s, a)/\alpha-H^{\pi_w}(s))$.
Therefore, we can use the policy to derive an estimate of the advantage function:
$\tilde A^{\pi_w}(s, a) = \alpha(\log \pi_w(s,a)+H^\pi(s))$.
We can thus think of all model-free methods as different facets of the same approach.

One remaining limitation is that both value-based and policy-based methods are model-free and they do not make use of any model of the environment. The next chapter describes algorithms with a model-based approach.

\chapter{Model-based methods for deep RL}
\label{ch:model-based}

In Chapters \ref{ch:value-based_methods} and \ref{ch:policy-based_methods}, we have discussed the model-free approach that rely either on a value-based or a policy-based method.
In this chapter, we introduce the model-based approach that relies on a model of the environment (dynamics and reward function) in conjunction with a planning algorithm.
In \S\ref{sec:integr_learn_plan}, the respective strengths of the model-based versus the model-free approaches are discussed, along with how the two approaches can be integrated.

\section{Pure model-based methods}
\label{sec:model-based}
A model of the environment is either explicitly given (e.g., in the game of Go for which all the rules are known a priori) or learned from experience. To learn the model, yet again function approximators bring  significant advantages in high-dimensional (possibly partially observable) environments \citep{oh2015action, mathieu2015deep, finn2016unsupervised, kalchbrenner2016video, duchesne2017machine, nagabandi2018neural}.
The model can then act as a proxy for the actual environment.

When a model of the environment is available, planning consists in interacting with the model to recommend an action.
In the case of discrete actions, lookahead search is usually done by generating potential trajectories.
In the case of a continuous action space, trajectory optimization with a variety of controllers can be used.

\subsection{Lookahead search}

A lookahead search in an MDP iteratively builds a decision tree where the current state is the root node.
It stores the obtained returns in the nodes and focuses attention on promising potential trajectories.
The main difficulty in sampling trajectories is to balance \textit{exploration} and \textit{exploitation}.
On the one hand, the purpose of exploration is to gather more information on the part of the search tree where few simulations have been performed (i.e., where the expected value has a high variance).
On the other hand, the purpose of exploitation is to refine the expected value of the most promising moves.

\begin{figure}[!ht]
 \centering
 \begin{tikzpicture}[scale=.75,font=\footnotesize
]
\def\lvldist{15mm}
\def\rlvldist{0.3}
\tikzstyle{level 1}=[level distance=\lvldist,sibling distance=20mm]
\tikzstyle{level 2}=[level distance=\lvldist,sibling distance=5mm]
\tikzstyle{level 3}=[level distance=\lvldist,sibling distance=5mm]

\tikzset{
solid node/.style={fill=black},
solidr node/.style={fill=black!45!green},
hollow node/.style={},
emph/.style={color=black,edge from parent/.style={dashed,thick,draw, black}},
gre/.style={color=black!45!green},
bla/.style={color=black},
emphgreen/.style={color=black!45!green, edge from parent/.style={dashed,thick,draw, color=black!45!green}},
}

\node(0)[solidr node,label= {above:{$s_t$}} ]{}

child[emph]{node[solid node]{}
child[emph]{node[solid node]{} }
child[emph]{node[solid node]{} }
child[emph]{node[solid node]{}}
edge from parent node[left,xshift=-3]{$$}
}
child[gre]{node[solid node]{}
  child[gre]{node[solid node]{}
    child[gre]{node[solid node]{}
      }
    child[emph]{node[solid node]{}}
    child[emph]{node[solid node]{}}
    }
  child[emph]{node[solid node]{}}
  child[emph]{node[solid node]{}}
  edge from parent node[right,xshift=3]{$$}
  }
child[emph]{node[solid node]{}};

\node(rand)[inner sep=0.1,minimum size=0] at (-1, -4.6) {};
\node(rand1)[fill=black!45!green,inner sep=2, label={[yshift=0.5cm, text width=2cm, text centered, color=black!45!green]right:{$a^{(i)} \in \mathcal A$}}] at (-1, -5.4-\rlvldist) {};
\node(rand2)[fill=black!45!green,inner sep=2] at (-1-\rlvldist, -5.4-2*\rlvldist) {};
\node(rand3)[fill=black!45!green,inner sep=2, label={[label distance=0.3cm, text width=2cm, text centered, color=black!45!green]right:{Monte-carlo \\ simulation}}] at (-1-\rlvldist, -5.4-3*\rlvldist) {};
\node(rand4)[fill=black!45!green,inner sep=2] at (-1, -5.4-4*\rlvldist) {};
\node(rand5)[fill=black!45!green,inner sep=2] at (-1+\rlvldist, -5.4-5*\rlvldist) {};
\node(rand6)[circle, color=black,draw=black!45!green, label={[label distance=-0.3cm, text width=2cm, text centered, color=black!45!green]left:{End \\ state}}] at (-1+\rlvldist, -5.4-6*\rlvldist) {};
\draw [gre] (rand) to (rand1);
\draw [gre] (rand1) to (rand2);
\draw [gre] (rand2) to (rand3);
\draw [gre] (rand3) to (rand4);
\draw [gre] (rand4) to (rand5);
\draw [gre] (rand5) to (rand6);

\node(st)[label=right:{$t$}] at (2.2, 0) {};
\node(st1)[label=right:{$t+1$}] at (2.2, -\lvldist) {};
\node(st2)[label=right:{$t+2$}] at (2.2, -2*\lvldist) {};
\node(st3)[] at (2.2, -3*\lvldist) {};

\draw [->] (st) to [out=-40,in=40] (st1);
\draw [->] (st1) to [out=-40,in=40] (st2);
\draw [dashed,->] (st2) to [out=-40,in=40] (st3);


\node(3)[solidr node, right of=0, xshift=180]{}

child[solid]{node[solid node]{}
child[solid]{node[solid node]{} }
child[solid]{node[solid node]{} }
child[solid]{node[solid node]{}}
edge from parent node[left,xshift=-3]{$$}
}
child[black]{node(1)[solid node]{}
child[black]{node(4)[solid node]{}
  child[black]{node(5)[solid node]{}
    child[solid]{node(sss4)[solid node, label={[yshift=0.5cm, text width=2cm, text centered, color=black]right:{$a^{(i)} \in \mathcal A$}}]{}}
    }
  child[solid]{node[solid node]{}}
  child[solid]{node[solid node]{}}
  }
child[solid]{node[solid node]{}}
child[solid]{node[solid node]{}}
edge from parent node[right,xshift=3]{$$}
}
child[solid]{node[solid node]{}};

\draw [dotted,->] (rand6) to [out=0,in=-180] (3);
\draw [dotted,->] (rand6) to [out=0,in=-128] (1);
\draw [dotted,->] (rand6) to [out=0,in=-140] (4);
\draw [dotted,->] (rand6) to [out=0,in=-160] (5);
\draw [dotted,->] (rand6) to [out=0,in=-180] (sss4);

\end{tikzpicture}
 \caption{Illustration of how a MCTS algorithm performs a Monte-Carlo simulation and builds a tree by updating the statistics of the different nodes. Based on the statistics gathered for the current node $s_t$, the MCTS algorithm chooses an action to perform on the actual environment.}
 \label{fig:MCTS}
\end{figure}
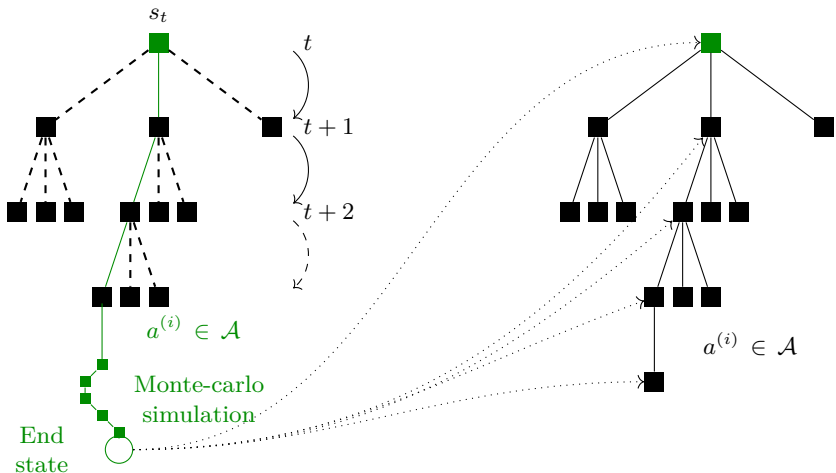

Monte-Carlo tree search (MCTS) techniques \citep{browne2012survey} are popular approaches to lookahead search.
Among others, they have gained popularity thanks to prolific achievements in the challenging task of computer Go \citep{brugmann1993monte, gelly2006modification, silver2016mastering}.
The idea is to sample multiple trajectories 
from the current state until a terminal condition is reached (e.g., a given maximum depth) (see Figure \ref{fig:MCTS} for an illustration).
From those simulation steps, the MCTS algorithm then recommends an action to take.

Recent works have developed strategies to directly learn end-to-end the model, along with how to make the best use of it, without relying on explicit tree search techniques \citep{pascanu2017learning}.
These approaches show improved sample efficiency, performance, and robustness to model misspecification compared to the separated approach (simply learning the model and then relying on it during planning).

\subsection{Trajectory optimization}
Lookahead search techniques are limited to discrete actions, and alternative techniques have to be used for the case of continuous actions.
If the model is differentiable, one can directly compute an analytic policy gradient by backpropagation of rewards along trajectories \citep{nguyen1990neural}.
For instance, PILCO \citep{deisenroth2011pilco} uses Gaussian processes to learn a probabilistic model of the dynamics. 
It can then explicitly use the uncertainty for planning and policy evaluation in order to achieve a good sample efficiency.
However, the gaussian processes have not been able to scale reliably to high-dimensional problems.

One approach to scale planning to higher dimensions is to aim at leveraging the generalization capabilities of deep learning.
For instance, \citet{wahlstrom2015pixels} uses a deep learning model of the dynamics (with an auto-encoder) along with a model in a latent state space.
Model-predictive control \citep{morari1999model} can then be used to find the policy by repeatedly solving a finite-horizon optimal control problem in the latent space.
It is also possible to build a probabilistic generative model in a latent space with the objective that it possesses a locally linear dynamics, which allows control to be performed more efficiently \citep{watter2015embed}.
Another approach is to use the trajectory optimizer as a teacher rather than a demonstrator:
guided policy search \citep{levine2013guided} takes a few sequences of actions suggested by another controller.
iIt then learns to adjust the policy from these sequences.
Methods that leverage trajectory optimization have demonstrated many capabilities, for instance in the case of simulated 3D bipeds and quadrupeds \citep[e.g.,][]{mordatch2015interactive}.

\section{Integrating model-free and model-based methods}
\label{sec:integr_learn_plan}
The respective strengths of the model-free versus model-based approaches depend on different factors.
First, the best suited approach depends on whether the agent has access to a model of the environment.
If that's not the case, the learned model usually has some inaccuracies that should be taken into account.
Note that learning the model can share the hidden-state representations with a value-based approach by sharing neural network parameters \citep{li2015recurrent}.

Second, a model-based approach requires working in conjunction with a planning algorithm (or controller), which is often computationally demanding. The time constraints for computing the policy $\pi(s)$ via planning must therefore be taken into account (e.g., for applications with real-time decision-making or simply due to resource limitations).

Third, for some tasks, the structure of the policy (or value function) is the easiest one to learn, but for other tasks, the model of the environment may be learned more efficiently due to the particular structure of the task (less complex or with more regularity).
Thus, the most performant approach depends on the structure of the model, policy, and value function (see the coming Chapter \ref{ch:generalization} for more details on generalization).
Let us consider two examples to better understand this key consideration. In a labyrinth where the agent has full observability, it is clear how actions affect the next state and the dynamics of the model may easily be generalized by the agent from only a few tuples (for instance, the agent is blocked when trying to cross a wall of the labyrinth). Once the model is known, a planning algorithm can then be used with high performance.
Let us now discuss another example where, on the contrary, planning is more difficult: an agent has to cross a road with random events happening everywhere on the road.
Let us suppose that the best policy is simply to move forward except when an object has just appeared in front of the agent.
In that case, the optimal policy may easily be captured by a model-free approach, while a model-based approach would be more difficult (mainly due to the stochasticity of the model which leads to many different possible situations, even for one given sequence of actions).

\begin{figure}[!ht]
 \centering
     \resizebox{0.5\textwidth}{!}{%
\begin{tikzpicture}
  \begin{scope}[blend group = soft light]
    \fill[red!30!white]   ( 90:1.7) circle (2.2);
    \fill[green!30!white] (210:1.7) circle (2.2);
    \fill[blue!30!white]  (330:1.7) circle (2.2);
  \end{scope}
  \node at ( 90:2.5)    {
        \begin{tabular}{c}
                        \text{Model-based}\\
                        \text{RL} \\
                        \end{tabular}
                      };
  \node at ( 210:2.5)   {
        \begin{tabular}{c}
                        \text{Value-based}\\
                        \text{RL} \\
                        \end{tabular}
                      };
  \node at ( 330:2.5)   {
        \begin{tabular}{c}
                        \text{Policy-based}\\
                        \text{RL} \\
                        \end{tabular}
                      };
\end{tikzpicture}
}
 \caption{Venn diagram in the space of possible RL algorithms.}
 \label{fig:venn_diagram_RL}
\end{figure}
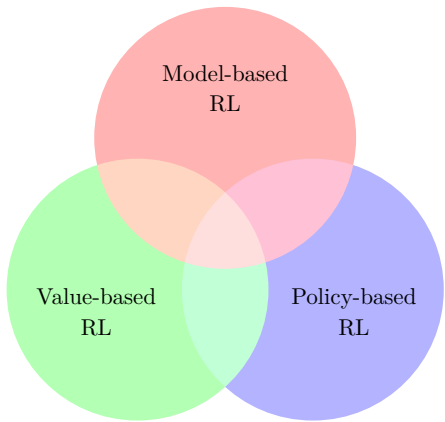

We now describe how it is possible to obtain advantages from both worlds by integrating learning and planning into one end-to-end training procedure so as to obtain an efficient algorithm both in performance (sample efficient) and in computation time. A Venn diagram of the different combinations is given in Figure \ref{fig:venn_diagram_RL}.

When the model is available, one direct approach is to use tree search techniques that make use of both value and policy networks (e.g., \cite{silver2016mastering}). When the model is not available and under the assumption that the agent has only access to a limited number of trajectories,
the key property is to have an algorithm that generalizes well (see Chapter \ref{ch:generalization} for a discussion on generalization).
One possibility is to build a model that is used to generate additional samples for a model-free reinforcement learning algorithm \citep{gu2016continuous}.
Another possibility is to use a model-based approach along with a controller such as MPC to perform basic tasks and use model-free fine-tuning in order to achieve task success \citep{nagabandi2017neural}.

Other approaches build neural network architectures that combine both model-free and model-based elements.
For instance, it is possible to combine a value function with steps of back-propagation through a model \citep{heess2015learning}.
The VIN architecture \citep{tamar2016value} is a fully differentiable neural network with a planning module that learns to plan from model-free objectives (given by a value function). It works well for tasks that involve planning-based reasoning (navigation tasks) from one initial position to one goal position and it demonstrates strong generalization in a few different domains.

In the same spirit, the predictron \citep{silver2016predictron} is aimed at developing a more generally applicable algorithm that is effective in the context of planning.
It works by implicitly learning an internal model in an abstract state space, which is used for policy evaluation.
The predictron is trained end-to-end to learn, from the abstract state space, (i) the immediate reward and (ii) value functions over multiple planning depths.
The predictron architecture is limited to policy evaluation, but the idea was extended to an algorithm that can learn an optimal policy in an architecture called VPN \citep{oh2017value}. Since VPN relies on n-step Q-learning, it requires however on-policy data.

Other works have proposed architectures that combine model-based and model-free approaches. 
Schema Networks \citep{kansky2017schema} learn the dynamics of an environment directly from data by enforcing some relational structure. 
The idea is to use a richly structured architecture such that it provides robust generalization thanks to an object-oriented approach for the model.

I2As \citep{weber2017imagination} does not use the model to directly perform planning but it uses the predictions as additional context in deep policy networks.
The proposed idea is that I2As could learn to interpret predictions from the learned model to construct implicit plans.

TreeQN \citep{farquhar2017treeqn} constructs a tree by recursively applying an implicit transition model in an implicitly learned abstract state space, built by estimating Q-values.
\citet{farquhar2017treeqn} also propose ATreeC, which is an actor-critic variant that augments TreeQN with a softmax layer to form a stochastic policy network.

The CRAR agent explicitly learns both a value function and a model via a shared low-dimensional learned encoding of the environment, which is meant to capture summarized abstractions and allow for efficient planning \citep{franccois2018combined}. 
By forcing an expressive representation, the CRAR approach creates an interpretable low-dimensional representation of the environment, even far temporally from any rewards or in the absence of model-free objectives.

Improving the combination of model-free and model-based ideas is one key area of research for the future development of deep RL algorithms.
We therefore expect to see smarter and richer structures in that domain.

\chapter{The concept of generalization}
\label{ch:generalization}

Generalization is a central concept in the field of machine learning, and reinforcement learning is no exception. In an RL algorithm (model-free or model-based), generalization refers to either
\begin{itemize}
\item the capacity to achieve good performance in an environment where limited data has been gathered, or
\item the capacity to obtain good performance in a related environment.
\end{itemize}


In the former case, the agent must learn how to behave in a test environment that is identical to the one it has been trained on. In that case, the idea of generalization is directly related to the notion of \textit{sample efficiency} (e.g., when the state-action space is too large to be fully visited).
In the latter case, the test environment has common patterns with the training environment but can differ in the dynamics and the rewards. For instance, the underlying dynamics may be the same but a transformation on the observations may have happened (e.g., noise, shift in the features, etc.).
That case is related to the idea of transfer learning (discussed in \S\ref{sec:transfer}) and meta-learning (discussed in \S\ref{sec:distrib_envs}).

Note that, in the online setting, one mini-batch gradient update is usually done at every step.
In that case, the community has also used the term sample efficiency to refer to how fast the algorithm learns, which is measured in terms of performance for a given  number of steps (number of learning steps=number of transitions observed). However, in that context, the result depends on many different elements.
It depends on the learning algorithm and it is, for instance, influenced by the possible variance of the target in a model-free setting.
It also depends on the exploration/exploitation, which will be discussed in \S\ref{sec:explo-exploit} (e.g, instabilities may be good). Finally, it depends on the actual generalization capabilities.

In this chapter, the goal is to study specifically the aspect of generalization.
We are not interested in the number of mini-batch gradient descent steps that are required but rather in the performance that a deep RL algorithm can have in the offline case where the agent has to learn from limited data.
Let us consider the case of a finite dataset $D$ obtained on the exact same task as the test environment. Formally, a dataset available to the agent $D \sim \mathcal D$\label{ntn:rand_var_Ds} can be defined as a set of four-tuples $<s, a, r, s'> \in \mathcal S \times \mathcal A \times \mathcal R \times \mathcal S$ gathered
by sampling independently and identically (i.i.d.)\footnote{That i.i.d. assumption can, for instance, be obtained from a given distribution of initial states by following a stochastic sampling policy that ensures a non-zero probability of taking any action in any given state. That sampling policy should be followed during at least $H$ time steps with the assumption that all states of the MDP can be reached in a number of steps smaller than $H$ from the given distribution of initial states.}
\begin{itemize}
\item a given number of state-action pairs $(s, a)$ from some fixed distribution with $\mathbb P(s, a) >0$, $\forall (s,a) \in \mathcal S \times \mathcal A$,
\item a next state $s' \sim T(s,a,\cdot)$, and
\item a reward $r=R(s,a,s')$.
\end{itemize}
We denote by $D_{\infty}$ the particular case of a dataset $D$ where the number of tuples tends to infinity.

A learning algorithm can be seen as a mapping of a dataset $D$ into a policy $\pi_{D}$ (independently of whether the learning algorithm has a model-based or a model-free approach). In that case, we can decompose the suboptimality of the expected return as follows:

\small
\begin{equation}
\begin{split}
\underset{D\sim \mathcal D}{\mathbb E}  \left[ V^{\pi^*}(s) -V^{\pi_{D}}(s)\right] & = \underset{D\sim \mathcal D}{\mathbb E} \left[ V^{\pi^*}(s)-V^{\pi_{D_{\infty}}}(s) + V^{\pi_{D_{\infty}}}(s)-V^{\pi_{D}}(s)\right]\\
& = \underbrace{(V^{\pi^*}(s)-V^{\pi_{D_{\infty}}}(s))}_\text{\shortstack{asymptotic bias}} \\
& + \underbrace{ \underset{D\sim \mathcal D}{\mathbb E} \Big[ V^{\pi_{D_{\infty}}}(s)-V^{\pi_{D}}(s)\Big]}_\text{\shortstack{
error due to finite size of the dataset $D$} }.
\end{split}
\label{bias-overfitting}
\end{equation}
\normalsize
This decomposition highlights two different terms: (i)~an asymptotic bias which is independent of the quantity of data and (ii)~an overfitting term directly related to the fact that the amount of data is limited.
The goal of building a policy $\pi_{D}$ from a dataset $D$ is to obtain the lowest overall suboptimality.
To do so, the RL algorithm should be well adapted to the task (or the set of tasks).

In the previous section, two different types of approaches (model-based and model-free) have been discussed, as well as how to combine them.
We have discussed the algorithms that can be used for different approaches but we have in fact left out many important elements that have an influence on the bias-overfitting tradeoff (e.g., \cite{zhang2018overfitting, zhang2018dissection} for illustrations of overfitting in deep RL).

\begin{figure}[ht!]
     \centering
    \resizebox{0.65\textwidth}{!}{%
  \begin{tikzpicture}[scale=1]
     \node[anchor=south west,inner sep=0] at (1.2,0) {\includegraphics[width=5cm]{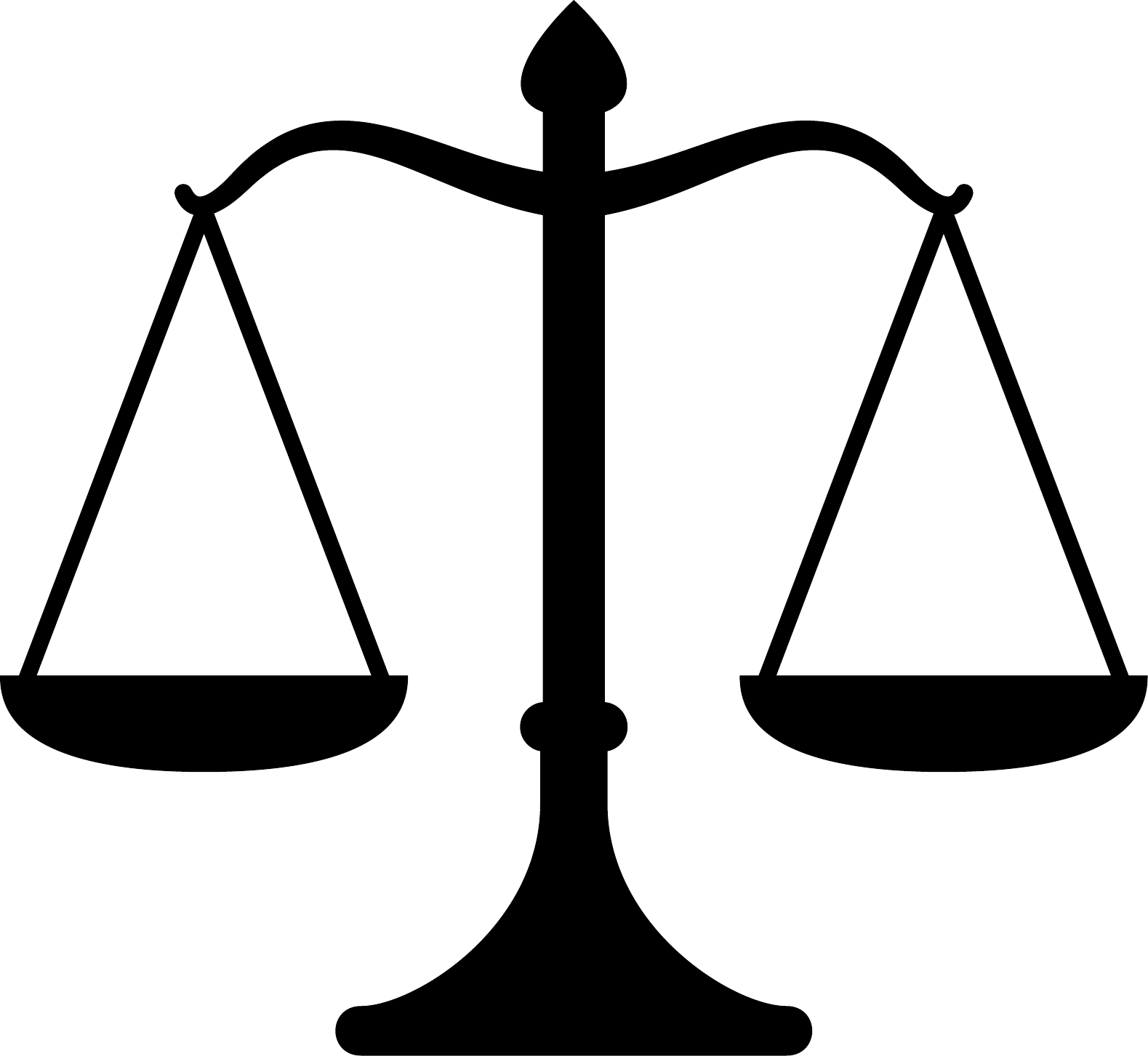}};
     \node at (2.1,2) {Data};
     \node at (5.35,2.1) {
    \resizebox{1.2cm}{!}{%
 \begin{tabular}{cc}
 Policy\\
 class
 \end{tabular}
 }
 };
 \node (A) at (1, 0){};
 \node (B) at (1, 5){};
 \node at (-0.2, 2.5){
 \begin{tabular}{cc}
 \% of the\\
 error\\
 due to\\
 overfitting
 \end{tabular}
 };
 \node (C) at (6.4, 0){};
 \node (D) at (6.4, 5){};
 \node at (7.8, 2.5){
 \begin{tabular}{cc}
 \% of the\\
 error due to\\
 asymptotic\\
 bias
 \end{tabular}
 };
 \draw[line width=2pt,->] (A) edge (B) ;
 \draw[line width=2pt,->] (C) edge (D) ;
 \end{tikzpicture}
 }
      \caption{Schematic representation of the bias-overfitting tradeoff.}
     \label{fig:balance}
 \end{figure}
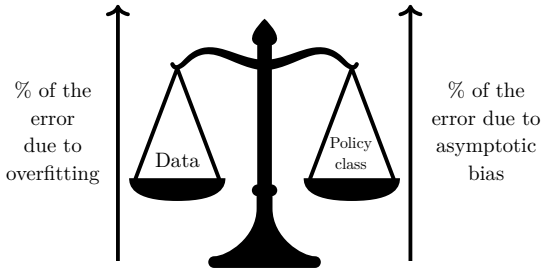

As illustrated in Figure \ref{fig:balance}, improving generalization can be seen as a tradeoff between (i) an error due to the fact that the algorithm trusts completely the frequentist assumption (i.e., discards any uncertainty on the limited data distribution) and (ii) an error due to the bias introduced to reduce the risk of overfitting.
For instance, the function approximator can be seen as a form of structure introduced to force some generalization, at the risk of introducing a bias.
When the quality of the dataset is low, the learning algorithm should favor more robust policies (i.e., consider a smaller class of policies with stronger generalization capabilities).
When the quality of the dataset increases, the risk of overfitting is lower and the learning algorithm can trust the data more, hence reducing the asymptotic bias.

As we will see, for many algorithmic choices, there is in practice a tradeoff to be made between asymptotic bias and overfitting that we simply call "bias-overfitting tradeoff". In this section, we discuss the following key elements that are at stake when one wants to improve generalization in deep RL:
\begin{itemize}
\item the state representation,
\item the learning algorithm (type of function approximator and model-free vs model-based),
\item the objective function (e.g., reward shaping, tuning the training discount factor), and
\item using hierarchical learning.
\end{itemize}



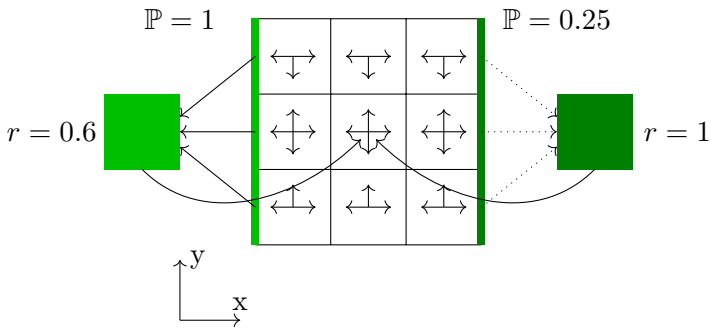
\begin{figure}[ht!]
     \centering
  \begin{tikzpicture}
  \draw[step=1cm] (-2,-2) grid (1,1);

  \fill[color=black!25!green] (-4,-1) rectangle (-3,0);
  \fill[color=black!25!green] (-2.05,-2) rectangle (-1.95,1);
  \fill[color=black!50!green] (2,-1) rectangle (3,0);
  \fill[color=black!50!green] (0.95,-2) rectangle (1.05,1);

  \foreach \i in {0,...,2}
    \draw [->]  (-2,-1.5+\i) -- (-3,-0.7+\i*0.2);
  \foreach \i in {0,...,2}
    \draw [->, dotted]  (1,-1.5+\i) -- (2,-0.7+\i*0.2);

  \foreach \i in {0,...,2}{
     \foreach \j in {0,...,2}{
    \draw [->]  (-1.5+\i,-1.5+\j) -- (-1.2+\i,-1.5+\j);
    \draw [->]  (-1.5+\i,-1.5+\j) -- (-1.8+\i,-1.5+\j);
    }
  }
  \foreach \i in {0,...,2}{
     \foreach \j in {0,...,1}{
      \draw [->]  (-1.5+\i,-1.5+\j) -- (-1.5+\i,-1.2+\j);
      \draw [->]  (-1.5+\i,-0.5+\j) -- (-1.5+\i,-0.8+\j);
    }
  }
  \draw [->] (-3.5,-1) to [out=-45,in=-135] (-0.6,-0.6);
  \draw [->] (2.5,-1) to [out=-135,in=-45] (-0.4,-0.6);

  \draw [->]  (-3,-3) -- (-3,-2.2) node[right] {y};
  \draw [->]  (-3,-3) -- (-2.2,-3) node[above] {x};

  \node[] at (-3,1) {$\mathbb P=1$};
  \node[] at (2,1) {$\mathbb P=0.25$};
  \node[] at (-4.7,-0.5) {$r=0.6$};
  \node[] at (3.6,-0.5) {$r=1$};

  \end{tikzpicture}
      \caption{Representation of a simple MDP that illustrates the need of generalization.}
     \label{fig:simple_MDP}
 \end{figure}

Throughout those discussions, a simple example is considered.
This example is, by no means, representative of the complexity of real-world problems but it is enlightening to simply illustrate the concepts that will be discussed.
Let us consider an MDP with $N_{\mathcal S}$ states,  $N_{\mathcal S}=11$ and $N_{\mathcal A}$ actions, $N_{\mathcal A}=4$.
Let us suppose that the main part of the environment is a square $3 \times 3$ grid world (each represented by a tuple $(x,y)$ with $x=\{0,1,2\}, y=\{0,1,2\}$), such as illustrated in Figure \ref{fig:simple_MDP}.
The agent starts in the central state $(1,1)$. In every state, it selects one of the $4$ actions corresponding to $4$ cardinal directions (up, down, left and right), which leads the agent to transition deterministically in a state immediately next to it, except when it tries to move out of the domain.
On the upper part and lower part of the domain, the agent is stuck in the same state if it tries to move out of the domain.
On the left, the agent transitions deterministically to a given state, which will provide a reward of $0.6$ for any action at the next time step.
On the right side of the square, the agent transitions with a probability $25\%$ to another state that will provide, at the next time step, a reward of $1$ for any action (the rewards are $0$ for all other states).
When a reward is obtained, the agent transitions back to the central state.

In this example, if the agent has perfect knowledge of its environment, the best expected cumulative reward (for a discount factor close to $1$) would be to always go to the left direction and repeatedly gather a reward of $0.6$ every 3 steps (as compared to gathering a reward of $1$ every 6 steps on average).
Let us now suppose that only limited information has been obtained on the MDP with only one tuple of experience $<s,a,r,s'>$ for each couple $<s,a>$.
According to the limited data in the frequentist assumption, there is a rather high probability ($\sim 58\%$) that at least one transition from the right side seems to provide a deterministic access to $r=1$.
In those cases and for either a model-based or a model-free approach, if the learning algorithm comes up with the optimal policy in an empirical MDP built from the frequentist statistics, it would actually suffer from poor generalization as it would choose to try to obtain the reward $r=1$.

We discuss hereafter the different aspects that can be used to avoid overfitting to limited data; we show that it is done by favoring robust policies within the policy class, usually at the expense of introducing some bias. At the end, we also discuss how the bias-overfitting tradeoff can be used in practice to obtain the best performance from limited data.

\section{Feature selection}
The idea of selecting the right features for the task at hand 
is key in the whole field of machine learning and also highly prevalent in reinforcement learning 
(see e.g., \cite{munos2002variable,ravindran2004algebraic,leffler2007efficient,kroon2009automatic,dinculescu2010approximate,li2011unbiased,ortner2014selecting,mandel2014offline,jiang2015abstraction,guo2017sample,franccois2017on}). 
The appropriate level of abstraction plays a key role in the bias-overfitting tradeoff and one of the key advantages of using a small but rich abstract representation is to allow for improved generalization. 

\begin{figure}[ht!]
     \centering
  \scalebox{0.8}{
  \begin{tikzpicture}
  \draw[step=1cm] (-2,-2) grid (1,1);

  \foreach \i in {0,...,2}{
     \foreach \j in {0,...,2}{
       \pgfmathsetmacro\result{int(\i+(-\j+2)*3)}
    \node at (-1.5+\i,-1.5+\j){$s^{(\result)}$};
    }
  }
  \draw [->]  (-4,-2) -- (-4,-1.2) node[right] {y};
  \draw [->]  (-4,-2) -- (-3.2,-2) node[above] {x};
  \node[minimum width=3cm, align=center] at (-0.5,-3.2){Environment};

  \draw[step=1cm] (2,-2) grid (5,1);
  \foreach \i in {0,...,2}{
     \foreach \j in {0,...,2}{
       \pgfmathsetmacro\result{int(-\j+2)}
    \node at (2.5+\i,-1.5+\j){$(\i,\result)$};
    }
  }
  \node[minimum width=3cm, align=center] at (3.5,-3.2){States \\representation\\ with a set of \\features $(x,y)$};

  \draw[step=1cm] (6,-2) grid (9,1);
  \foreach \i in {0,...,2}{
     \foreach \j in {0,...,2}{
    \node at (6.5+\i,-1.5+\j){($\i$)};
    }
  }
  \node[minimum width=3cm, align=center] at (7.5,-3.2){Feature\\selection where\\only the $x$-coordinate\\has been kept};
  \end{tikzpicture}
  }
      \caption{Illustration of the state representation and feature selection process. In this case, after the feature selection process, all states with the same $x$-coordinate are considered as indistinguishable.}
     \label{fig:feature_sel}
 \end{figure}
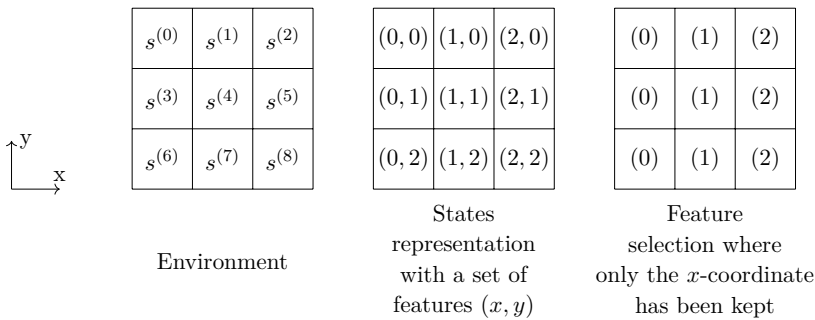

\paragraph{Overfitting} When considering many features on which to base the policy (in the example the y-coordinate of the state as illustrated in Figure \ref{fig:feature_sel}), an RL algorithm may take into consideration spurious correlations, which leads to overfitting (in the example, the agent may infer that the y-coordinate changes something to the expected return because of the limited data).

\paragraph{Asymptotic bias} Removing features that discriminate states with a very different role in the dynamics introduces an asymptotic bias. Indeed, the same policy would be enforced on undistinguishable states, hence leading to a sub-optimal policy.

\paragraph{}

In deep RL, one approach is to first infer a factorized set of generative factors from the observations.
This can be done for instance with an encoder-decoder architecture variant \citep{higgins2017darla,zhang2018decoupling}. 
These features can then be used as inputs to a reinforcement learning algorithm. The learned representation can, in some contexts, greatly help for generalization as it provides a more succinct representation that is less prone to overfitting.
However, an auto-encoder is often too strong of a constraint. On the one hand, some features may be kept in the abstract representation because they are important for the reconstruction of the observations, though they are otherwise irrelevant for the task at hand (e.g., the color of the cars in a self-driving car context). On the other hand, crucial information about the scene may also be discarded in the latent representation, particularly if that information takes up a small proportion of the observations $x$ in pixel space \citep{higgins2017darla}.
Note that in the deep RL setting, the abstraction representation is intertwined with the use of deep learning.
This is discussed in detail in the following section.

\section{Choice of the learning algorithm and function approximator selection}

The function approximator in deep learning characterizes how the features will be treated into higher levels of abstraction (a fortiori it can thus give more or less weight to some features).
If there is, for instance, an attention mechanism in the first layers of a deep neural network, the mapping made up of those first layers can be seen as a feature selection mechanism.

On the one hand, if the function approximator used for the value function and/or the policy and/or the model is too simple, an asymptotic bias may appear.
On the other hand, when the function approximator has poor generalization, there will be a large error due to the finite size of the dataset (overfitting).
In the example above, a particularly good choice of a model-based or model-free approach associated with a good choice of a function approximator could infer that the y-coordinate of the state is less important than the x-coordinate, and generalize that to the policy.

Depending on the task, finding a performant function approximator is easier in either a model-free or a model-based approach. The choice of relying more on one or the other approach is thus also a crucial element to improve generalization, as discussed in \S\ref{sec:integr_learn_plan}.

One approach to mitigate non-informative features is to force the agent to acquire a set of symbolic rules adapted to the task and to reason on a more abstract level.
This abstract level reasoning and the improved generalization have the potential to induce high-level cognitive functions such as transfer learning and analogical reasoning \citep{garnelo2016towards}.
For instance, the function approximator may embed a relational learning structure \citep{santoro2017simple} and thus build on the idea of relational reinforcement learning \citep{dvzeroski2001relational}.


\subsection{Auxiliary tasks}
\label{sec:aux_tasks}


In the context of deep reinforcement learning, it was shown by \citet{jaderberg2016reinforcement} that augmenting a deep reinforcement learning agent with auxiliary tasks within a jointly learned representation can drastically improve sample efficiency in learning.
This is done by maximizing simultaneously many pseudo-reward functions such as immediate reward prediction ($\gamma=0$), predicting pixel changes in the next observation, or predicting activation of some hidden unit of the agent's neural network.
The argument is that learning related tasks introduces an inductive bias that causes a model to build features in the neural network that are useful for the range of tasks \citep{ruder2017overview}. Hence, this formation of more significant features leads to less overfitting.

In deep RL, it is possible to build an abstract state such that it provides sufficient information for simultaneously fitting an internal meaningful dynamics as well as the estimation of the expected value of an optimal policy. By explicitly learning both the model-free and model-based components through the state representation, along with an approximate entropy maximization penalty, the CRAR agent \citep{franccois2018combined} shows how it is possible to learn a low-dimensional representation of the task.
In addition, this approach can directly make use of a combination of model-free and model-based, with planning happening in a smaller latent state space.

\section{Modifying the objective function}
In order to improve the policy learned by a deep RL algorithm, one can optimize an objective function that diverts from the actual objective. By doing so, a bias is usually introduced but this can in some cases help with generalization. The main approaches to modify the objective function are either (i)~to modify the reward of the task to ease learning (reward shaping), or (ii)~tune the discount factor at training time.

\subsection{Reward shaping}
Reward shaping is a heuristic for faster learning. In practice, reward shaping uses prior knowledge by giving intermediate rewards for actions that lead to desired outcome. It is usually formalized as a function $F(s,a,s')$ added to the original reward function $R(s,a,s')$ of the original MDP \citep{ng1999policy}. This technique is often used in deep reinforcement learning to improve the learning process in settings with sparse and delayed rewards \citep[e.g.,][]{lample2017playing}.

\subsection{Discount factor}
When the model available to the agent is estimated from data, the policy found using a shorter planning horizon can actually be better than a policy learned with the true horizon  \citep{petrik2009biasing, jiang2015dependence}.
On the one hand, artificially reducing the planning horizon leads to a bias since the objective function is modified.
On the other hand, if a long planning horizon is targeted (the discount factor $\gamma$ is close to 1), there is a higher risk of overfitting.
This overfitting can intuitively be understood as linked to the accumulation of the errors in the transitions and rewards estimated from data as compared to the actual transition and reward probabilities. 
In the example above (Figure \ref{fig:simple_MDP}), in the case where the upper right or lower right states would seem to lead deterministically to $r=1$ from the limited data, one may take into account that it requires more steps and thus more uncertainty on the transitions (and rewards). In that context, a low training discount factor would reduce the impact of rewards that are temporally distant.
In the example, a discount factor close to 0 would discount the estimated rewards at three time steps much more strongly than the rewards two time steps away, hence practically discarding the potential rewards that can be obtained by going through the corners as compared to the ones that only require moving along the x-axis.

In addition to the bias-overfitting tradeoff, a high discount factor also requires specific care in value iteration algorithms as it can lead to instabilities in convergence.
This effect is due to the mappings used in the value iteration algorithms with bootstrapping (e.g., Equation \ref{op_mapping_Bellman_Q} for the Q-learning algorithm) that propagate errors more strongly with a high discount factor.
This issue is discussed by \citet{gordon1999approximate} with the notion of \textit{non-expansion/expansion mappings}.
When bootstrapping is used in a deep RL value iteration algorithm, the risk of instabilities and overestimation of the value function is empirically stronger for a discount factor close to one \citep{franccois2015discount}.

\section{Hierarchical learning}
\label{sec:hierarchical}
The possibility of learning temporally extended actions (as opposed to atomic actions that last for one time-step) has been formalized under the name of options \citep{sutton1999between}. Similar ideas have also been denoted in the literature as macro-actions \citep{mcgovern1997roles} or abstract actions \citep{hauskrecht1998hierarchical}.
The usage of options is an important challenge in RL because it is essential when the task at hand requires working on long time scales while developing generalization capabilities and easier transfer learning between the strategies.
A few recent works have brought interesting results in the context of fully differentiable (hence learnable in the context of deep RL) options discovery.
In the work of \cite{bacon2016option}, an \textit{option-critic} architecture is presented with the capability of learning simultaneously the internal policies and the termination conditions of options, as well as the policy over options.
In the work of \cite{vezhnevets2016strategic}, the deep recurrent neural network is made up of two main elements. The first module generates an action-plan (stochastic plan of future actions)
while the second module maintains a commitment-plan which determines when the action-plan has to be updated or terminated.
Many variations of these approaches are also of interest \citep[e.g.,][]{kulkarni2016hierarchical,mankowitz2016adaptive}.
Overall, building a learning algorithm that is able to do hierarchical learning can be a good way of constraining/favoring some policies that have interesting properties and thus improving generalization.

\section{How to obtain the best bias-overfitting tradeoff}
From the previous sections, it is clear that there is a large variety of algorithmic choices and parameters that have an influence on the bias-overfitting tradeoff (including the choice of approach between model-based and model-free).
An overall combination of all these elements provides a low overall sub-optimality.

For a given algorithmic parameter setting and keeping all other things equal, the right level of complexity is the one at which the increase in bias is equivalent to the reduction of overfitting (or the increase in overfitting is equivalent to the reduction of bias).
However, in practice, there is usually not an analytical way to find the right tradeoffs to be made between all the algorithmic choices and parameters. Still, there are a variety of practical strategies that can be used. We now discuss them for the batch setting case and the online setting case.

\subsection{Batch setting}
In the batch setting case, the selection of the policy parameters to effectively balance the bias-overfitting tradeoff can be done similarly to that in supervised learning (e.g., cross-validation) as long as the performance criterion can be estimated from a subset of the trajectories from the dataset $D$ not used during training (i.e., a validation set).

One approach is to fit an MDP model to the data via regression (or simply use the frequentist statistics for finite state and action space). The empirical MDP can then be used to evaluate the policy. 
This purely model-based estimator has alternatives that do not require fitting a model.
One possibility is to use a policy evaluation step obtained by generating artificial trajectories from the data, without explicitly referring to a model, thus designing a Model-free Monte Carlo-like (MFMC) estimator \citep{fonteneau2013batch}.
Another approach is to use the idea of \textit{importance sampling} that lets us obtain an estimate of $V^\pi(s)$ from trajectories that come from a behavior policy $\beta \neq \pi$, where $\beta$ is assumed to be known \citep{precup2000eligibility}.
That approach is unbiased but the variance usually grows exponentially in horizon, which renders the method unsuitable when the amount of data is low.
A mix of the regression-based approach and the importance sampling approach is also possible \citep{jiang2016doubly,thomas2016data}, and the idea is to use a \textit{doubly-robust estimator} that is both unbiased and with a lower variance than the importance sampling estimators.

Note that there exists a particular case where the environment's dynamics are known to the agent, but contain a dependence on an exogenous time series (e.g., trading in energy markets, weather-dependent dynamics) for which the agent only has finite data. In that case, the exogenous signal can be broken down in training time series and validation time series \citep{franccois2016deep}. This allows training on the environment with the training time series and this allows estimating any policy on the environment with the validation time series.

\subsection{Online setting}
\label{sec:online}
In the online setting, the agent continuously gathers new experience.
The bias-overfitting tradeoff still plays a key role at each stage of the learning process in order to achieve good sampling efficiency.
Indeed, a performant policy from given data is part of the solution to an efficient exploration/exploitation tradeoff. 
For that reason, progressively fitting a function approximator as more data becomes available can in fact be understood as a way to obtain a good bias-overfitting tradeoff throughout learning.
With the same logic, progressively increasing the discount factor allows optimizing the bias-overfitting tradeoff through learning \citep{franccois2015discount}.
Besides, optimizing the bias-overfitting tradeoff also suggests the possibility to dynamically adapt the feature space and/or the function approximator. For example, this can be done through ad hoc regularization, or by adapting the neural network architecture, using for instance the NET2NET transformation \citep{chen2015net2net}.

\chapter{Particular challenges in the online setting}
\label{ch:challenges_online}
As discussed in the introduction, reinforcement learning can be used in two main settings: (i) the batch setting (also called offline setting), and (ii) the online setting.
In a batch setting, the whole set of transitions $(s,a,r,s')$ to learn the task is fixed. This is in contrast to the \textit{online} setting where the agent can gather new experience gradually.
In the online setting, two specific elements have not yet been discussed in depth.
First, the agent can influence how to gather experience so that it is the most useful for learning.
This is the \textit{exploration/exploitation} dilemma that we discuss in Section \ref{sec:explo-exploit}.
Second, the agent has the possibility to use a replay memory \citep{lin1992self} that allows for a good data-efficiency.
We discuss in Section \ref{sec:exp_replay} what experience to store and how to reprocess that experience.

\section{Exploration/Exploitation dilemma}
\label{sec:explo-exploit}
The exploration-exploitation dilemma is a well-studied tradeoff in RL \citep[e.g.,][]{thrun1992efficient}.
Exploration is about obtaining information about the environment (transition model and reward function) while exploitation is about maximizing the expected return given the current knowledge.
As an agent starts accumulating knowledge about its environment, it has to make a tradeoff between learning more about its environment (exploration) or pursuing what seems to be the most promising strategy with the experience gathered so far (exploitation).

\subsection{Different settings in the exploration/exploitation dilemma}
There exist mainly two different settings.
In the first setting, the agent is expected to perform well without a separate training phase.
Thus, an explicit tradeoff between exploration versus exploitation appears so that the agent should explore only when the learning opportunities are valuable enough for the future to compensate what direct exploitation can provide. The sub-optimality $\underset{s_0}{\mathbb{E}} \ V^*(s_0)-V^\pi(s_0)$ of an algorithm obtained in this context is known as the \textit{cumulative regret} \footnote{This term is mainly used in the bandit community where the agent is in only one state and where a distribution of rewards is associated to each action; see e.g., \cite{bubeck2011pure}.}.
The deep RL community is usually not focused on this case, except when explicitly stated such as in the works of \citet{wang2016learning, duan2016rl}.

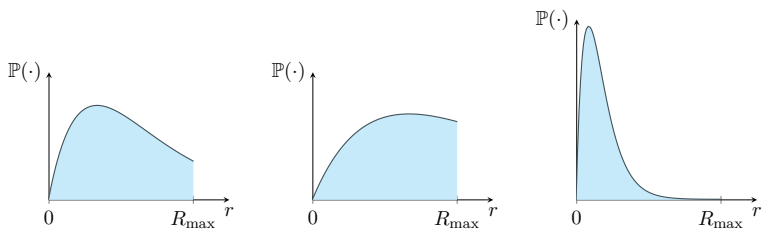
\begin{figure}[ht!]
 \centering
\scalebox{0.7}{
\begin{tikzpicture}
[	distrib/.style={rectangle,draw=black!50,fill=black!10,thick, inner sep=5pt,minimum size=4mm, align=center},
  task/.style={circle,draw=black!50,fill=black!5,thick, inner sep=5pt,minimum size=25mm, align=center},
  RLalgo/.style={rectangle,draw=black!50,fill=black!10,thick, inner sep=10pt,minimum width=30mm, align=center},
    declare function={gamma(\z)=
    0.00870357*(1/\z)^(2.5)- (174.2106599*(1/\z)^(3.5))/25920- (715.6423511*(1/\z)^(4.5))/1244160)*exp((-ln(1/\z)-1)*\z;},
    declare function={gammapdf(\x,\k,\theta) = 1/(\theta^\k)*1/(gamma(\k))*\x^(\k-1)*exp(-\x/\theta);},
    declare function={gamma2(\z)=1;},
    declare function={fct1(\x,\k,\theta) = 1/(\theta^\k))*\x^(\k-1)*exp(-1*\x/\theta);},
    declare function={fct2(\x,\k,\theta) = 1/(\theta^\k))*\x^(\k-1)*exp(-0.5*\x/\theta);},
    declare function={fct3(\x,\k,\theta) = 1/(\theta^\k))*\x^(\k-1)*exp(-4*\x/\theta);},
];

\begin{axis}[
  no markers, domain=0:6, samples=100,
  axis lines=left, xlabel=$r$, ylabel=$\mathbb P(\cdot)$,
  every axis y label/.style={at=(current axis.above origin),anchor=east},
  every axis x label/.style={at=(current axis.right of origin),anchor=north},
  height=4cm, width=5cm,
  xtick={0,6}, ytick=\empty,
  xticklabels={0,$R_{\max}$},
  xmax=7.5,
  ymax=0.25
  ]
\addplot [very thick,cyan!20!black] {fct1(x,2,2)};
\addplot [fill=cyan!20, draw=none] {fct1(x,2,2)} \closedcycle;
\end{axis}

\begin{axis}[
  no markers, domain=0:6, samples=100,
  axis lines=left, xlabel=$r$, ylabel=$\mathbb P(\cdot)$,
  every axis y label/.style={at=(current axis.above origin),anchor=east},
  every axis x label/.style={at=(current axis.right of origin),anchor=north},
  height=4cm, width=5cm,
  xtick={0,6}, ytick=\empty,
  xticklabels={0,$R_{\max}$},
  xmax=7.5,
  ymax=0.55,
  xshift=5cm
  ]
\addplot [very thick,cyan!20!black] {fct2(x,2,2)};
\addplot [fill=cyan!20, draw=none] {fct2(x,2,2)} \closedcycle;
\end{axis}

\begin{axis}[
  no markers, domain=0:6, samples=100,
  axis lines=left, xlabel=$r$, ylabel=$\mathbb P(\cdot)$,
  every axis y label/.style={at=(current axis.above origin),anchor=east},
  every axis x label/.style={at=(current axis.right of origin),anchor=north},
  height=5cm, width=5cm,
  xtick={0,6}, ytick=\empty,
  xticklabels={0,$R_{\max}$},
  xmax=7.5,
  ymax=0.048,
  xshift=10cm
  ]
\addplot [very thick,cyan!20!black] {fct3(x,2,2)};
\addplot [fill=cyan!20, draw=none] {fct3(x,2,2)} \closedcycle;
\end{axis}

  \node at (2,-0.5) (distrib1) {};
  \node at (6.5,-0.5) (distrib2) {};
  \node at (11,-0.5) (distrib3) {};


\end{tikzpicture}
}
 \caption{Illustration of the reward probabilities of 3 arms in a multi-armed bandit problem.}
 \label{fig:exploration}
\end{figure}

In the more common setting, the agent is allowed to follow a \textit{training policy} during a first phase of interactions with the environment so as to accumulate training data and hence learn a \textit{test policy}.
In the training phase, exploration is only constrained by the interactions it can make with the environment (e.g., a given number of interactions). The test policy should then be able to maximize a cumulative sum of rewards in a separate phase of interaction. The sub-optimality $\underset{s_0}{\mathbb{E}} \ V^*(s_0)-V^\pi(s_0)$ obtained in this case of setting is known as the \textit{simple regret}. 
Note that an implicit exploration/exploitation is still important.
On the one hand, the agent has to ensure that the lesser-known parts of the environment are not promising (exploration).
On the other hand, the agent is interested in gathering experience in the most promising parts of the environment (which relates to exploitation) to refine the knowledge of the dynamics.
For instance, in the bandit task provided in Figure \ref{fig:exploration}, it should be clear with only a few samples that the option on the right is less promising and the agent should gather experience mainly on the two most promising arms to be able to discriminate the best one.

\subsection{Different approaches to exploration}
The exploration techniques are split into two main categories: (i)~directed exploration and (ii)~undirected exploration \citep{thrun1992efficient}.

In the undirected exploration techniques, the agent does not rely on any exploration specific knowledge of the environment \citep{thrun1992efficient}.
For instance, the technique called $\epsilon$-greedy takes a random action with probability $\epsilon$ and follows the policy that is believed to be optimal with probability $1-\epsilon$.
Other variants such as softmax exploration (also called Boltzmann exploration) takes an action with a probability that depends on the associated expected return.

Contrary to the undirected exploration, directed exploration techniques make use of a memory of the past interactions with the environment. 
For MDPs, directed exploration can scale polynomially with the size of the state space while undirected exploration scales in general exponentially with the size of the state space (e.g., E$^3$ by \cite{kearns2002near}; R-max by \cite{brafman2003r};~...).
Inspired by the Bayesian setting, directed exploration can be done via heuristics of exploration bonus \citep{kolter2009near} or by maximizing Shannon information gains \citep[e.g.,][]{sun2011planning}.

Directed exploration is, however, not trivially applicable in high-dimensional state spaces \citep[e.g.,][]{kakade2003exploration}.
With the development of the generalization capabilities of deep learning, some possibilities have been investigated.
The key challenge is to handle, for high-dimensional spaces, the exploration/exploitation tradeoff in a principled way -- with the idea to encourage the exploration of the environment where the uncertainty due to limited data is the highest.
When rewards are not sparse,  a measure of the uncertainty on the value function can be used to drive the exploration \citep{dearden1998bayesian,dearden1999model}.
When rewards are sparse, this is even more challenging and exploration should in addition be driven by some novelty measures on the observations (or states in a Markov setting).

Before discussing the different techniques that have been proposed in the deep RL setting, one can note that the success of the first deep RL algorithms such as DQN also come from the exploration that arises naturally.
Indeed, following a simple $\epsilon$-greedy scheme online often proves to be already relatively efficient thanks to the natural instability of the Q-network that drives exploration (see Chapter \ref{ch:value-based_methods} for why there are instabilities when using bootstrapping in a fitted Q-learning algorithm with neural networks).

Different improvements are directly built on that observation. For instance, the method of "Bootstrapped DQN" \citep{osband2016deep} makes an explicit use of randomized value functions.
Along similar lines, efficient exploration has been obtained by the induced stochasticity of uncertainty estimates given by a dropout Q-network \citep{DBLP:conf/icml/GalG16} or parametric noise added to its weights \citep{lipton2016efficient,plappert2017parameter,fortunato2017noisy}.
One specificity of the work done by \cite{fortunato2017noisy} is that, similarly to Bayesian deep learning, the variance parameters are learned by gradient descent from the reinforcement learning loss function.

Another common approach is to have a directed scheme thanks to \textit{exploration rewards} given to the agent via heuristics that estimate novelty \citep{schmidhuber2010formal,stadie2015incentivizing, houthooft2016vime}.
In \citep{bellemare2016unifying,ostrovski2017count}, an algorithm provides the notion of novelty through a pseudo-count from an arbitrary density model that provides an estimate of how many times an action has been taken in similar states. This has shown good results on one of the most difficult Atari 2600 games, Montezuma's Revenge.


In \citep{florensa2017stochastic}, useful skills are learned in pre-training environments, which can then be utilized in the actual environment to improve exploration and train a high-level policy over these skills. Similarly, an agent that learns a set of auxiliary tasks may use them to efficiently explore its environment \citep{riedmiller2018learning}. These ideas are also related to the creation of options studied in \citep{machado2017laplacian}, where it is suggested that exploration may be tackled by learning options that lead to specific modifications in the state representation derived from proto-value functions. 

Exploration strategies can also make use of a model of the environment along with planning.
In that case, a strategy investigated in \citep{salge2014changing,mohamed2015variational,gregor2016variational,chiappa2017recurrent} is to have the agent choose a sequence of actions by planning that leads to a representation of state as different as possible to the current state.
In \citep{pathak2017curiosity,haber2018learning}, the agent optimizes both a model of its environment and a separate model that predicts the error/uncertainty of its own model.
The agent can thus seek to take actions that adversarially challenge its knowledge of the environment \citep{savinov2018episodic}.

By providing rewards on unfamiliar states, it is also possible to explore efficiently the environments. 
To determine the bonus, the current observation can be compared with the observations in memory.
One approach is to define the rewards based on how many environment steps it takes to reach the current observation from those in memory \citep{savinov2018episodic}.
Another approach is to use a bonus positively correlated to the error of predicting features from the observations (e.g., features given by a fixed randomly initialized neural network) \citep{burda2018exploration}.

Other approaches require either demonstrations or guidance from human demonstrators.
One line of work suggests using natural language to guide the agent by providing exploration bonuses when an instruction is correctly executed \citep{kaplan2017beating}.
In the case where demonstrations from expert agents are available, another strategy for guiding exploration in these domains is to imitate good trajectories. In some cases, it is possible to use demonstrations from experts even when they are given in an environment setup that is not exactly the same \citep{aytar2018playing}.


\section{Managing experience replay}
\label{sec:exp_replay}
In online learning, the agent has the possibility to use a replay memory \citep{lin1992self} that allows for data-efficiency by storing the past experience of the agent in order to have the opportunity to reprocess it later.
In addition, a replay memory also ensures that the mini-batch updates are done from a reasonably stable data distribution kept in the replay memory (for $N_{\text{replay}}$ sufficiently large) which helps for convergence/stability.
This approach is particularly well-suited in the case of off-policy learning as using experience from past (i.e. different) policies does not introduce any bias (usually it is even good for exploration).
In that context, methods based for instance on a DQN learning algorithm or model-based learning can safely and efficiently make use of a replay memory.
In an online setting, the replay memory keeps all information for the last $N_{\text{replay}} \in \mathbb N$\label{ntn:size_repl_mem} time steps, where $N_{\text{replay}}$ is constrained by the amount of memory available.

While a replay memory allows processing the transitions in a different order than they are experienced, there is also the possibility to use prioritized replay.
This allows for consideration of the transitions with a different frequency than they are experienced depending on their significance (that could be which experience to store and which ones to replay).
In \citep{schaul2015prioritized}, the prioritization increases with the magnitude of the transitions' TD error, with the aim that the "unexpected" transitions are replayed more often.

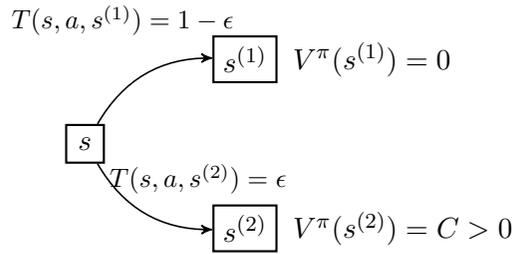
\begin{figure}[!ht]
\begin{center}
\begin{tikzpicture}[->, >=stealth', auto, semithick, node distance=3cm]
\tikzstyle{state}=[fill=white,draw=black,thick,text=black,scale=1,minimum width=0.5cm,minimum height=0.5cm]
\node[state]    (A)                     {$s$};
\node[state]    (B)[above right of=A, yshift = -1cm]   {$s^{(1)}$};
\node[state]    (C)[below right of=A, yshift = 1cm]   {$s^{(2)}$};
\path
(A) edge[bend left]     (B) node[above, yshift=1.3cm, xshift=0.5cm] {\small{$T(s,a,s^{(1)})=1-\epsilon$}}
(A) edge[bend right]    (C) node[above, yshift=-0.8cm, xshift=1.5cm] {\small{$T(s,a,s^{(2)})=\epsilon$}}

(B) node[right, xshift=0.5cm] {$V^\pi(s^{(1)})=0$}
(C) node[right, xshift=0.5cm] {$V^\pi(s^{(2)})=C>0$};
\end{tikzpicture}
\end{center}
\caption{Illustration of a state $s$ where for a given action $a$, the value of $Q^{\pi}(s,a;\theta)$ would be biased if prioritized experience replay is used ($\epsilon<<1$).}
\label{fig:prio_replay}
\end{figure}

A disadvantage of prioritized replay is that, in general, it also introduces a bias; indeed, by modifying the apparent probabilities of transitions and rewards, the expected return gets biased. This can readily be understood by considering the simple example illustrated in Figure \ref{fig:prio_replay} 
where an agent tries to estimate the expected return for a given tuple $<s,a>$.
In that example, a cumulative return of $0$ is obtained with probability $1-\epsilon$ (from next state~$s^{(1)}$) while a cumulative return of $C>0$ is obtained with probability~$\epsilon$ (from next state~$s^{(2)}$). In that case, using prioritized experience replay will bias the expected return towards a value higher than $\epsilon C$ since any transition leading to $s^{(2)}$ will be replayed with a probability higher than~$\epsilon$.

Note that this bias can be partly or completely corrected using weighted importance sampling, and this correction is important near convergence at the end of training \citep{schaul2015prioritized}.

\chapter{Benchmarking Deep RL}
\label{ch:benchmarks}
Comparing deep learning algorithms is a challenging problem due to the stochastic nature of the learning process and the narrow scope of the datasets examined during algorithm comparisons.
This problem is exacerbated in deep reinforcement learning.
Indeed, deep RL involves both stochasticity in the environment and stochasticity inherent to model learning, which makes ensuring fair comparisons and reproducibility especially difficult.
To this end, simulations of many sequential decision-making tasks have been created to serve as benchmarks.
In this section, we present several such benchmarks.
Next, we give key elements to ensure consistency and reproducibility of experimental results.
Finally, we also discuss some open-source implementations for deep RL algorithms.

\section{Benchmark Environments}

\subsection{Classic control problems}
Several classic control problems have long been used to evaluate reinforcement learning algorithms.
These include balancing a pole on a cart (Cartpole)~\citep{barto1983neuronlike}, trying to get a car up a mountain using momentum (Mountain Car)~\citep{moore1990efficient}, 
swinging a pole up using momentum and subsequently balancing it (Acrobot)~\citep{sutton1998reinforcement}. 
These problems have been commonly used as benchmarks for tabular RL and RL algorithms using linear function approximators~\citep{whiteson2011protecting}. Nonetheless, these simple environments are still sometimes used to benchmark deep RL algorithms~\citep{ho2016generative,duan2016benchmarking,lillicrap2015continuous}.

\subsection{Games}


Board-games have also been used for evaluating artificial intelligence methods for decades~\citep{shannon1950,turing1953digital,samuel1959some,sutton1988learning,littman1994markov,schraudolph1994temporal,tesauro1995temporal,campbell2002deep}.
In recent years, several notable works have stood out in using deep RL for mastering Go~\citep{silver2016mastering} or Poker~\citep{brownlibratus,moravvcik2017deepstack}.

In parallel to the achievements in board games, 
video games have also been used to further investigate reinforcement learning algorithms.
In particular,
\begin{itemize}
\item many of these games have large observation space and/or large action space; 
\item they are often non-Markovian, which require specific care (see \S\ref{sec:POMDP_meta});
\item they also usually require very long planning horizons (e.g., due to sparse rewards).
\end{itemize}

\begin{figure}[htp]
\centering
\subfloat[Space invaders]{
\includegraphics[width=.3\textwidth]{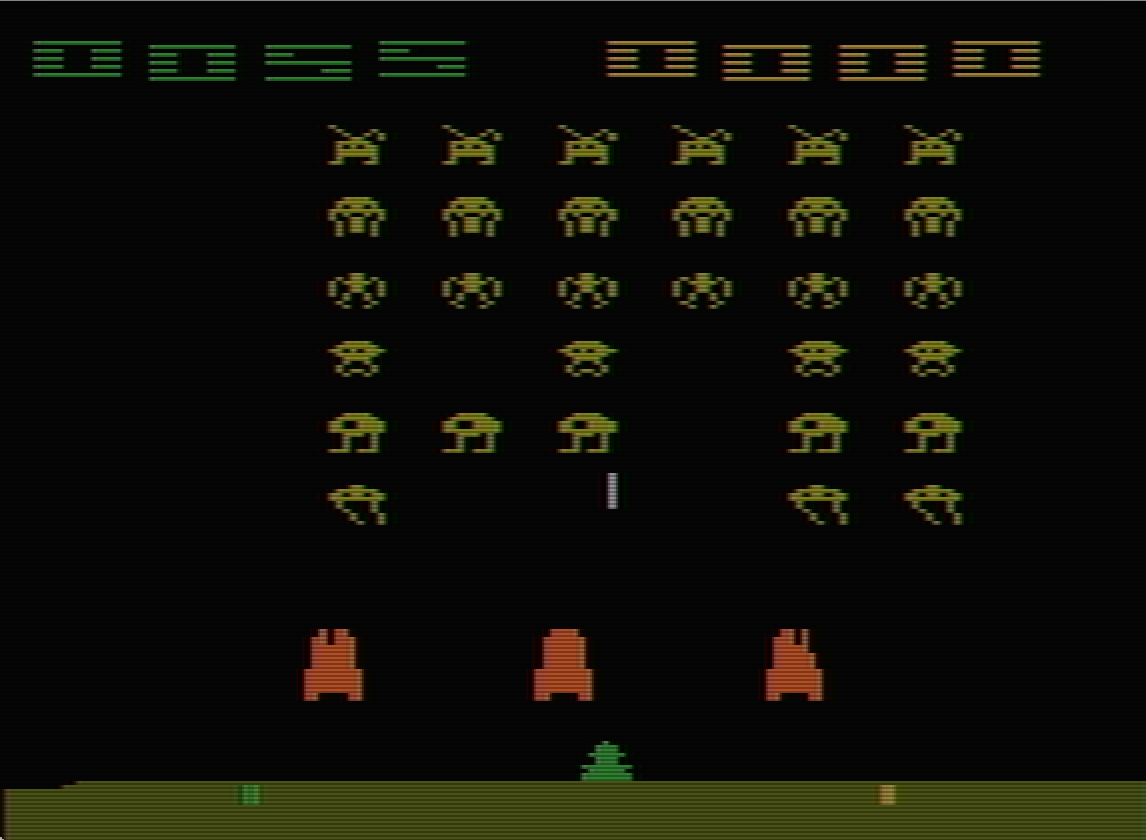}\hfill
}
\subfloat[Seaquest]{
\includegraphics[width=.3\textwidth]{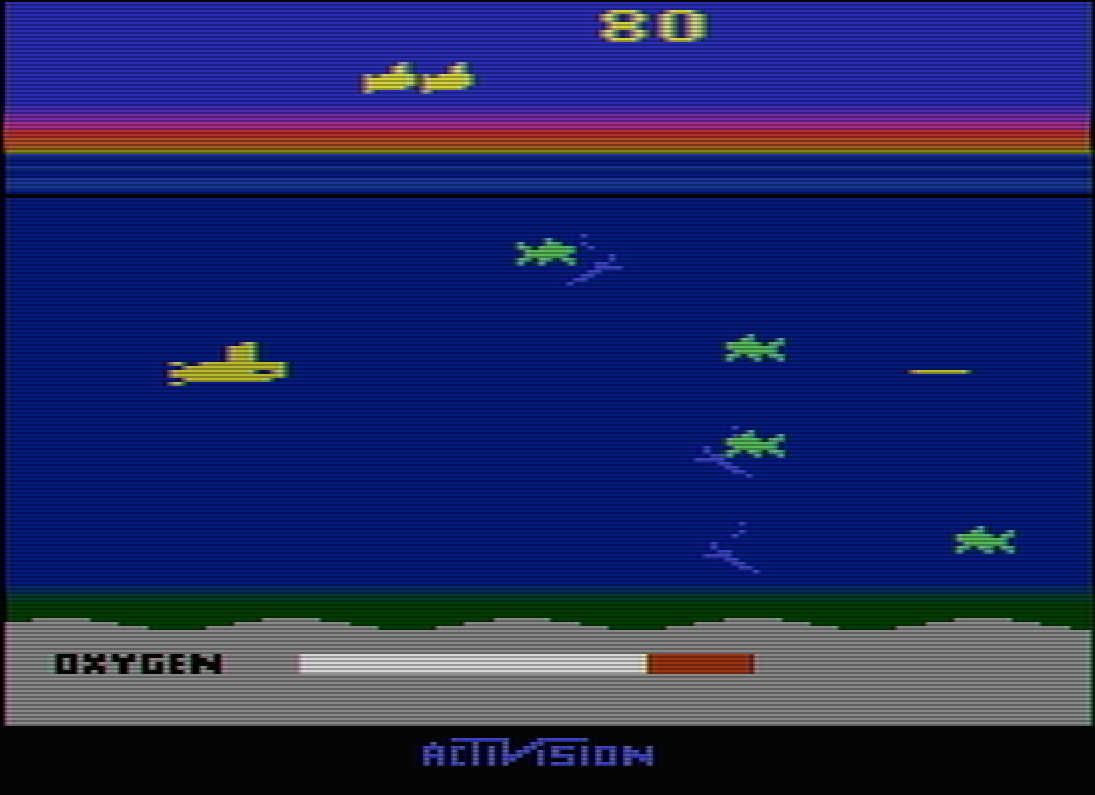}\hfill
}
\subfloat[Breakout]{
\includegraphics[width=.3\textwidth]{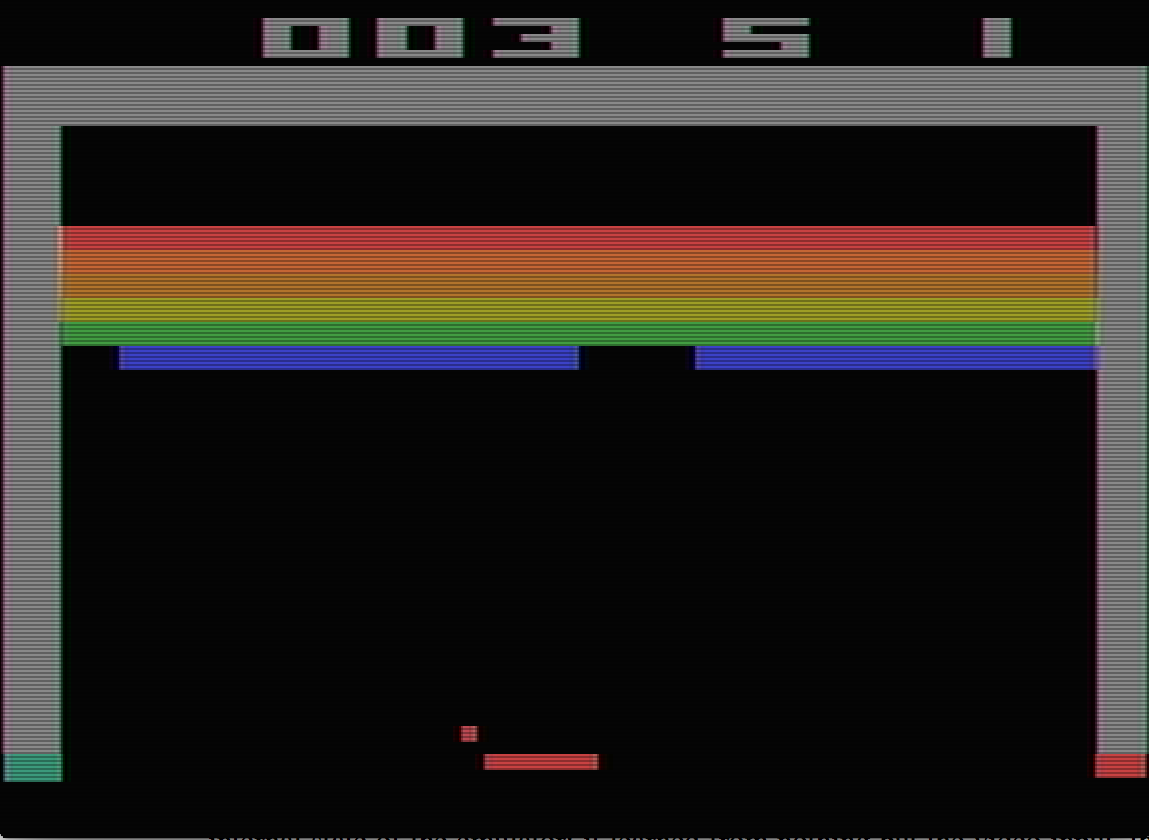}
}
\caption{Illustration of three Atari games.}
\label{fig:Atari_games}
\end{figure}

Several platforms based on video games have been popularized. 
The Arcade Learning Environment (ALE)~\citep{bellemare2013arcade} was developed to test reinforcement algorithms across a wide range of different tasks.
The system encompasses a suite of iconic Atari games, including Pong, Asteroids, Montezuma's Revenge, etc.
Figure~\ref{fig:Atari_games} shows sample frames from some of these games.
On most of the Atari games, deep RL algorithms have reached super-human level  ~\citep{mnih2015human}.
Due to the similarity in state and action spaces between different Atari games or different variants of the same game, they are also a good test-bed for evaluating generalization of reinforcement learning algorithms ~\citep{machado2017revisiting}, multi-task learning~\citep{parisotto2015actor} and for transfer learning~\citep{rusu2015policy}.

The General Video Game AI (GVGAI) competition framework \citep{perez20162014} 
was created and released with the purpose of providing researchers a platform for testing and comparing their algorithms on a large variety of games and under different constraints. The agents are required to either play multiple unknown games with or without access to game simulations, or to design new game levels or rules. 

VizDoom~\citep{kempka2016vizdoom} implements the Doom video game as a simulated environment for reinforcement learning. 
VizDoom has been used as a platform for investigations of reward shaping~\citep{lample2017playing}, curriculum learning~\citep{wu2016training}, predictive planning~\citep{dosovitskiy2016learning}, and meta-reinforcement learning~\citep{duan2016rl}.

The open-world nature of Minecraft also provides a convenient platform for exploring reinforcement learning and artificial intelligence. Project Malmo~\citep{johnson2016malmo} is a framework that provides easy access to the Minecraft video game. The environment and framework provide layers of abstraction that facilitate tasks ranging from simple navigation to collaborative problem solving. Due to the nature of the simulation, several works have also investigated lifelong-learning, curriculum learning, and hierarchical planning using Minecraft as a platform~\citep{tessler2017deep,matiisen2017teacher,branavan2012learning,oh2016control}.

Similarly, Deepmind Lab~\citep{beattie2016deepmind} provides a 3D platform adapted from the Quake video game. The Labyrinth maze environments provided with the framework have been used in work on hierarchical, lifelong and curriculum learning~\citep{jaderberg2016reinforcement, mirowski2016learning,teh2017distral}.

Finally, ``StarCraft II"~\citep{vinyals2017starcraft} and ``Starcraft: Broodwar"~\citep{wender2012applying,synnaeve2016torchcraft} provide similar benefits in exploring lifelong-learning, curriculum learning, and other related hierarchical approaches. In addition, real-time strategy (RTS) games -- as with the Starcraft series -- are also an ideal testbed for multi-agent systems. Consequently, several works have investigated these aspects in the Starcraft framework~\citep{foerster2017stabilising,peng2017multiagent,brys2014multi}.

\subsection{Continuous control systems and robotics domains}

While games provide a convenient platform for reinforcement learning, the majority of those environments investigate discrete action decisions. In many real-world systems, as in robotics, it is necessary to provide frameworks for continuous control.

In that setting, the MuJoCo~\citep{todorov2012mujoco} simulation framework is used to provide several locomotion benchmark tasks. These tasks typically involve learning a gait to move a simulated robotic agent as fast as possible. 
The action space is the amount of torque to apply to motors on the agents' joints, while the observations provided are typically the joint angles and positions in the 3D space. Several frameworks have built on top of these locomotion tasks to provide hierarchical task environments~\citep{duan2016benchmarking} and multi-task learning platforms~\citep{henderson2017multitask}.

\begin{figure}[ht!]
    \centering
    \includegraphics[width=0.6\textwidth]{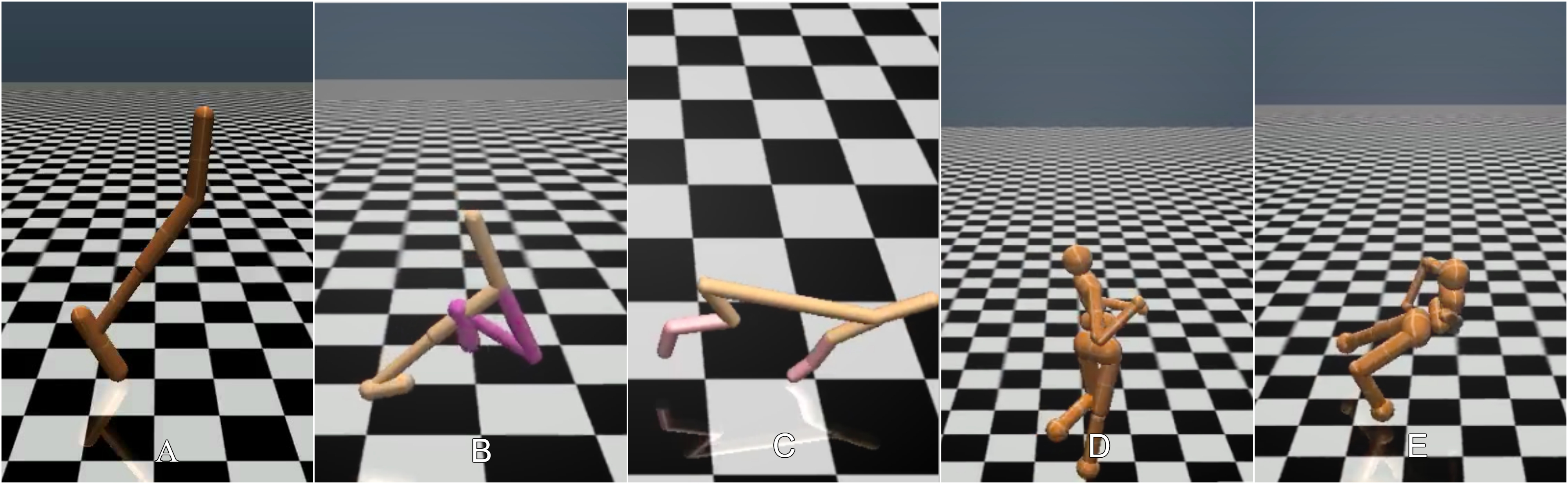}
    \caption{Screenshots from MuJoCo locomotion benchmark environments provided by OpenAI Gym.}
    \label{fig:mujoco}
\end{figure}

Because the MuJoCo simulator is closed-source and requires a license, an open-source initiative called Roboschool~\citep{schulman2017proximal} provides the same locomotion tasks along with more complex tasks involving humanoid robot simulations (such as learning to run and chase a moving flag while being hit by obstacles impeding progress).
These tasks allow for evaluation of complex planning in reinforcement learning algorithms.




Physics engines have also been used to investigate transfer learning to real-world applications.
For instance, the Bullet physics engine \citep{pybullet} has been used to learn locomotion skills in simulation, for character animation in games \citep{2017-TOG-deepLoco} or for being transferred to real robots \citep{tan2018sim}.
This also includes manipulation tasks~\citep{rusu2016sim,duan2017one} where a robotic arm stacks cubes in a given order.
Several works integrate Robot Operating System (ROS) with physics engines (such as ODE, or Bullet) to provide RL-compatible access to near real-world robotic simulations~\citep{zamora2016extending,ueno2017re}.
Most of them can also be run on real robotic systems using the same software.

There exists also a toolkit that leverages the Unity platform for creating simulation environments \citep{juliani2018unity}.
This toolkit enables the development of learning environments that are rich in sensory and physical complexity and supports the multi-agent setting.


\subsection{Frameworks}
Most of the previously cited benchmarks have open-source code available.
There also exists easy-to-use wrappers for accessing many different benchmarks.
One such example is OpenAI Gym~\citep{gym}.
This wrapper provides ready access to environments such as algorithmic, Atari, board games, Box2d games, classical control problems, MuJoCo robotics simulations, toy text problems, and others.
Gym Retro\footnote{\url{https://github.com/openai/retro}} is a wrapper similar to OpenAI Gym and it provides over 1,000 games across a variety of backing emulators. The goal is to study the ability of deep RL agents to generalize between games that have similar concepts but different appearances.
Other frameworks such as $\mu\text{niverse}$\footnote{\url{https://github.com/unixpickle/muniverse}} and SerpentAI\footnote{\url{https://github.com/SerpentAI/SerpentAI}} also provide wrappers for specific games or simulations.

\section{Best practices to benchmark deep RL}

Ensuring best practices in scientific experiments is crucial to continued scientific progress. Across various fields, investigations in reproducibility have found problems in numerous publications, resulting in several works providing experimental guidelines in proper scientific practices~\citep{sandve2013ten,baker20161,halsey2015fickle,casadevall2010reproducible}.
To this end, several works have investigated proper metrics and experimental practices when comparing deep RL algorithms~\citep{hendersonRL2017,islam2017reproducibility,machado2017revisiting,whiteson2011protecting}.

\subsubsection{Number of Trials, Random Seeds and Significance Testing}

Stochasticity plays a large role in deep RL, both from randomness within initializations of neural networks and stochasticity in environments. Results may vary significantly simply by changing the random seed. When comparing the performance of algorithms, it is therefore important to run many trials across different random seeds. 

In deep RL, it has become common to simply test an algorithm's effectiveness with an average across a few learning trials.
While this is a reasonable benchmark strategy, techniques derived from significance testing~\citep{demvsar2006statistical,bouckaert2004evaluating,bouckaert2003choosing,dietterich1998approximate} have the advantage of providing statistically grounded arguments in favor of a given hypothesis.
In practice for deep RL, significance testing can be used to take into account the standard deviation across several trials with different random seeds and environment conditions.
For instance, a simple 2-sample $t$-test can give an idea of whether performance gains are significantly due to the algorithm performance or to noisy results in highly stochastic settings. 
In particular, while several works have used the top-$K$ trials and simply presented those as performance gains, this has been argued to be inadequate for fair comparisons \citep{machado2017revisiting,hendersonRL2017}.

In addition, one should be careful not to over-interpret the results. It is possible that a hypothesis can be shown to hold for one or several given environments and under one or several given set of hyperparameters, but fail in other settings.

\subsubsection{Hyperparameter Tuning and Ablation Comparisons}

Another important consideration is ensuring a fair comparison between learning algorithms. In this case, an ablation analysis compares alternate configurations across several trials with different random seeds.
It is especially important to tune hyperparameters to the greatest extent possible for baseline algorithms. Poorly chosen hyperparameters can lead to an unfair comparison between a novel and a baseline algorithm. In particular, network architecture, learning rate, reward scale, training discount factor, and many other parameters can affect results significantly.
Ensuring that a novel algorithm is indeed performing much better requires proper scientific procedure when choosing such hyperparameters \citep{hendersonRL2017}.

\subsubsection{Reporting Results, Benchmark Environments, and Metrics}


Average returns (or cumulative reward) across evaluation trajectories are often reported as a comparison metric.
While some literature~\citep{gu2016q,gu2017interpolated} has also used metrics such as average maximum return or maximum return within $Z$ samples, these may be biased to make results for highly unstable algorithms appear more significant.
For example, if an algorithm reaches a high maximum return quickly, but then diverges, such metrics would ensure this algorithm appears successful.
When choosing metrics to report, it is important to select those that provide a fair comparison.
If the algorithm performs better in average maximum return, but worse by using an average return metric, it is important to highlight both results and describe the benefits and shortcomings of such an algorithm \citep{hendersonRL2017}.

This is also applicable to the selection of which benchmark environments to report during evaluation.
Ideally, empirical results should cover a large mixture of environments to determine in which settings an algorithm performs well and in which settings it does not.
This is vital for determining real-world performance applications and capabilities.

\section{Open-source software for Deep RL}
A deep RL agent is composed of a learning algorithm (model-based or model-free) along with specific structure(s) of function approximator(s).
In the online setting (more details are given in Chapter \ref{ch:challenges_online}), the agent follows a specific exploration/exploitation strategy and typically uses a memory of its previous experience for sample efficiency.

While many papers release implementations of various deep RL algorithms, there also exist some frameworks built to facilitate the development of new deep RL algorithms or to apply existing algorithms to a variety of environments. We provide a list of some of the existing frameworks in Appendix \ref{app:frameworks}.

\chapter{Deep reinforcement learning beyond MDPs}
\label{ch:different_settings}

We have so far mainly discussed how an agent is able to learn how to behave in a given Markovian environment where all the interesting information (the state $s_t \in \mathcal S$) is obtained at every time step $t$. 
In this chapter, we discuss more general settings with (i) non-Markovian environments, (ii) transfer learning and (iii) multi-agent systems.

\section{Partial observability and the distribution of (related) MDPs}
\label{sec:POMDP_meta}

In domains where the Markov hypothesis holds, it is straightforward to show that the policy need not depend on what happened at previous time steps to recommend an action (by definition of the Markov hypothesis). 
This section describes two different cases that complicate the Markov setting: the partially observable environments and the distribution of (related) environments.

Those two settings are at first sight quite different conceptually.
However, in both settings, at each step in the sequential decision process, the agent may benefit from taking into account its whole observable history up to the current time step $t$ when deciding what action to perform. In other words, a history of observations can be used as a pseudo-state (pseudo-state because that refers to a different and abstract stochastic control process). Any missing information in the history of observations (potentially long before time $t$) can introduce a bias in the RL algorithm (as described in Chapter \ref{ch:generalization} when some features are discarded).

\subsection{The partially observable scenario}
In this setting, the agent only receives, at each time step, an observation of its environment that does not allow it to identify the state with certainty.
A Partially Observable Markov Decision Process (POMDP) \citep{sondik1978optimal,kaelbling1998planning} is a discrete time stochastic control process defined as follows:

\begin{definition}
A POMDP is a 7-tuple $(\mathcal S,\mathcal A,T,R,\Omega,O,\gamma)$ where:
\begin{itemize}
\item $\mathcal S$ is a finite set of states $\{1, \ldots, N_{\mathcal S}\}$,
\item $\mathcal A$ is a finite set of  actions $\{1, \ldots, N_{\mathcal A}\}$,
\item $T: \mathcal S \times \mathcal A \times \mathcal S \to [0,1]$ is the transition function (set of conditional transition probabilities between states),
\item $R: \mathcal S \times \mathcal A \times \mathcal S \to \mathcal R$ is the reward function, where $\mathcal R$ is a continuous set of possible rewards in a range $R_{\text{max}} \in \mathbb{R}^+$ (e.g., $[0,R_{\text{max}}]$ without loss of generality),
\item $\Omega$ is a finite set of observations $\{1, \ldots, N_{\Omega}\}$\label{ntn:N_Omega},
\item $O: \mathcal S \times \Omega \to [0,1]$\label{ntn:cond_obs} is a set of conditional observation probabilities, and
\item $\gamma \in [0, 1)$ is the discount factor.
\end{itemize}
\end{definition}

The environment starts in a distribution of initial states $b(s_0)$.
At each time step $t \in \mathbb N_0$, the environment is in a state $s_t \in \mathcal S$. At the same time, the agent receives an observation $\omega_t \in \Omega$ that depends on the state of the environment with probability $O(s_t, \omega_t)$, after which the agent chooses an action $a_t \in \mathcal A$.
Then, the environment transitions to state $s_{t+1} \in \mathcal S$ with probability $T(s_t,a_t,s_{t+1})$ and the agent receives a reward  $r_t \in \mathcal R$ equal to $R(s_t, a_t, s_{t+1})$.

When the full model ($T$, $R$ and $O$) are known, methods such as Point-Based Value Iteration (PBVI) algorithm \citep{pineau2003point} for POMDP planning can be used to solve the problem. If the full POMDP model is not available, other reinforcement learning techniques have to be used.

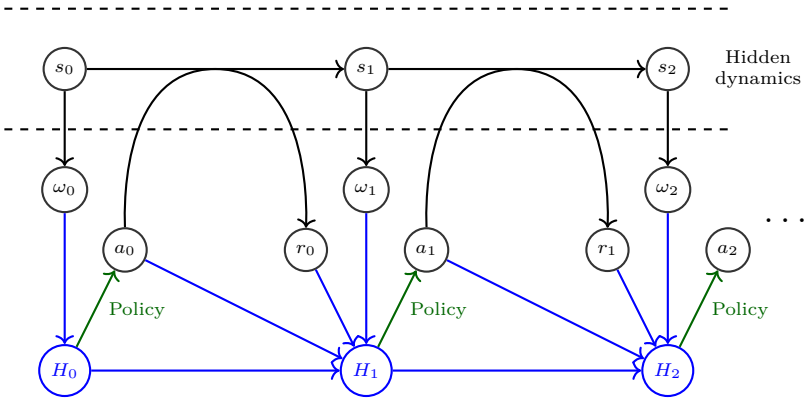
\begin{figure}[ht!]
\centering
\def\dist{5}
\def\ra{0.9}
\begin{tikzpicture}[->,thick, scale=0.8]
\scriptsize
\tikzstyle{main}=[circle, minimum size = \ra, thick, draw =black!80, node distance = 12mm]
\tikzstyle{rr}=[rounded rectangle, rounded rectangle west arc=5pt, rounded rectangle east arc=50pt, minimum size = 7mm, thick, draw =black!80, node distance = 12mm]

\foreach \name in {0,...,2}
    \node[main, fill = white!100] (s\name) at (\dist*\name,0) {$s_\name$};
\foreach \name in {0,...,2}
    \node[main, fill = white!100] (omega\name) at (\dist*\name,-2) {$\omega_\name$};
\foreach \name in {0,...,2}
    \node[main, color=blue] (H\name) at (\dist*\name,-5) {$H_\name$};
\foreach \name in {0,...,2}
    \node[main, fill = white!100] (a\name) at (\dist*\name+1,-3) {$a_\name$};
\foreach \name in {0,...,1}
    \node[main, fill = white!100] (r\name) at (\dist*\name+4,-3) {$r_\name$};
\foreach \name in {0,...,1}{
    \node[] (tb\name) at (\dist*\name+2,0) {};
    \node[] (tf\name) at (\dist*\name+3,0) {};
}

\node[font=\Large] (dots) at (\dist*2+2,-2.5) {\ldots};

\draw [dashed,-] (-1,1) -- (11,1);
\draw [dashed,-] (-1,-1) -- (11,-1);

\foreach \name in {0,...,1}
    \draw [] plot [smooth, tension=2] coordinates { (a\name.north) (\dist*\name+\dist/2,0) (r\name.north) };

\foreach \cur/\next in {0/1,1/2}
       {
        \path (s\cur) edge (s\next);
        \path (s\cur) edge (omega\cur);
        \path [blue] (omega\cur) edge (H\cur);
        \path [blue] (H\cur) edge (H\next);
        \path [black!60!green] (H\cur) edge (a\cur);
        \node [black!60!green] (pol\cur) at (\dist*\cur+1.2,-4) {Policy};
        \path [blue] (r\cur) edge (H\next);
        \path [blue] (a\cur) edge (H\next);
       }
        \path (s2) edge (omega2);
        \path [blue] (omega2) edge (H2);
        \node [text width=2cm, align=center] (hid) at (11.5,0) {Hidden dynamics};

        \path [black!60!green] (H2) edge (a2);
        \node [black!60!green] (pol2) at (\dist*2+1.2,-4) {Policy};

\end{tikzpicture}
\caption{Illustration of a POMDP. The actual dynamics of the POMDP is depicted in dark while the information that the agent can use to select the action at each step is the whole history $H_t$ depicted in blue.}
\label{fig:POMDP}
\end{figure}

A naive approach to building a space of candidate policies is to consider the set of mappings taking only the very last observation(s) as input. However, in a POMDP setting, this leads to candidate policies that are typically not rich enough to capture the system dynamics, thus suboptimal. 
In that case, the best achievable policy is stochastic \citep{singh1994learning}, and it can be obtained using policy gradient.
The alternative is to use a history of previously observed features to better estimate the hidden state dynamics.
We denote by $\mathcal H_t=\Omega \times (\mathcal A \times \mathcal R \times \Omega)^{t}$ the set of histories observed up to time $t$ for $t \in \mathbb N_0$ (see Fig. \ref{fig:POMDP}),  and by  $\mathcal H=\bigcup\limits_{t=0}^{\infty} \mathcal H_{t}$ the space of all possible observable histories.

A straightforward approach is to take the whole history $H_{t} \in \mathcal H$ as input \citep{braziunas2003pomdp}.
However, increasing the size of the set of candidate optimal policies generally implies:
(i) more computation to search within this set \citep{singh1994learning, mccallum1996reinforcement} and,
(ii) an increased risk of including candidate policies suffering from overfitting due to lack of sufficient data, which thus leads to a bias-overfitting tradeoff when learning policies from data \citep{franccois2017on}.

In the case of deep RL, the architectures used usually have a smaller number of parameters and layers than in supervised learning due to the more complex RL setting, but the trend of using ever smarter and complex architectures in deep RL happens similarly to supervised learning tasks.
Architectures such as convolutional layers or recurrency are particularly well-suited to deal with a large input space 
because they offer interesting generalization properties. 
A few empirical successes on large scale POMDPs make use of convolutional layers \citep{mnih2015human} and/or recurrent layers \citep{hausknecht2015deep}, such as LSTMs \citep{hochreiter1997long}. 

\subsection{The distribution of (related) environments}
\label{sec:distrib_envs}
\begin{figure}[ht!]
\tikzstyle{arr} = [
        -triangle 90,
        line width=0.05cm,
        postaction={draw, line width=0.2cm, shorten >=0.2cm, -}
]

\newsavebox{\genericfilt}
\savebox{\genericfilt}{%
    \begin{tikzpicture}[font=\small,
            >=stealth,
        ]
        \draw[thick, blue] (0,0) --(0.2,0) ..controls (0.5,1)..(1,1)..controls(1.2,1) and (1.8,0)..
            node[black,left]{} (2,0) -- (2.5,0);
        \draw (1,0) node[below]{Distribution of tasks};
    \end{tikzpicture}%
}

\newsavebox{\labya}
\savebox{\labya}{%
\scalebox{0.6}{
    \begin{tikzpicture}[font=\small,
            >=stealth,
        ]
\draw[very thick, fill=white, white] (0,0) rectangle (3,4);
\draw[thin, black!80] (0,0) grid (3,4);

\foreach \x/\y in {0/2, 1/2, 2/2, 0/3, 1/3, 2/3,0/1,2/1}
  \draw[fill=black!50] (\x,\y) rectangle (\x+1,\y+1);
\foreach \x/\y in {0/1, 1/1, 1/2}
  \draw[pattern=north west lines, pattern color=blue] (\x,\y) rectangle (\x+1,\y+1);

\draw[thin, black, line width=1mm] (0,0) rectangle (3,4);
\draw[very thick, fill=white, white] (-0.02,1) rectangle (0.0,2);

\node at (1.5,0.5) {\pgftext{\includegraphics[width=24pt]{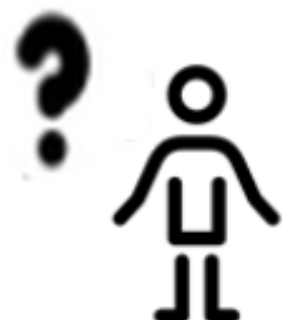}}};
\draw[thin, black, line width=1mm] (0,0) rectangle (3,4);
\draw[line width=1mm, black!10] (-0.01,2) rectangle (0.,3);
\draw[arr]  (0.2,2.5) -- (-0.7,2.5) ;

    \end{tikzpicture}%
}
}

\newsavebox{\labyb}
\savebox{\labyb}{%
\scalebox{0.6}{
    \begin{tikzpicture}[font=\small,
            >=stealth,
        ]
\draw[very thick, fill=white, white] (0,0) rectangle (3,4);
\draw[thin, black!80] (0,0) grid (3,4);

\foreach \x/\y in {0/2, 1/2, 2/2, 0/3, 1/3, 2/3,0/1,2/1}
  \draw[fill=black!50] (\x,\y) rectangle (\x+1,\y+1);
\foreach \x/\y in {2/2, 1/1, 1/3}
  \draw[pattern=north west lines, pattern color=blue] (\x,\y) rectangle (\x+1,\y+1);

\node at (1.5,0.5) {\pgftext{\includegraphics[width=24pt]{lost.pdf}}};
\draw[thin, black, line width=1mm] (0,0) rectangle (3,4);
\draw[line width=1mm, black!10] (-0.01,3) rectangle (0.,4);
\draw[arr]  (0.2,3.5) -- (-0.7,3.5) ;
    \end{tikzpicture}%
}
}

\newsavebox{\labyc}
\savebox{\labyc}{%
\scalebox{0.6}{
    \begin{tikzpicture}[font=\small,
            >=stealth,
        ]
\draw[very thick, fill=white, white] (0,0) rectangle (3,4);
\draw[thin, black!80] (0,0) grid (3,4);

\foreach \x/\y in {0/0, 0/1, 0/2, 0/3, 1/0, 2/0,1/1,1/3}
  \draw[fill=black!50] (\x,\y) rectangle (\x+1,\y+1);
\foreach \x/\y in {1/1, 1/2,0/3}
  \draw[pattern=north west lines, pattern color=blue] (\x,\y) rectangle (\x+1,\y+1);

\node at (2.5,2.5) {\pgftext{\includegraphics[width=24pt]{lost.pdf}}};
\draw[thin, black, line width=1mm] (0,0) rectangle (3,4);
\draw[line width=1mm, black!10] (-0.02,2) rectangle (0.0,3);
\draw[arr]  (0.2,2.5) -- (-0.7,2.5) ;
    \end{tikzpicture}%
}
}

\newsavebox{\labytest}
\savebox{\labytest}{%
\scalebox{0.6}{
    \begin{tikzpicture}[font=\small,
            >=stealth,
        ]
\draw[very thick, fill=white, white] (0,0) rectangle (3,4);
\draw[thin, black!80] (0,0) grid (3,4);

\foreach \x/\y in {0/0, 0/1, 0/2, 0/3, 1/1, 1/2,1/3,2/2,2/3}
  \draw[fill=black!50] (\x,\y) rectangle (\x+1,\y+1);
\foreach \x/\y in {1/1, 1/2,1/3}
  \draw[pattern=north west lines, pattern color=blue] (\x,\y) rectangle (\x+1,\y+1);

\node at (2.5,0.5) {\pgftext{\includegraphics[width=24pt]{lost.pdf}}};
\draw[thin, black, line width=1mm] (0,0) rectangle (3,4);
\draw[line width=1mm, black!10] (-0.02,2) rectangle (0.0,3);
\draw[arr]  (0.2,2.5) -- (-0.7,2.5) ;

    \end{tikzpicture}%
}
}

 \centering
\scalebox{0.65}{
\begin{tikzpicture}
[	distrib/.style={rectangle,draw=black!50,fill=black!10,thick, inner sep=5pt,minimum size=4mm, align=center},
  task/.style={circle,draw=black!50,fill=black!5,thick, inner sep=5pt,minimum size=25mm, align=center},
  RLalgo/.style={rectangle,draw=black!50,fill=black!10,thick, inner sep=10pt,minimum width=30mm, align=center},
];

  \node at (0,0) 	[distrib] (distrib) {\usebox{\genericfilt}};
  \node at (-4.5,-4) [task] (task1) {\usebox{\labya}};
  \node at (0,-4) 	[task] (task2) {\usebox{\labyb}};
  \node at (4.5,-4) 	[task] (task3) {\usebox{\labyc}};

  \node at (0,-8) [RLalgo] (algo) {RL algorithm};

  \node at (0,-12) [task] (test_task) {\usebox{\labytest}};
  \node at (-7,0)[inner sep=0,minimum size=0] (node1) {};
  \node at (-7,-7)[inner sep=0,minimum size=0] (node2) {};

 \draw [->] (distrib) -- (task1);
 \draw [->] (distrib) -- (task2);
 \draw [->] (distrib) -- (task3);

 \draw [->] (task1) -- (algo);
 \draw [->] (task2) -- (algo);
 \draw [->] (task3) -- (algo);

 \draw [->] (algo) -- (test_task);
 \draw [->] (distrib) -- (node1) -- (node2) -- (test_task);

\draw [decorate,decoration={brace,amplitude=10pt,raise=4pt},yshift=0pt]
(7,0) -- (7,-7.75) node [black,midway,xshift=1.5cm, minimum width=3cm, align=center] {Training \\on a set \\of tasks};

\draw [decorate,decoration={brace,amplitude=10pt,raise=4pt},yshift=0pt]
(7,-8.25) -- (7,-14) node [black,midway,xshift=1.5cm, minimum width=3cm, align=center] {Testing \\on related \\ tasks};

\end{tikzpicture}
}
 \caption{Illustration of the general setting of meta learning on POMDPs for a set of labyrinth tasks. In this illustration, it is supposed that the agent only sees the nature of the environment just one time step away from him.}
 \label{fig:distrib_envs}
\end{figure}

In this setting, the environment of the agent is a distribution of different (yet related) tasks that differ for instance in the reward function or in the probabilities of transitions from one state to another. 
Each task $T_i \sim \mathcal T$ can be defined by the observations $\omega_t \in \Omega$ (which are equal to $s_t$ if the environments are Markov), the rewards $r_t \in \mathcal R$, as well as the effect of the actions $a_t \in \mathcal A$ taken at each step. Similarly to the partially observable context, we denote the history of observations by $H_t$, where $H_t \in \mathcal H_t=\Omega \times (\mathcal A \times \mathcal R \times \Omega)^{t}$. The agent aims at finding a policy $\pi(a_t | H_t; \theta)$ with the objective of maximizing its expected return, defined (in the discounted setting) as
$$\underset{T_i \sim \mathcal T}{\mathbb{E}} \left[ \sum \nolimits_{k=0}^{\infty} \gamma^{k} r_{t+k} \mid H_t, \pi  \right].$$
An illustration of the general setting of meta learning on non-Markov environments is given in Figure \ref{fig:distrib_envs}.

Different approaches have been investigated in the literature.
The Bayesian approach aims at explicitly modeling the distribution of the different environments, if a prior is available \citep{ghavamzadeh2015bayesian}. However, it is often intractable to compute the Bayesian-optimal strategy and one has to rely on more practical approaches that do not require an explicit model of the distribution. 
The concept of \textit{meta-learning} or \textit{learning to learn} aims at discovering, from experience, how to behave in a range of tasks and how to negotiate the exploration-exploitation tradeoff \citep{hochreiter2001learning}.
In that case, deep RL techniques have been investigated by, e.g., \cite{wang2016learning, duan2016rl} with the idea of using recurrent networks trained on a set of environments drawn i.i.d. from the distribution. 

Some other approaches have also been investigated.
One possibility is to train a neural network to imitate the behavior of known optimal policies on MDPs drawn from the distribution \citep{castronovo2017approximate}.
The parameters of the model can also be explicitly trained such that a small number of gradient steps in a new task from the distribution will produce fast learning on that task \citep{finn2017model}.

There exists different denominations for this setting with a distribution of environments. 
For instance, the denomination of "multi-task setting" is usually used in settings where a short history of observations is sufficient to clearly distinguish the tasks.
As an example of a multi-task setting, a deep RL agent can exceed median human performance on the set of 57 Atari games with a single set of weights \citep{hessel2018multi}.
Other related denominations are the concepts of "contextual" policies \citep{da2012learning} and "universal"/"goal conditioned" value functions \citep{schaul2015universal} that refer to learning policies or value function within the same dynamics for multiple goals (multiple reward functions).

\section{Transfer learning}
\label{sec:transfer}

Transfer learning is the task of efficiently using previous knowledge from a source environment to achieve a good performance in a target environment. In a transfer learning setting, the target environment should not be in the distribution of the source tasks. However, in practice, the concept of transfer learning is sometimes closely related to meta learning, as we discuss hereafter.


\subsection{Zero-shot learning}
The idea of zero-shot learning is that an agent should be able to act appropriately in a new task directly from experience acquired on other similar tasks.
For instance, one use case is to learn a policy in a simulation environment and then use it in a real-world context where gathering experience is not possible or severely constrained (see \S\ref{sec:challenges_real-world}).
To achieve this, the agent must either (i)~develop generalization capacities described in Chapter \ref{ch:generalization} or (ii)~use specific transfer strategies that explicitly retrain or replace some of its components to adjust to new tasks. 

To develop generalization capacities, one approach is to use an idea similar to data augmentation in supervised learning so as to make sense of variations that were not encountered in the training data.
Exactly as in the meta-learning setting (\S\ref{sec:distrib_envs}), the actual (unseen) task may appear to the agent as just another variation if there is enough data augmentation on the training data.
For instance, the agent can be trained with deep RL techniques on different tasks simultaneously, and it is shown by \cite{parisotto2015actor} that it can generalize to new related domains where the exact state representation has never been observed.
Similarly, the agent can be trained in a simulated environment while being provided with different renderings of the observations. In  that case, the learned policy can transfer well to real images \citep{sadeghi2016cad2rl,tobin2017domain}.
The underlying reason for these successes is the ability of the deep learning architecture to generalize between states that have similar high-level representations and should therefore have the same value function/policy in different domains.
Rather than manually tuning the randomization of simulations, one can also adapt the simulation parameters by matching the policy behavior in simulation to the real world \citep{chebotar2018closing}.
Another approach to zero-shot transfer is to use algorithms that enforce states that relate to the same underlying task but have different renderings to be mapped into an abstract state that is close \citep{tzeng2015adapting,franccois2018combined}.

\subsection{Lifelong learning or continual learning}
A specific way of achieving transfer learning is to aim at \textit{lifelong learning} or \textit{continual learning}.
According to \cite{silver2013lifelong}, lifelong machine learning relates to the capability of a system to learn many tasks over a lifetime from one or more domains.

In general, deep learning architectures can generalize knowledge across multiple tasks by sharing network parameters.
A direct approach is thus to train function approximators (e.g. policy, value function, model, etc.) sequentially in different environments.
The difficulty of this approach is to find methods that enable the agent to retain knowledge in order to more efficiently learn new tasks.
The problem of retaining knowledge in deep reinforcement learning is complicated by the phenomenon of catastrophic forgetting, where generalization to previously seen data is lost at later stages of learning.

The straightforward approach is to either (i)~use experience replay from all previous experience (as discussed in \S\ref{sec:exp_replay}), or (ii)~retrain occasionally on previous tasks similar to the meta-learning setting (as discussed in \S\ref{sec:distrib_envs}).

When these two options are not available, or as a complement to the two previous approaches, one can use deep learning techniques that are robust to forgetting, such as progressive networks \citep{rusu2016sim}.
The idea is to leverage prior knowledge by adding, for each new task, lateral connections to previously learned features (that are kept fixed).
Other approaches to limiting catastrophic forgetting include slowing down learning on the weights important for previous tasks \citep{kirkpatrick2016overcoming} and decomposing learning into skill hierarchies \citep{stone2000layered,tessler2017deep}.

\begin{figure}[ht!]

\tikzstyle{arr} = [
        -triangle 90,
        line width=0.05cm,
        postaction={draw, line width=0.2cm, shorten >=0.2cm, -}
]

\centering
\def\dist{2}
\def\ra{0.9}
\begin{tikzpicture}[->,thick, scale=0.8]
\scriptsize
\tikzstyle{main}=[circle, minimum size = \ra, thick, draw =black!80, node distance = 12mm, fill = black!10]
\tikzstyle{rr}=[rounded rectangle, rounded rectangle west arc=5pt, rounded rectangle east arc=50pt, minimum size = 7mm, thick, draw =black!80, node distance = 12mm]

\draw[arr]  (5,-3.) -- (6,-3.) ;

\foreach \name in {0,...,4}
    \node[main] (t\name) at (\dist*\name,0) {Task \name};

\node[font=\Large] (dots) at (\dist*4+2,0) {\dots};
\node[font=\Large] (agent) at (\dist*2,-3) {Agent};

\draw [dashed,-] (-1,1) -- (11,1);
\draw [dashed,-] (-1,-1) -- (11,-1);

\node [text width=2cm, align=center] (hid) at (11.5,0) {Sequence of tasks};

\draw[-open triangle 45] (t2.260) -- node[rotate=60,above] {} (agent.105);
\draw[open triangle 45-] (t2.280) -- node[rotate=60,below] {} (agent.75);

\end{tikzpicture}
\caption{Illustration of the continual learning setting where an agent has to interact sequentially with related (but different) tasks.}
\label{fig:continual}
\end{figure}
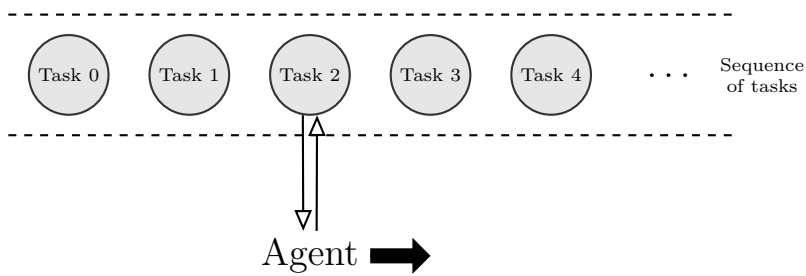

\subsection{Curriculum learning}
\label{sec:curriculum}
A particular setting of continual learning is curriculum learning.
Here, the goal is to explicitly design a sequence of source tasks for an agent to train on such that the final performance or learning speed is improved on a target task.
The idea is to start by learning small and easy aspects of the target task and then to gradually increase the difficulty level \citep{bengio2009curriculum, narvekar2016source}. 
For instance, \citet{florensa2018automatic} use generative adversarial training to automatically generate goals for a contextual policy such that they are always at the appropriate level of difficulty.
As the difficulty and number of tasks increase, one possibility to satisfy the bias-overfitting tradeoff is to consider network transformations through learning.

\section{Learning without explicit reward function}
In reinforcement learning, the reward function defines the goals to be achieved by the agent (for a given environment and a given discount factor).
Due to the complexity of environments in practical applications, defining a reward function can turn out to be rather complicated.
There are two other possibilities: (i) given demonstrations of the desired task, we can use imitation learning or extract a reward function using inverse reinforcement learning; (ii) a human may provide feedback on the agent's behavior in order to define the task.

\subsection{Learning from demonstrations}
In some circumstances, the agent is only provided with trajectories of an expert agent (also called the teacher), without rewards. 
Given an observed behavior, the goal is to have the agent perform similarly. Two approaches are possible:
\begin{itemize}
\item \underline{Imitation learning} uses supervised learning to map states to actions from the observations of the expert's behavior \citep[e.g.,][]{giusti2016machine}. Among other applications, this approach has been used for self-driving cars to map raw pixels directly to steering commands thanks to a deep neural network \citep{bojarski2016end}.

\item \underline{Inverse reinforcement learning (IRL)} determines a possible reward function given observations of optimal behavior. 
When the system dynamics is known (except the reward function), this is an appealing approach particularly when the reward function provides the most generalizable definition of the task \citep{ng2000algorithms,abbeel2004apprenticeship}. For example, let us consider a large MDP for which the expert always ends up transitioning to the same state. In that context, one may be able to easily infer, from only a few trajectories, what the probable goal of the task is (a reward function that explains the behavior of the teacher), as opposed to directly learning the policy via imitation learning, which is much less efficient.
Note that recent works bypass the requirement of the knowledge of the system dynamics \citep{boularias2011relative,kalakrishnan2013learning,finn2016guided} by using techniques based on the principle of maximum causal entropy \citep{ziebart2010modeling}.
\end{itemize}
A combination of the two approaches has also been investigated by \cite{neu2012apprenticeship,ho2016generative}.
In particular, \cite{ho2016generative} use adversarial methods to learn a discriminator (i.e., a reward function) such that the policy matches the distribution of demonstrative samples.

It is important to note that in many real-world applications, the teacher is not exactly in the same context as the agent. Thus, transfer learning may also be of crucial importance \citep{schulman2016learning, liu2017imitation}. 

Another setting requires the agent to learn directly from a sequence of observations without corresponding actions (and possibly in a slightly different context). This may be done in a meta-learning setting by providing positive reward to the agent when it performs as it is expected based on the demonstration of the teacher. The agent can then act based on new unseen trajectories of the teacher, with the objective that is can generalize sufficiently well to perform new tasks \citep{paine2018one}.

\subsection{Learning from direct feedback}
Learning from feedback investigates how an agent can interactively learn behaviors from a human teacher who provides positive and negative feedback signals. In order to learn complex behavior, human trainer feedbacks has the potential to be more performant than a reward function defined a priori \citep{macglashan2017interactive, warnell2017deep}. This setting can be related to the idea of curriculum learning discussed in \S\ref{sec:curriculum}.

In the work of \cite{hadfield2016cooperative}, the cooperative inverse reinforcement learning framework considers a two-player game between a human and a robot interacting with an environment with the purpose of maximizing the human's reward function.
In the work of \cite{christiano2017deep}, it is shown how learning a separate reward model using supervised learning lets us significantly reduce the amount of feedback required from a human teacher. They also present the first practical applications of using human feedback in the context of deep RL to solve tasks with a high dimensional observation space.

\section{Multi-agent systems}
A multi-agent system is composed of multiple interacting agents within an environment \citep{littman1994markov}.

\begin{definition}
A multi-agent POMDP with N agents is a tuple $(\mathcal S,\mathcal A_N, \ldots, \mathcal A_N,T,R_1,\ldots,R_N,\Omega,O_1, \ldots, O_N,\gamma)$ where:
\begin{itemize}
\item $\mathcal S$ is a finite set of states $\{1, \ldots, N_{\mathcal S}\}$ (describing the possible configurations of all agents),
\item $\mathcal A=\mathcal A_1 \times \ldots \times \mathcal A_n$ is a finite set of  actions $\{1, \ldots, N_{\mathcal A}\}$,
\item $T: \mathcal S \times \mathcal A \times \mathcal S \to [0,1]$ is the transition function (set of conditional transition probabilities between states),
\item $\forall i, R_i: \mathcal S \times \mathcal A_i \times \mathcal S \to \mathcal R$ is the reward function for agent $i$, where $\mathcal R$ is a continuous set of possible rewards in a range $R_{\text{max}} \in \mathbb{R}^+$ (e.g., $[0,R_{\text{max}}]$ without loss of generality),
\item $\Omega$ is a finite set of observations $\{1, \ldots, N_{\Omega}\}$,
\item $\forall i, O_i: \mathcal S \times \Omega \to [0,1]$ is a set of conditional observation probabilities, and
\item $\gamma \in [0, 1)$ is the discount factor.
\end{itemize}
\end{definition}

For this type of system, many different settings can be considered.

\begin{itemize}
\item \underline{Collaborative versus non-collaborative setting}.
In a pure collaborative setting, agents have a shared reward measurement ($R_i=R_j, \forall~i,j~\in~[1, \ldots, N]$).
In a mixed or non-collaborative (possibly adversarial) setting each agent obtains different rewards.
In both cases, each agent $i$ aims to maximize a discounted sum of its rewards $\sum_{t=0}^H \gamma^t r_t^{(i)}$.
\item \underline{Decentralized versus centralized setting}. 
In a decentralized setting, each agent selects its own action conditioned only on its local information.
When collaboration is beneficial, this decentralized setting can lead to the emergence of communication between agents in order to share information \citep[e.g.,][]{sukhbaatar2016learning}.

In a centralized setting, the RL algorithm has access to all observations $w^{(i)}$ and all rewards $r^{(i)}$.
The problem can be reduced to a single-agent RL problem on the condition that a single objective can be defined (in a purely collaborative setting, the unique objective is straightforward). 
Note that even when a centralized approach can be considered (depending on the problem), an architecture that does not make use of the multi-agent structure usually leads to sub-optimal learning \citep[e.g.,][]{sunehag2017value}.
\end{itemize}

In general, multi-agent systems are challenging because agents are independently updating their policies as learning progresses, and therefore the environment appears non-stationary to any particular agent.
For training one particular agent, one approach is to select randomly the policies of all other agents from a pool of previously learned policies. 
This can stabilize training of the agent that is learning and prevent overfitting to the current policy of the other agents \citep{silver2016mastering}.

In addition, from the perspective of a given agent, the environment is usually strongly stochastic even with a known, fixed policy for all other agents.
Indeed, any given agent does not know how the other agents will act and consequently, it doesn't know how its own actions contribute to the rewards it obtains.
This can be partly explained due to partial observability, and partly due to the intrinsic stochasticity of the policies followed by other agents (e.g., when there is a high level of exploration).
For these reasons, a high variance of the expected global return is observed, which makes learning challenging (particularly when used in conjunction with bootstrapping).
In the context of the collaborative setting, a common approach is to use an actor-critic architecture with a centralized critic during learning and decentralized actor (the agents can be deployed independently).
These topics have already been investigated in works by \cite{foerster2017counterfactual, sunehag2017value,lowe2017multi} as well as in the related work discussed in these papers.
Other works have shown how it is possible to take into account a term that either considers the learning of the other agent \citep{foerster2018learning} or an internal model that predicts the actions of other agents \citep{jaques2018intrinsic}.

Deep RL agents are able to achieve human-level performance in 3D multi-player first-person video games such as Quake III Arena Capture the Flag \citep{jaderberg2018human}.
Thus, techniques from deep RL have a large potential for many real-world tasks that require multiple agents to cooperate in domains such as robotics, self-driving cars, etc.

\chapter{Perspectives on deep reinforcement learning}
\label{ch:real-world}
In this section, we first mention some of the main successes of deep RL.
Then, we describe some of the main challenges for tackling an even wider range of real-world problems.
Finally, we discuss some parallels that can be found between deep RL and neuroscience.

\section{Successes of deep reinforcement learning}
Deep RL techniques have demonstrated their ability to tackle a wide range of problems that were previously unsolved. Some of the most renowned achievements are

\begin{itemize}
\item  beating previous computer programs in the game of backgammon \citep{tesauro1995temporal},
\item attaining superhuman-level performance in playing Atari games from the pixels \citep{mnih2015human},
\item mastering the game of Go \citep{silver2016mastering}, as well as
\item beating professional poker players in the game of heads up no-limit Texas hold'em: Libratus \citep{brownlibratus} and Deepstack \citep{moravvcik2017deepstack}.
\end{itemize}

These achievements in popular games are important because they show the potential of deep RL in a variety of complex and diverse tasks that require working with high-dimensional inputs.
Deep RL has also shown lots of potential for real-world applications such as
robotics \citep{kalashnikov2018qt},
self-driving cars \citep{you2017virtual},
finance \citep{deng2017deep},
smart grids \citep{franccois2016deep},
dialogue systems \citep{fazel2017learning},
etc.
In fact, Deep RL systems are already in production environments.
For example, \citet{gauci2018horizon} describe how Facebook uses Deep RL such as for pushing notifications and for faster video loading with smart prefetching.

RL is also applicable to fields where one could think that supervised learning alone is sufficient, such as sequence prediction \citep{ranzato2015sequence,bahdanau2016actor}. Designing the right neural architecture for supervised learning tasks has also been cast as an RL problem \citep{zoph2016neural}. Note that those types of tasks can also be tackled with evolutionary strategies \citep{miikkulainen2017evolving, real2017large}.

Finally, it should be mentioned that deep RL has applications in classic and fundamental algorithmic problems in the field of computer science, such as the travelling salesman problem \citep{bello2016neural}. This is an NP-complete problem and the possibility to tackle it with deep RL shows the potential impact that it could have on several other NP-complete problems, on the condition that the structure of these problems can be exploited.


\section{Challenges of applying reinforcement learning to real-world problems}
\label{sec:challenges_real-world}
The algorithms discussed in this introduction to deep RL can, in principle, be used to solve many different types of real-world problems. In practice, even in the case where the task is well defined (explicit reward function), there is one fundamental difficulty: it is often not possible to let an agent interact freely and sufficiently in the actual environment (or set of environments), due to either safety, cost or time constraints. 
We can distinguish two main cases in real-world applications:
\begin{enumerate}
\item The agent may not be able to interact with the true environment but only with an inaccurate simulation of it. This scenario occurs for instance in robotics \citep{zhu2016target, gu2017deep}. When first learning in a simulation, the difference with the real-world domain is known as the \textit{reality gap} (see e.g. \cite{jakobi1995noise}).
\item The acquisition of new observations may not be possible anymore (e.g., the batch setting).
This scenario happens for instance in medical trials, in tasks with dependence on weather conditions or in trading markets (e.g., energy markets and stock markets).
\end{enumerate}
Note that a combination of the two scenarios is also possible in the case where the dynamics of the environment may be simulated but where there is a dependence on an exogenous time series that is only accessible via limited data \citep{franccois2016deep}.

In order to deal with these limitations, different elements are important:
\begin{itemize}
\item One can aim to develop a simulator that is as accurate as possible.
\item One can design the learning algorithm so as to improve generalization and/or use transfer learning methods (see Chapter~\ref{ch:generalization}).
\end{itemize}


\section{Relations between deep RL and neuroscience}
One of the interesting aspects of deep RL is its relations to neuroscience.
During the development of algorithms able to solve challenging sequential decision-making tasks, biological plausibility was not a requirement from an engineering standpoint.
However, biological intelligence has been a key inspiration for many of the most successful algorithms. 
Indeed, even the ideas of reinforcement and deep learning have strong links with neuroscience and biological intelligence.

\paragraph{Reinforcement}
In general, RL has had a rich conceptual relationship to neuroscience.
RL has used neuroscience as an inspiration and it has also been a tool to explain neuroscience phenomena~\citep{niv2009reinforcement}.
RL models have also been used as a tool in the related field of neuroeconomics~\citep{camerer2005neuroeconomics}, which uses models of human decision-making to inform economic analyses.

The idea of reinforcement (or at least the term) can be traced back to the work of \citet{pavlov1927conditioned} in the context of animal behavior.
In the Pavlovian conditioning model, reinforcement is described as the strengthening/weakening effect of a behavior whenever that behavior is preceded by a specific stimulus.
The Pavlovian conditioning model led to the development of the Rescorla-Wagner Theory \citep{rescorla1972theory}, which assumed that learning is driven by the error between predicted and received reward, among other prediction models.
In computational RL, those concepts have been at the heart of many different algorithms, such as in the development of temporal-difference (TD) methods \citep{sutton1984temporal, schultz1997neural, russek2017predictive}.
These connections were further strengthened when it was found that the dopamine neurons in the brain act in a similar manner to TD-like updates to direct learning in the brain~\citep{schultz1997neural}.

Driven by such connections, many aspects of reinforcement learning have also been investigated directly to explain certain phenomena in the brain.
For instance, computational models have been an inspiration to explain cognitive phenomena
such as exploration~\citep{cohen2007should} and temporal discounting of rewards~\citep{story2014does}.
In cognitive science, \citet{kahneman2011thinking} has also described that there is a dichotomy between two modes of thoughts: a "System 1" that is fast and instinctive and a "System 2" that is slower and more logical. In deep reinforcement, a similar dichotomy can be observed when we consider the model-free and the model-based approaches.
As another example, the idea of having a meaningful abstract representation in deep RL can also be related to how animals (including humans) think. Indeed, a conscious thought at a particular time instant can be seen as a low-dimensional combination of a few concepts in order to take decisions \citep{bengio2017consciousness}.

There is a dense and rich literature about the connections between RL and neuroscience and, as such, the reader is referred to the work of \citet{sutton1998introduction,niv2009reinforcement,lee2012neural,holroyd2002neural,dayan2008reinforcement,dayan2008decision,montague2013reinforcement,niv2009theoretical} for an in-depth history of the development of reinforcement learning and its relations to neuroscience.

\paragraph{Deep learning}
Deep learning also finds its origin in models of neural processing in the brain of biological entities.
However, subsequent developments are such that deep learning has become partly incompatible with current knowledge of neurobiology \citep{bengio2015towards}.
There exists nonetheless many parallels.
One such example is the convolutional structure used in deep learning that is inspired by the organization of the animal visual cortex \citep{fukushima1982neocognitron, lecun1998gradient}.

\paragraph{}
Much work is still needed to bridge the gap between machine learning and general intelligence of humans (or even animals).
Looking back at all the achievements obtained by taking inspiration from neuroscience, it is natural to believe that further understanding of biological brains could play a vital role in building more powerful algorithms and conversely. In particular, we refer the reader to the survey by \cite{hassabis2017neuroscience} where the bidirectional influence between deep RL and neuroscience is discussed. 

\chapter{Conclusion}
\label{ch:conclusion}
Sequential decision-making remains an active field of research with many theoretical, methodological and experimental challenges still open.
The important developments in the field of deep learning have contributed to many new avenues where RL methods and deep learning are combined.
In particular, deep learning has brought important generalization capabilities, which opens new possibilities to work with large, high-dimensional state and/or action spaces.
There is every reason to think that this development will continue in the coming years with more efficient algorithms and many new applications.

\section{Future development of deep RL}
In deep RL, we have emphasized in this manuscript that one of the central questions is the concept of generalization.
To this end, the new developments in the field of deep RL will surely develop the current trends of taking explicit algorithms and making them differentiable so that they can be embedded in a specific form of neural network and can be trained end-to-end.
This can bring algorithms with richer and smarter structures that would be better suited for reasoning on a more abstract level, which would allow to tackle an even wider range of applications than they currently do today.
Smart architectures could also be used for hierarchical learning where much progress is still needed in the domain of temporal abstraction.

We also expect to see deep RL algorithms going in the direction of meta-learning and lifelong learning where previous knowledge (e.g., in the form of pre-trained networks) can be embedded so as to increase performance and training time.
Another key challenge is to improve current transfer learning abilities between simulations and real-world cases.
This would allow learning complex decision-making problems in simulations (with the possibility to gather samples in a flexible way), and then use the learned skills in real-world environments, with applications in robotics, self-driving cars, etc.

Finally, we expect deep RL techniques to develop improved curiosity driven abilities to be able to better discover by themselves their environment.

\section[Applications and societal impact of deep RL]{Applications and societal impact of deep RL and artificial intelligence in general}
In terms of applications, many areas are likely to be impacted by the possibilities brought by deep RL.
It is always difficult to predict the timelines for the different developments, but the current interest in deep RL could be the beginning of profound transformations in information and communication technologies, with applications in clinical decision support, marketing, finance, resource management, autonomous driving, robotics, smart grids, and more.

Current developments in artificial intelligence (both for deep RL or in general for machine learning) follows the development of many tools brought by information and communications technologies.
As with all new technologies, this comes with different potential opportunities and challenges for our society.

On the positive side, algorithms based on (deep) reinforcement learning promise great value to people and society.
They have the potential to enhance the quality of life by automating tedious and exhausting tasks with robots \citep{levine2016end, gandhi2017learning, pinto2017asymmetric}.
They may improve education by providing adaptive content and keeping students engaged \citep{mandel2014offline}.
They can improve public health with, for instance, intelligent clinical decision-making \citep{fonteneau2008variable,bennett2013artificial}.
They may provide robust solutions to some of the self-driving cars challenges \citep{bojarski2016end, you2017virtual}.
They also have the possibility to help managing ecological resources \citep{dietterich2009machine} or reducing greenhouse gas emissions by, e.g., optimizing traffic \citep{li2016traffic}.
They have applications in computer graphics, such as for character animation \citep{2017-TOG-deepLoco}.
They also have applications in finance \citep{deng2017deep}, smart grids \citep{franccois2017contributions}, etc.

However, we need to be careful that deep RL algorithms are safe, reliable and predictable \citep{amodei2016concrete,bostrom2017superintelligence}.
As a simple example, to capture what we want an agent to do in deep RL, we frequently end up, in practice, designing the reward function, somewhat arbitrarily.
Often this works well, but sometimes it produces unexpected, and potentially catastrophic behaviors. For instance, to remove a certain invasive species from an environment, one may design an agent that obtains a reward every time it removes one of these organisms. However, it is likely that to obtain the maximum cumulative rewards, the agent will learn to let that invasive species develop and only then would eliminate many of the invasive organisms, which is of course not the intended behavior.
All aspects related to safe exploration are also potential concerns in the hypothesis that deep RL algorithms are deployed in real-life settings.

In addition, as with all powerful tools, deep RL algorithms also bring societal and ethical challenges \citep{brundage2018malicious}, raising the question of how they can be used for the benefit of all.
Even tough different interpretations can come into play when one discusses human sciences, we mention in this conclusion some of the potential issues that may need further investigation.

The ethical use of artificial intelligence is a broad concern.
The specificity of RL as compared to supervised learning techniques is that it can naturally deal with sequences of interactions, which is ideal for chatbots, smart assistants, etc.
As it is the case with most technologies, regulation should, at some point, ensure a positive impact of its usage.

In addition, machine learning and deep RL algorithms will likely yield to further automation and robotisation than it is currently possible.
This is clearly a concern in the context of autonomous weapons, for instance \citep{walsh2017s}.
Automation also influences the economy, the job market and our society as a whole.
A key challenge for humanity is to make sure that the future technological developments in artificial intelligence do not create an ecological crisis \citep{harari2017sapiens} or deepen the inequalities in our society with potential social and economic instabilities \citep{piketty2013capital}.

We are still at the very first steps of deep RL and artificial intelligence in general.
The future is hard to predict; however, it is key that the potential issues related to the use of these algorithms are progressively taken into consideration in public policies. If that is the case, these new algorithms can have a positive impact on our society.

\appendix
\section*{}
\pagestyle{empty}
\addcontentsline{toc}{subsection}{Deep RL frameworks}
\renewcommand{\thesubsection}{\Alph{subsection}}

\subsection{Deep RL frameworks}
\label{app:frameworks}
Here is a list of some well-known frameworks used for deep RL:

\begin{itemize}
  \item DeeR \citep{franccoislavet2016deer} is focused on being (i) easily accessible and (ii) modular for researchers. 
  \item Dopamine \citep{dopamine} provides standard algorithms along with baselines for the ATARI games. 
  \item ELF \citep{tian2017elf} is a research platform for deep RL, aimed mainly to real-time strategy games. 
  \item OpenAI baselines \citep{baselines} is a set of popular deep RL algorithms, including DDPG, TRPO, PPO, ACKTR.
  The focus of this framework is to provide implementations of baselines.
  \item PyBrain \citep{schaul2010pybrain} is a machine learning library with some RL support.
  \item rllab \citep{duan2016benchmarking} provides a set of benchmarked implementations of deep RL algorithms.
  \item TensorForce~\citep{schaarschmidt2017tensorforce} is a framework for deep RL built around Tensorflow with several algorithm implementations.
  It aims at moving reinforcement computations into the Tensorflow graph for performance gains and efficiency. As such, it is heavily tied to the Tensorflow deep learning library. It provides many algorithm implementations including TRPO, DQN, PPO, and A3C.
\end{itemize}

Even though, they are not tailored specifically for deep RL, we can also cite the two following frameworks for reinforcement learning:
\begin{itemize}
\item RL-Glue \citep{tanner2009rl} provides a standard interface that allows to connect RL agents, environments, and experiment programs together.
\item RLPy \citep{geramifard2015rlpy} is a framework focused on value-based RL using linear function approximators with discrete actions.
\end{itemize}

Table \ref{tab:summary_frameworks} provides a summary of some properties of the aforementioned libraries.

\begin{table}[ht!]
\begin{center}
\begin{tabular}{ |p{3cm}|p{1.6cm}|p{1.6cm}|p{2.2cm}|  }
\hline
Framework  &  Deep RL  & Python interface & Automatic GPU support \\
\hline
DeeR & yes & yes   & yes \\
Dopamine & yes & yes   & yes \\
ELF    & yes & no & yes \\
OpenAI baselines    & yes & yes & yes \\
PyBrain & yes & yes & no \\
RL-Glue & no & yes & no \\
RLPy & no & yes & no \\
rllab    & yes & yes & yes \\
TensorForce   & yes & yes & yes \\
\hline
\end{tabular}
\end{center}
\caption{Summary of some characteristics for a few existing RL frameworks.}
\label{tab:summary_frameworks}
\end{table}

\backmatter  

\printbibliography

\end{document}